%% file: ms.tex
\documentclass[twoside,11pt]{article}

\usepackage{jmlr2e}
\usepackage{booktabs}
\usepackage{amsmath}
\usepackage{tikz}
\usepackage[font=small,format=hang]{caption}
\usepackage{subcaption}
\usepackage{fontawesome}
\usepackage{ifthen}
\usepackage{xstring}
\usepackage{pgfopts}
\usepackage{algorithm}
\usepackage[noend]{algpseudocode}
\usepackage{hyperref}

\usetikzlibrary{positioning, backgrounds, decorations.pathreplacing, arrows, arrows.meta, calc}

\pdfoutput=1

\hypersetup{pdfauthor={J. Gomez Robles},pdftitle={Learning to reinforcement learn for Neural Architecture Search},pdfkeywords={}}

\ShortHeadings{Learning to reinforcement learn for Neural Architecture Search}{J. Gomez Robles and J. Vanschoren}
\firstpageno{1}

\begin{document}

\title{Learning to reinforcement learn for Neural Architecture Search}

\author{\name Jorge Gomez Robles \email j.gomez.robles@student.tue.nl \\
\name Joaquin Vanschoren \email j.vanschoren@tue.nl\\ 
       \addr Department of Mathematics and Computer Science\\
       Eindhoven University of Technology\\
       Eindhoven 5612 AZ, The Netherlands
}

\maketitle

\input{abstract.tex}
\input{introduction.tex}
\input{preliminaries.tex}
\input{related.tex}
\input{methodology.tex}
\input{software.tex}
\input{experiments.tex}
\input{results.tex}
\input{conclusions.tex}

\input{ack.tex}

\vskip 0.2in
\bibliography{ms}

\newpage
\appendix
\input{appendix_datasets.tex}

\input{appendix_networks.tex}

\end{document}

%% file: abstract.tex
\begin{abstract}
Reinforcement learning (RL) is a goal-oriented learning solution that has proven to be successful for Neural Architecture Search (NAS) on the CIFAR and ImageNet datasets. However, a limitation of this approach is its high computational cost, making it unfeasible to replay it on other datasets. Through meta-learning, we could bring this cost down by adapting previously learned policies instead of learning them from scratch. In this work, we propose a deep meta-RL algorithm that learns an adaptive policy over a set of environments, making it possible to transfer it to previously unseen tasks. The algorithm was applied to various proof-of-concept environments in the past, but we adapt it to the NAS problem. We empirically investigate the agent's behavior during training when challenged to design chain-structured neural architectures for three datasets with increasing levels of hardness, to later fix the policy and evaluate it on two unseen datasets of different difficulty. Our results show that, under resource constraints, the agent effectively adapts its strategy during training to design better architectures than the ones designed by a standard RL algorithm, and can design good architectures during the evaluation on previously unseen environments. We also provide guidelines on the applicability of our framework in a more complex NAS setting by studying the progress of the agent when challenged to design multi-branch architectures.
\end{abstract}

\begin{keywords}
  Neural Architecture Search, Deep Meta-Reinforcement Learning, Image Classification
\end{keywords}

%% file: introduction.tex
\section{Introduction}

Neural networks have achieved remarkable results in many fields, such as that of Image Classification. Crucial aspects of this success are the choice of the neural architecture and the chosen hyperparameters for the particular dataset of interest; however, this is not always straightforward. Although state-of-the-art neural networks can inspire the design of other architectures, this process heavily relies on the designer's level of expertise, making it a challenging and cumbersome task that is prone to deliver underperforming networks.

In an attempt to overcome these flaws, researchers have explored various techniques under the name of Neural Architecture Search (NAS)~\citep{NASsurvey}. In NAS, the ultimate goal is to come up with an algorithm that takes any arbitrary dataset as input and outputs a well-performing neural network for some learning task of interest, so that we can accelerate the design process and remove the dependency on human intervention. Nevertheless, coming up with a solution of this kind is a complicated endeavor where researchers have to deal with several aspects such as the type of the networks that they consider, the scope of the automation process, or the search strategy applied. A particular search strategy for NAS is reinforcement learning (RL), where a so-called \textit{agent} learns how to design neural networks by sampling architectures and using their numeric performance on a specific dataset as the reward signals that guide the search. Popular standard RL algorithms such as \textsc{Q-learning} or \textsc{Reinforce} have been used to design state-of-the-art Convolutional Neural Networks (CNNs) for classification tasks on the CIFAR and ImageNet datasets \citep{ENAS, PathNAS, BlockQNN, ZophNAS1, BakerNAS}, but little attention is paid to deliver architectures for other datasets. In an attempt to fill that gap, a suitable alternative is deep meta-RL~\citep{LtRL, RL2}, where the \textit{agent} acts on various environments to learn an \textit{adaptive} policy that can be transferred to new environments.

In this work, we apply deep meta-RL to NAS, which, to the best of our knowledge, is a novel contribution. The environments that we consider are associated with standard image classification tasks on datasets with different levels of hardness sampled from a meta-dataset~\citep{MetaDataset}. Our main experiments focus on the design of chain-structured networks and show that, under resource constraints, the resulting policy can adapt to new environments, outperform standard RL, and design better architectures than the ones inspired by state-of-the-art networks. We also experiment with extending our approach to the design of multi-branch architectures so that we can give directions for future work.

The remainder of this report is structured as follows. First, in  Section ~\ref{sec:preliminaries}, we introduce the preliminary concepts required to understand our work. Next, in Section~\ref{sec:related}, we discuss the related work for both reinforcement learning and NAS. In Section~\ref{sec:methodology}, we formally introduce our methodology, and in Section~\ref{sec:software}, the framework developed to implement it. In Section~\ref{sec:experiments}, we define the experiments, and in Section~\ref{sec:results}, we show the results. Finally, in Section~\ref{sec:conclusions}, the conclusions are set out.

%% file: preliminaries.tex
\section{Preliminaries}\label{sec:preliminaries}

\subsection{Reinforcement learning}\label{sec:preliminaries:rl}

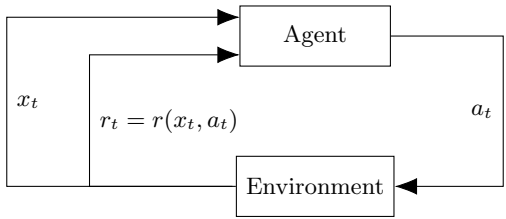
\begin{figure}[ht]
\begin{center}
\begin{tikzpicture}

\tikzstyle{box} = [rectangle, minimum width=2cm, minimum height=0.8cm, text centered, draw=black, fill=gray!0]

\node (agent) [box] at (0,0) {\footnotesize Agent};
\node (env) [box, below of=agent, yshift=-1cm] {\footnotesize Environment};

\draw [-{Latex[length=3mm]}] 
    (env) -- +(-3,0) |- node[right, pos=0.25] {\footnotesize $r_t=r(x_t, a_t)$} ($(agent)+(-1, -0.25)$);

\draw [-{Latex[length=3mm]}] 
    ($(env)+(-1.1, 0)$) -- +(-3, 0) |- node[right, pos=0.25] {\footnotesize $x_t$} ($(agent)+(-1, 0.25)$);
    
\draw [-{Latex[length=3mm]}] 
    (agent) -- +(2.5,0) |- node[left, pos=0.25] {\footnotesize $a_t$} (env);

\end{tikzpicture}
\caption{Graphic representation of the reinforcement learning interaction. Every time the agent performs an action $a_t$, the environment modifies its state $x_{t-1}$ to $x_t$, computes the reward $r_t$ and sends both values to the agent, who uses them to optimize its policy.}
\label{fig:preliminaries:rl}
\end{center}
\end{figure}

Reinforcement learning (RL) is an approach to automate goal-directed learning~\citep{RLIntroBook}. It relies on two entities that interact with each other: an \textit{environment} that delivers information of its \textit{state}, and an \textit{agent} that using such information learns how to achieve a \textit{goal} in the environment. The interaction is a bilateral communication where the \text{agent} performs \textit{actions} to modify the \textit{state} of the environment, which responds with a numeric \textit{reward} measuring how good the action was to achieve the \textit{goal}. Typically, the sole interest of the agent is to improve its decision-making strategy, known as the \textit{policy}, to maximize the total reward received over the whole interaction trial since this will lead it to the desired goal. More strictly, RL is formalized using finite Markov Decision Processes (MDPs) as in Definition~\ref{def:preliminaries:rl} borrowed from~\cite{RL2}, resulting in the agent-environment interaction illustrated in Figure~\ref{fig:preliminaries:rl}.

\begin{definition}[Reinforcement Learning]
\label{def:preliminaries:rl}
We define a discrete-time finite-horizon discounted MDP $M = (\mathcal{X}, \mathcal{A}, \mathcal{P}, r, \rho_0, \gamma, T)$, in which $\mathcal{X}$ is a state set, $\mathcal{A}$ an action set, $\mathcal{P}: \mathcal{X} \times \mathcal{A} \times \mathcal{X} \mapsto \mathbb{R}_+$ a transition probability distribution, $r: \mathcal{X} \times \mathcal{A} \mapsto [-R_{max}, R_{max}]$ a bounded reward function, $\rho_0 : \mathcal{X} \mapsto \mathbb{R}_+ $ an initial state distribution, $\gamma \in [0,1]$ a discount factor, and $T$ the horizon.
\textsc{Reinforcement learning} typically aims to optimize a stochastic policy $\pi_\theta : \mathcal{X} \times \mathcal{A} \mapsto \mathbb{R}_+$ by maximizing the expected reward, modeled as $\eta(\pi_\theta) = \mathbb{E}_\tau[\sum_{t=0}^{T}{\gamma^tr(x_t, a_t)}]$, where $\tau = (s_0, \alpha_0, ...)$ denotes the whole trajectory, $x_t \in \mathcal{X}$, $x_0 \sim \rho_0(x_0)$, $a_t \in \mathcal{A}$, $a_t \sim \pi_\theta(a_t | x_t)$, and $x_{t+1} \sim \mathcal{P}(x_{t+1}| x_t, a_t)$.

\end{definition}

\subsection{Neural Architecture Search}\label{sec:preliminaries:nas}

\begin{figure}[ht]
\begin{center}
\begin{tikzpicture}

\tikzstyle{box} = [rectangle, minimum width=2cm, minimum height=1cm, text width=2.3cm, text centered, draw=black, fill=gray!0]

\tikzstyle{arrow} = [->,>=stealth]

\node (ss) [box] at (0,0) {\small Search space $\mathcal{X}$};
\node (st) [box, right of=ss, xshift=3cm] {\small Search strategy};
\node (pss) [box, right of=st, xshift=5cm] {\small Performance estimation strategy $r$};

\draw [arrow] (ss) -- (st);

\draw [arrow] (st.east) to [out=30,in=150] node [above,midway,text centered] {\footnotesize $ A \in \mathcal{X}$ } (pss.west);

\draw [arrow] (pss.west) to [out=210,in=330] node [below,midway, text width=2cm, text centered] {\footnotesize performance estimate of $A$} (st.east);

\end{tikzpicture}
\caption{An illustration of the three NAS variables interacting~\citep{NASsurvey}. At any moment during the search, the \textit{search strategy} samples an architecture $A$ from the \textit{search space} and sends it to the \textit{performance estimation strategy}, which returns the performance estimate. By design, the \textit{search space} and the \textit{performance estimation strategy} are named after the variables in Definition~\ref{def:preliminaries:rl} since they are typically equivalent in NAS within the reinforcement learning setting.}
\label{fig:preliminaries:nas}
\end{center}
\end{figure}
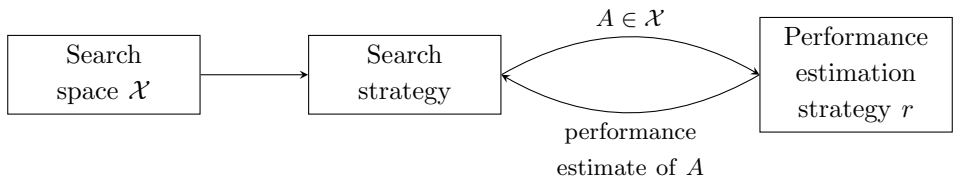

Neural Architecture Search (NAS) is the process of automating the design of neural networks. In order to formalize this definition, it is convenient to refer to the survey of~\citet{NASsurvey}, which characterize a NAS work with three variables: the \textit{search space}, the \textit{search strategy}, and the \textit{performance estimation strategy}. Figure~\ref{fig:preliminaries:nas} illustrates the interaction between these variables.

\begin{figure}[ht]
\begin{center}
\begin{tikzpicture}

\tikzstyle{squareA} = [rectangle, minimum width=0.4cm, minimum height=0.4cm, draw=black, fill=gray!0]

\tikzstyle{squareB} = [rectangle, minimum width=0.4cm, minimum height=0.4cm, draw=black, fill=gray!50]

\tikzstyle{squareC} = [rectangle, minimum width=0.4cm, minimum height=0.4cm, draw=black, fill=gray!100]

\tikzstyle{arrow} = [->,>=stealth]

\node (aa) [squareA] at (-0.75,-0.5) {};
\node (ab) [squareB, below of=aa] {};
\node (ac) [squareC, below of=ab] {};
\node (ad) [squareA, below of=ac] {};
\node (ae) [squareB, below of=ad] {};
\node (af) [squareC, below of=ae] {};

\draw [arrow] (aa) -- (ab);
\draw [arrow] (ab) -- (ac);
\draw [arrow] (ac) -- (ad);
\draw [arrow] (ad) -- (ae);
\draw [arrow] (ae) -- (af);

\draw[dashed] (0.75, 0.5) -- (0.75, -6.5);

\node (ba) [squareA] at (3,-2) {};
\node (bb) [squareB, below of=ba, left of=ba] {};
\node (bc) [squareB, below of=ba] {};
\node (bd) [squareB, below of=ba, right of=ba] {};
\node (be) [squareC, below of=bc] {};

\draw [arrow] (ba) -- (bb);
\draw [arrow] (ba) -- (bc);
\draw [arrow] (ba) -- (bd);
\draw [arrow] (bb) -- (be);
\draw [arrow] (bc) -- (be);
\draw [arrow] (bd) -- (be);

\draw[dashed] (5.25, 0.5) -- (5.25, -6.5);

\node (ca) [squareA] at (9,0) {};
\node (cb) [squareB, below of=ca, left of=ca] {};
\node (cc) [squareB, below of=ca] {};
\node (cd) [squareB, below of=ca, right of=ca] {};
\node (ce) [squareC, below of=cc] {};

\draw [arrow] (ca) -- (cb);
\draw [arrow] (ca) -- (cc);
\draw [arrow] (ca) -- (cd);
\draw [arrow] (cb) -- (ce);
\draw [arrow] (cc) -- (ce);
\draw [arrow] (cd) -- (ce);

\node (da) [squareA] at (7.5,-3) {};
\node (db) [squareB, below of=da, left of=da] {};
\node (dc) [squareB, below of=da] {};
\node (dd) [squareB, below of=da, right of=da] {};
\node (de) [squareC, below of=dc] {};

\draw [arrow] (da) -- (db);
\draw [arrow] (da) -- (dc);
\draw [arrow] (da) -- (dd);
\draw [arrow] (db) -- (de);
\draw [arrow] (dc) -- (de);
\draw [arrow] (dd) -- (de);

\node (ea) [squareA] at (10.5,-3) {};
\node (eb) [squareB, below of=ea, left of=ea] {};
\node (ec) [squareB, below of=ea] {};
\node (ed) [squareB, below of=ea, right of=ea] {};
\node (ee) [squareC, below of=ec] {};

\draw [arrow] (ea) -- (eb);
\draw [arrow] (ea) -- (ec);
\draw [arrow] (ea) -- (ed);
\draw [arrow] (eb) -- (ee);
\draw [arrow] (ec) -- (ee);
\draw [arrow] (ed) -- (ee);

\draw [arrow] (ce) -- (da);
\draw [arrow] (ce) -- (ea);

\node (f) [squareC, below of=de, right of=de, xshift=0.5cm] {};

\draw [arrow] (de) -- (f);
\draw [arrow] (ee) -- (f);

\end{tikzpicture}
\caption{Examples of networks belonging to different search spaces. On the left, a \textit{chain-structured} network. On the center, a \textit{multi-branch} network. On the right, the same multi-branch structure used as a \textit{cell} repeated multiple times to build a more complex network.}
\label{fig:preliminaries:ss}
\end{center}
\end{figure}
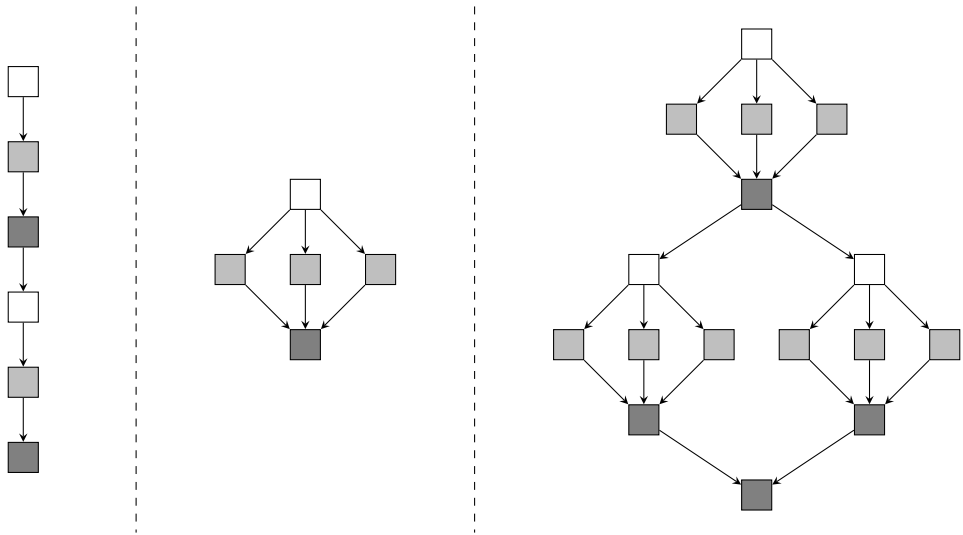

The \textit{search space} is the set of architectures considered in the search process. It is possible to define different spaces by constraining attributes of the networks, such as the maximum depth allowed, the type of layers to use, or the connections permitted between layers. A common abstraction inspired in popular networks is to separate the search spaces in \textit{chain} structures and \textit{multi-branch} structures that can be either \textit{complete} neural networks or \textit{cells} that can be used to build more complex networks, as illustrated in Figure~\ref{fig:preliminaries:ss}.

On the other hand, the \textit{search strategy} is simply the algorithm used to perform the search. The choices range from naive approaches such as random search to more sophisticated ones like reinforcement learning~\citep{BakerNAS, ZophNAS1}, evolutionary algorithms~\citep{AmoebaNet}, or gradient descent search~\citep{DARTS}.

Lastly, the \textit{performance estimation strategy} is the function used to measure the goodness of the sampled architectures. Formally, it is a function $R_{D}: \mathcal{X} \mapsto \mathbb{R}$ evaluating an architecture on a dataset $D$. The \textit{vanilla} estimation strategy is the test accuracy after training of a network, but different alternatives are proposed to try delivering an accurate estimate in a short time since expensive training creates a bottleneck in the search process.

%% file: related.tex
\section{Related work}\label{sec:related}

\subsection{Reinforcement learning}\label{sec:related:rl}

The key to reinforcement learning is the algorithm used to optimize the \textit{policy} $\pi_\theta$ (see Definition~\ref{def:preliminaries:rl}). Through the years, researchers have proposed different algorithms, such as \textsc{Reinforce}~\citep{Reinforce}, Q-learning~\citep{Qlearning}, Actor-Critic~\citep{ActorCritic}, Deep-Q-Network (DQN)~\citep{DQN}, Trust Region Policy Optimization (TRPO)~\citep{TRPO}, and Proximal Policy Optimization (PPO)~\citep{PPO}. These algorithms have successfully solved problems in a variety of fields, from robotics~\citep{RLRobotics} and video games~\citep{MonteCarlo, DQN} to traffic-control~\citep{RLTraffic} or computational resources management~\citep{RLresourcemanagement}, showing the power and utility of the reinforcement learning framework. Despite its success, a theoretical flaw of RL is that the policy learned only captures the one-to-one \textit{state-action} relation of the environment in question, making it necessary to perform individual runs for every new environment of interest. The latter is a costly trait of this standard form of RL since it typically requires thousands of steps to converge to an optimal policy.

A research area that addresses this problem is meta-RL, in which agents are trained to learn transferable policies that do not require training from scratch on new problems. We identify two types of meta-RL algorithms: the ones that learn a \textit{good} initialization for the neural networks representing the policy\footnote{The term \textit{deep} meta-reinforcement learning comes from the usage of \textit{deep} models, such as neural networks, to represent the policy to be learned.}, and the ones that learn policies that can adapt their decision-making strategy to new environments, ideally without further training. First, Model Agnostic Meta-Learning (MAML)~\citep{MAML} learns an initialization that allows few-shot learning in RL for a set of similar environments.
Second, two algorithms have been proposed to learn policies that, once deployed, can adapt their decision-making strategy to new environments, ideally without further training: \textit{Learning to reinforcement learn}~\citep{LtRL} and RL$^2$~\citep{RL2}. These methods aim to learn a more sophisticated policy modeled by a Recurrent Neural Network (RNN) that captures the relation between states, actions, and meta-data of past actions. What emerges is a policy that can adapt its strategy to different environments. The main difference between the two works is the set of environments considered: for ~\citet{LtRL} they come from a parameterized distribution, whereas for~\citet{RL2} they are relatively unrelated. We believe that the latter makes these algorithms more suitable for scenarios where a distribution of environments cannot be guaranteed, although the \textit{general universality of the gradient-based learner}~\citep{MAMLUniversality} suggests that MAML could address this scenario too. Third, Simple Neural AttentIve Learner (SNAIL) extends the idea behind \textit{Learning to reinforcement learn} and RL$^2$ by using a more powerful temporal-focused model than the simple RNN. We note that none of these approaches have been applied to NAS.

\subsection{Neural Architecture Search}\label{sec:related:nas}

As introduced earlier in Section~\ref{sec:preliminaries:nas}, it is possible to address Neural Architecture Search (NAS) in different ways. Remarkable results have been achieved by applying optimization techniques such as Bayesian optimization, evolutionary algorithms, gradient-based search, and reinforcement learning. We are interested in reinforcement learning for NAS, due to the variety of works that have achieved state-of-the-art results.
For other work in NAS, we refer to the survey of~\citet{NASsurvey}.


Although the ultimate goal of NAS is to come up with a straightforward fully-automated solution that can deliver a neural architecture for a machine learning task on any dataset of interest, there exist several factors that impede that ambition. Perhaps the most important of these factors is the high computational cost of NAS with reinforcement learning, which imposes constraints on different elements that impact the scope of the solutions. The first bottleneck is the computation of the reward, which typically is the test accuracy of the sampled architectures after training. Because of the expensiveness of such computation,  researchers have proposed various \textit{performance estimation strategies} to avoid expensive training procedures, and they have also imposed some constraints over the \textit{search space} considered so that a lower number of architectures get sampled and evaluated. For the first aspect, we observe several relaxations: reducing the number of training epochs~\citep{BakerNAS,BlockQNN}, sharing of weights between similar networks~\citep{PathNAS, ENAS}, or entirely skipping the train-evaluation procedure by predicting the performance of the architectures~\citep{BlockQNN}. Although these alternatives have successfully reduced the computation time, they pay little attention to the effect of their potentially unfair estimations on the decision-making of the agent, and therefore, one should treat them carefully. On the other hand, for the \textit{search space}, crucial choices are the cardinality of the space and the complexity of the architectures. Some researchers opt for ample spaces with various types of layers and no restrictions in their connections~\citep{ZophNAS1,ZophNAS2,BlockQNN}, whereas others prefer them smaller, such as a chain-structured space~\citep{BakerNAS}, or a multi-branch space modeled as a fully-connected graph with low cardinality~\citep{ENAS}.

It is important to note that an approach can dramatically reduce its computation time with relaxations on the search space alone. For instance, the same methodology can decrease its computational cost by a factor of 7 (28 days to 4 days, with hundreds of GPUs in both cases) if the space is restricted in the types of layers and the number of elements allowed in the architectures~\citep{ZophNAS1, ZophNAS2}. Furthermore, a constrained search space used jointly with some performance estimation strategy can reduce the cost to only 1 day with 1 GPU such as in BlockQNN-V2~\citep{BlockQNN} and ENAS~\citep{ENAS}; however, this drastic reduction of the computational time should be treated with caution. In the case of BlockQNN-V2, the estimation of the performance of the networks (i.e., accuracy at a given epoch) depends on a surrogate prediction model that is not studied in detail by the authors, thus leaving room for potentially wrong predictions. On the other hand, a recent investigation~\citep{ENASbad} shows that the quality of the networks delivered by ENAS is not a consequence of reinforcement learning, but of the search space, which contains a majority of well-performing architectures that can be explored with a less expensive procedure such as random search, therefore losing its character of \textit{artificially smart} search.

Another factor impacting a NAS with reinforcement learning work is the \textit{input dataset} used. Although they usually transfer the best CIFAR-based architecture designed by the agent to the ImageNet dataset~\citep{BakerNAS, ZophNAS1, ZophNAS2, PathNAS, ENAS}, none of them make the agent design networks for other datasets. Furthermore, none of the works give insight on how using a different dataset could affect the complexity of the search. We believe that the lack of study for other datasets is ascribed to the costly task-oriented design of the reinforcement learning algorithms used, \textsc{Q-learning} and \textsc{Reinforce}, that requires to train the agent from scratch for every environment (i.e., a dataset) of interest. The authors do not justify the choice of these algorithms; hence, it would be desirable to study other reinforcement learning algorithms in the same NAS scenarios.

%% file: methodology.tex
\section{Methodology}\label{sec:methodology}

We aim to improve the performance of Neural Architecture Search (NAS) with reinforcement learning (RL) by using meta-learning. We, therefore, build a meta-RL system that can learn across environments and adapt to them. We split the system into two components: the NAS variables and the reinforcement learning framework. For the reinforcement learning framework, we make use of a deep meta-RL algorithm that follows the same line of \textit{Learning to reinforcement learn}~\citep{LtRL} and \textit{RL$^2$}~\citep{RL2}, with some minor adaptations in the meta-data employed and the design of the episodes. The environments that we consider are neural architecture design tasks for different datasets sampled from the meta-dataset collection~\citep{MetaDataset}. On the other hand, for the NAS elements, we work with a slightly modified version of the search space of BlockQNN~\citep{BlockQNN} and similarly, we use the test accuracy after early-stop training as the reward associated with the sampled architectures. In the remainder of the section, we discuss these elements further.

\subsection{The Neural Architecture Search elements}\label{sec:methodology:nas}

As described in Section~\ref{sec:preliminaries:nas}, three NAS variables characterize a research work in this area: the search strategy, the search space, and the performance estimation strategy. In our case, we constrain the search strategy to a deep meta-reinforcement learning algorithm that we explain in detail in Section~\ref{sec:methodology:rl}, and thus, here we only elaborate on the remaining two. 

\subsubsection{The search space}\label{sec:methodology:nas:ss}

The set of architectures considered in our work is inspired by BlockQNN~\citep{BlockQNN}, which defines the search space as all architectures that can be generated by sequentially stacking $d \in \mathbb{N}$ vectors from a so-called Network Structure Code (NSC) space containing encodings of the most relevant layers for CNNs. An NSC vector has information of the type of a layer, the value of its most important hyperparameter, its position on the network, and the allowed incoming connections (i.e., the inputs) so that it becomes possible to represent any architecture as a list of NSCs. The NSC definition is flexible in that it can easily be modified or extended and, moreover, it allows us to define an equivalent discrete action space for the reinforcement learning agent as described in Section~\ref{sec:methodology:rl:as}.

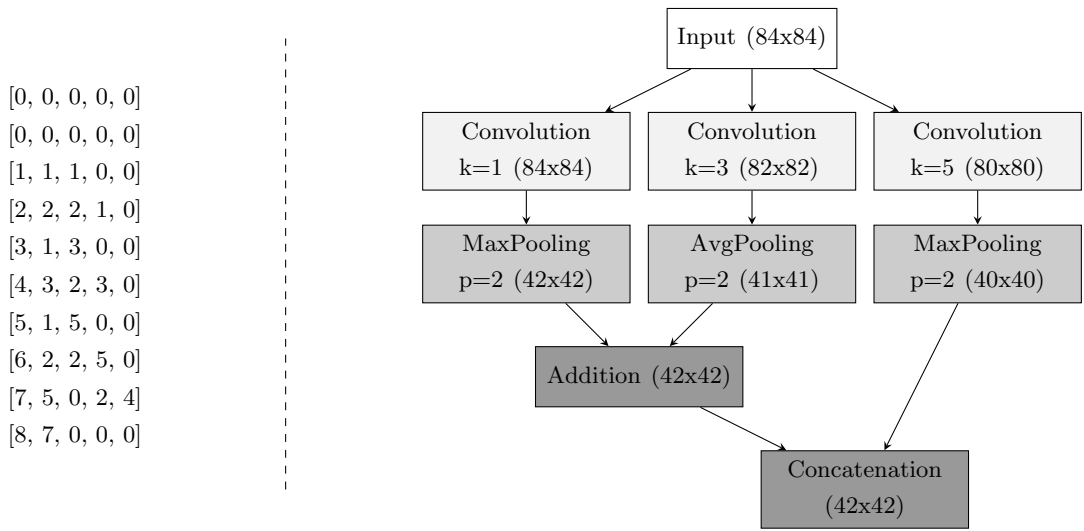
\begin{figure}[ht]
\begin{center}
\begin{tikzpicture}
\node[align=left, below] at (0,-.5){\footnotesize [0, 0, 0, 0, 0]};
\node[align=left, below] at (0,-1.0){\footnotesize [0, 0, 0, 0, 0]};
\node[align=left, below] at (0,-1.5){\footnotesize [1, 1, 1, 0, 0]};
\node[align=left, below] at (0,-2.0){\footnotesize [2, 2, 2, 1, 0]};
\node[align=left, below] at (0,-2.5){\footnotesize [3, 1, 3, 0, 0]};
\node[align=left, below] at (0,-3.0){\footnotesize [4, 3, 2, 3, 0]};
\node[align=left, below] at (0,-3.5){\footnotesize [5, 1, 5, 0, 0]};
\node[align=left, below] at (0,-4.0){\footnotesize [6, 2, 2, 5, 0]};
\node[align=left, below] at (0,-4.5){\footnotesize [7, 5, 0, 2, 4]};
\node[align=left, below] at (0,-5.0){\footnotesize [8, 7, 0, 0, 0]};

\draw[dashed] (2.8, -6) -- (2.8, 0);

\tikzstyle{input} = [rectangle, minimum width=2cm, minimum height=0.8cm, text centered, draw=black, fill=gray!0, text width=2cm]

\tikzstyle{convolution} = [rectangle, minimum width=2cm, minimum height=0.8cm, text centered, draw=black, fill=gray!10, text width=2.5cm]

\tikzstyle{maxpool} = [rectangle, minimum width=2cm, minimum height=0.8cm, text centered, draw=black, fill=gray!40, text width=2.5cm]

\tikzstyle{avgpool} = [rectangle, minimum width=2cm, minimum height=0.8cm, text centered, draw=black, fill=gray!40, text width=2.5cm]

\tikzstyle{concat} = [rectangle, minimum width=2cm, minimum height=0.8cm, text centered, draw=black, fill=gray!80, text width=2.5cm]

\tikzstyle{arrow} = [->,>=stealth]

\node (l0) [input] at (9,0) {\footnotesize Input (84x84)};

\node (l1) [convolution, below of=l0, xshift=-3cm, yshift=-0.5cm] {\footnotesize Convolution k=1 (84x84)};

\node (l3) [convolution, below of=l0, yshift=-0.5cm] {\footnotesize Convolution k=3 (82x82)};

\node (l5) [convolution, below of=l0, xshift=3cm, yshift=-0.5cm] {\footnotesize Convolution k=5 (80x80)};

\node (l2) [maxpool, below of=l1, yshift=-0.5cm] {\footnotesize MaxPooling p=2 (42x42)};

\node (l4) [avgpool, below of=l3, yshift=-0.5cm] {\footnotesize AvgPooling p=2 (41x41)};

\node (l6) [maxpool, below of=l5, yshift=-0.5cm] {\footnotesize MaxPooling p=2 (40x40)};

\node (l7) [concat, below of=l2, yshift=-0.5cm, xshift=1.5cm] {\footnotesize Addition (42x42)};

\node (l8) [concat, below of=l7, yshift=-0.5cm, xshift=3.0cm] {\footnotesize Concatenation (42x42)};

\draw [arrow] (l0) -- (l1);
\draw [arrow] (l0) -- (l3);
\draw [arrow] (l0) -- (l5);

\draw [arrow] (l1) -- (l2);
\draw [arrow] (l3) -- (l4);
\draw [arrow] (l5) -- (l6);

\draw [arrow] (l2) -- (l7);
\draw [arrow] (l4) -- (l7);

\draw [arrow] (l7) -- (l8);
\draw [arrow] (l6) -- (l8);

\end{tikzpicture}
\caption{Example of an architecture sampled from the search space of our approach. On the left, a list of Neural Structure Codes (NSCs); on the right, the corresponding network after the application of the rules. For the sake of simplicity, in this example, the convolutions are assumed to have one filter only.}
\label{fig:methodology:nas:ss:rules}
\end{center}
\end{figure}

In Table~\ref{tab:methodology:nas:ss:nsc}, we present the NSC space for our implementation. Given a list of NSC vectors representing an architecture, the network is built following the next rules: firstly, based on BlockQNN's results, if a \textit{convolution} layer is found then a Pre-activation Convolutional Cell\footnote{The PCC stacks a ReLU, a convolution, and a batch normalization unit.} (PCC)~\citep{PCC} with 32 units\footnote{The selection of the number of units is made to reduce the cost of the training of the networks.} is used; secondly, the \textit{concatenation} and \textit{addition} operations create padded versions of their inputs if they have different shapes; thirdly, if at the end of the building process the network has two or more leaves then they get merged with a \textit{concatenation} operation\footnote{The last two rules do not apply for chain-structured architectures since no merge operations are needed.}. Figure~\ref{fig:methodology:nas:ss:rules} illustrates these rules.

\begin{table}[ht]
\centering
\begin{tabular}{@{}cccccc@{}}
\toprule
\textbf{Name}   & \textbf{Index}  & \textbf{Type} & \textbf{Kernel size\footnotemark{}} & \textbf{Predecessor 1} & \textbf{Predecessor 2} \\ \midrule
Convolution     & \textbf{T}      & 1             & \{1, 3, 5\}          & \textbf{K}             & $\emptyset$            \\
Max Pooling     & \textbf{T}      & 2             & \{2, 3\}             & \textbf{K}             & $\emptyset$            \\
Average Pooling & \textbf{T}      & 3             & \{2, 3\}             & \textbf{K}             & $\emptyset$            \\
Addition        & \textbf{T}      & 5             & $\emptyset$          & \textbf{K}             & \textbf{K}             \\
Concatenation   & \textbf{T}      & 6             & $\emptyset$          & \textbf{K}             & \textbf{K}             \\
Terminal        & \textbf{T}      & 7             & $\emptyset$          & $\emptyset$            & $\emptyset$            \\ \bottomrule
\end{tabular}
\caption{The subset of the NSC space used, presented as in BlockQNN~\citep{BlockQNN}. The changes with respect to the original BlockQNN space are: a) the \textit{identity} operator (\textbf{Type} 4) is omitted; b) the pool size values changed from the original set $\{1, 3\}$ to $\{2, 3\}$ because a pool size of 1 does not contribute to any reduction. The set $\textbf{T}=\{1, 2 \dots , d \}$ refers to the position of each layer in the network, where $d$ is the maximum depth, and $\textbf{K} = \{0, 1, 2, \dots , \text{current layer index} - 1 \}$ the index of its predecessor. }
\label{tab:methodology:nas:ss:nsc}
\end{table}

\footnotetext{The kernel size is an attribute for the convolutions, whereas for the pooling elements it refers to the layer's pool size.}

\subsubsection{The performance estimation strategy}\label{sec:methodology:nas:pss}

Our estimation of the long-term performance of the designed networks closely follows the early-stop approach of BlockQNN-V1~\citep{BlockQNN}, but we ignore the penalization of the network's FLOPs and density since we have empirically ascertained that it is too strict when the classification task is difficult (i.e., when low accuracy values are expected).

The choice of an early-stop strategy is made to help reduce the computational cost of our approach. In short, for every sampled architecture $\mathcal{N}$ a prediction module is appended, and the network is then trained for a small number of epochs to obtain its accuracy on a test set, which is the final estimation of its long-term performance. The datasets considered are balanced, and their train and test splits are designed beforehand (see Section~\ref{sec:methodology:rl:environments}).

As in BlockQNN, for the prediction module we stack a fully-connected dense layer with 1024 units and ReLU activation function, a dropout layer with rate of 0.4, a dense layer with the number of units equals to the desired number of classes to predict and linear activation function, and a softmax that outputs the probabilities per class. The training is performed to minimize the cross-entropy loss using the Adam Optimizer~\citep{Adam} with the parameters used in BlockQNN: $\beta_1=0.9$, $\beta_2=0.999$, $\epsilon_{\textsc{adam}}=10e^{-8}$, and $\alpha_\textsc{adam}=0.001$ that is reduced by a factor of 0.2 every five epochs. After training, the network is evaluated on a test set by fixing the network's weights and selecting the class with the highest probability to be the final prediction per observation in the set so that the standard accuracy $\text{ACC}_{\mathcal{N}}$ can be returned.

\subsection{The reinforcement learning framework}\label{sec:methodology:rl}

The deep-meta-RL framework that we propose is different from standard RL in two main aspects. First, the \textit{agent} is challenged to face more than one \textit{environment} during training, and second, the distribution over the \textit{reward} domain learned by the agent is now dependant on the whole history of \textit{states}, \textit{actions}, and \textit{rewards}, instead of the simple \textit{state-action} pairs. 
In the remainder of the section, we describe each of the RL elements.

\subsubsection{The states}\label{sec:methodology:rl:states}

\begin{figure}[ht]
\begin{center}
\begin{tikzpicture}

\node[align=left, below] at (0,-.5){\footnotesize [0, 0, 0, 0, 0]};
\node[align=left, below] at (0,-1){\footnotesize [1, 1, 5, 0, 0]};
\node[align=left, below] at (0,-1.5){\footnotesize [2, 2, 2, 1, 0]};
\node[align=left, below] at (0,-2){\footnotesize [3, 6, 0, 0, 2]};


\draw[dashed] (2.1, -3) -- (2.1, 0);

\node[align=left, below] at (4,-.5){\footnotesize 1, 0 \dots 0, 0};
\node[align=left, below] at (6.5,-.5){\footnotesize 1, 0, 0, 0, 0, 0};
\node[align=left, below] at (9,-.5){\footnotesize 1, 0, 0, 0, 0};
\node[align=left, below] at (11.25,-.5){\footnotesize 1, 0, 0, 0, 0};

\node[align=left, below] at (4,-1){\footnotesize 0, 1 \dots 0, 0};
\node[align=left, below] at (6.5,-1){\footnotesize 0, 0, 0, 0, 0, 1};
\node[align=left, below] at (9,-1){\footnotesize 1, 0, 0, 0, 0};
\node[align=left, below] at (11.25,-1){\footnotesize 1, 0, 0, 0, 0};

\node[align=left, below] at (4,-1.5){\footnotesize 0, 0 \dots 0, 0};
\node[align=left, below] at (6.5,-1.5){\footnotesize 0, 0, 1, 0, 0, 0};
\node[align=left, below] at (9,-1.5){\footnotesize 0, 1, 0, 0, 0};
\node[align=left, below] at (11.25,-1.5){\footnotesize 1, 0, 0, 0, 0};

\node[align=left, below] at (4,-2){\footnotesize 0, 0 \dots 1, 0};
\node[align=left, below] at (6.5,-2){\footnotesize 1, 0, 0, 0, 0, 0};
\node[align=left, below] at (9,-2){\footnotesize 1, 0, 0, 0, 0};
\node[align=left, below] at (11.25,-2){\footnotesize 0, 0, 1, 0, 0};


\draw [decorate,decoration={brace,amplitude=5pt,raise=4pt},xshift=-4pt,yshift=0pt]
(12.5,-0.6) -- (12.5,-2.5) node [black,midway,xshift=0.6cm]{\footnotesize $d$};

\draw [decorate,decoration={brace,amplitude=5pt,mirror,raise=4pt},xshift=-4pt,yshift=0pt]
(3.2,-2.5) -- (12.4,-2.5) node [black,midway,yshift=-0.6cm]{\footnotesize $2d + k + 11$};

\draw [decorate,decoration={brace,amplitude=5pt,raise=4pt},xshift=-4pt,yshift=0pt]
(3.3,-0.6) -- (5.0,-0.6) node [black,midway,yshift=0.6cm]{\footnotesize $8$};

\draw [decorate,decoration={brace,amplitude=5pt,raise=4pt},xshift=-4pt,yshift=0pt]
(5.6,-0.6) -- (7.7,-0.6) node [black,midway,yshift=0.6cm]{\footnotesize $k+1$};

\draw [decorate,decoration={brace,amplitude=5pt,raise=4pt},xshift=-4pt,yshift=0pt]
(8.3,-0.6) -- (10,-0.6) node [black,midway,yshift=0.6cm]{\footnotesize $d+1$};

\draw [decorate,decoration={brace,amplitude=5pt,raise=4pt},xshift=-4pt,yshift=0pt]
(10.5,-0.6) -- (12.3,-0.6) node [black,midway,yshift=0.6cm]{\footnotesize $d+1$};

\end{tikzpicture}
\caption{The different representations of a state. In this example, a state $x_t$ contains $d=4$ NSC vectors. On the left, a network as a list of NSC vectors; on the right, the same network in its encoded representation. In our work, $k=5$ as observed in Table~\ref{tab:methodology:nas:ss:nsc}.}
\label{fig:methodology:rl:states:img}
\end{center}
\end{figure}
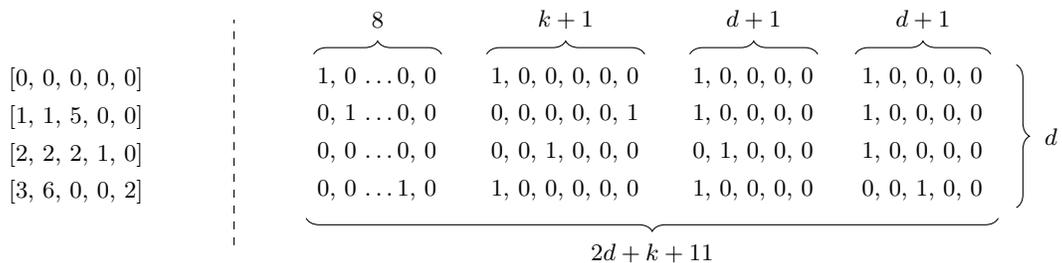

A state $x_i \in \mathcal{X}$ is a multidimensional array of size $d \times 5$, storing $d$ NSC vectors sorted by layer index.
While this representation is programmatically easy to control, it is not ideal in a machine learning setting. In particular, we note that every element of an NSC vector is a categorical variable. Therefore, when required, every NSC vector in $x_t$ is transformed as follows: 
the layer's type\footnote{This size is the result of having 7 types of layers (see Section~\ref{sec:methodology:nas:ss}) plus the type 0 representing an empty layer.} is encoded into a one-hot vector of size 8, the predecessors into a one-hot vector of size $(d+1)$, and the kernel size into a one-hot vector of size $(k+1)$ with $k=\max(\text{kernel\_size})$. The transformation ignores the layer index because the state implicitly incorporates the information of the position of each layer due to sorting. This encoding results in a multidimensional array\footnote{When working with chain-structured networks the second predecessor is always omitted, reducing the dimensionality of the encoding to $d \times (d + k + 10)$.} of size $d \times (2d + k + 11)$. Figure~\ref{fig:methodology:rl:states:img} illustrates this transformation.

\subsubsection{The action space}\label{sec:methodology:rl:as}

We formulate the action space $\mathcal{A}$ as a discrete space of 14 actions listed in Table~\ref{tab:methodology:rl:as}. Each action $a_i \in \mathcal{A}$ can either append a new element from the NSC space to a state $x_j \in \mathcal{X}$ or control two pointers, $p_1$ and $p_2$, for the indices of the predecessors to use for the next NSC vector. We note that for the chained-structured networks no pointers are required since the predecessor is always the previous layer, and neither do the merging operations \textit{addition} and \textit{concatenation}, making it possible to reduce the action space to 8 actions only.

\begin{table}[ht]
\centering
\begin{tabular}{@{}cl@{}}
\toprule
\textbf{Action ID} & \textbf{Description}                                                               \\ \midrule
A0                  & Add \textit{convolution} with $\text{kernel\_size}=1$, using predecessor $p_1$     \\
A1                  & Add \textit{convolution} with $\text{kernel\_size}=3$, using predecessor $p_1$     \\
A2                  & Add \textit{convolution} with $\text{kernel\_size}=5$, using predecessor $p_1$     \\
A3                  & Add \textit{max-pooling} with $\text{pool\_size}=2$, using predecessor $p_1$       \\
A4                  & Add \textit{max-pooling} with $\text{pool\_size}=3$, using predecessor $p_1$       \\
A5                  & Add \textit{avg-pooling} with $\text{pool\_size}=2$, using predecessor $p_1$       \\
A6                  & Add \textit{avg-pooling} with $\text{pool\_size}=3$, using predecessor $p_1$       \\
A7                  & Add \textit{terminal} state.                                                       \\
A8                  & Add \textit{addition} with predecessors $p_1$ and $p_2$                            \\
A9                  & Add \textit{concatenation} with predecessors $p_1$ and $p_2$                       \\
A10                 & Shift $p_1$ one position up (i.e., $p_1 = p_1 + 1$)                                                        \\
A11                 & Shift $p_1$ one position down (i.e., $p_1 = p_1 - 1$)                                                  \\
A12                 & Shift $p_2$ one position up (i.e., $p_2 = p_2 + 1$)                                                        \\
A13                 & Shift $p_2$ one position down (i.e., $p_2 = p_2 - 1$)                                                      \\ \bottomrule
\end{tabular}
\caption{The action space proposed, which is compliant with the NSC space of section~\ref{sec:methodology:nas:ss}. 
}
\label{tab:methodology:rl:as}
\end{table}

\subsubsection{The environments}\label{sec:methodology:rl:environments}

In our work, an environment is a neural architecture design task for image classification on a specific dataset of interest. The goal for an agent on this environment is to come up with the best architecture possible after interacting for a certain number of time-steps. At any time-step $t$, the environment's state is $x_t \in \mathcal{X}$, which is the NSC representation of a neural network $N_t$. The reward $r_t \in [0, 1]$ associated with $x_t$ is a function of the network's accuracy $\text{ACC}_{N_t} \in [0, 1]$ (Section~\ref{sec:methodology:nas:pss}). The initial state $x_0$ of the environment is an empty architecture. 

An agent can interact with the environment through a set of episodes by performing actions $a_t \in \mathcal{A}$. In our terminology, an \textit{episode} is the trajectory from a reset of the environment's state until a termination signal. The environment triggers the termination signal in the following cases: a) the predecessors $p_1$ and $p_2$ are out of bounds after the execution of $a_t$, b) $a_t$ is a \textit{terminal} action, c) $x_t$ contains $d$ NSC elements (the maximum depth) after performing $a_t$, d) the total number of actions executed in the current episode is higher than a given number $\tau$, or e) the action led to an invalid architecture. The agent-environment interaction process is formalized in Algorithm~\ref{alg:methodology:rl:environments:interaction}.

\begin{algorithm}
\caption{Agent-environment interaction}\label{euclid}
\begin{algorithmic}[1]
    \Procedure{interact}{Agent, Environment, Dataset, $t_{max}$, $\sigma$}
    \State $done \gets \text{False}$
    \State $t \gets 0$
    \State $\text{Environment.reset\_to\_initial\_state()}$
    \While {$t < t_{max}$}
        \State $a_t \gets \text{Agent.get\_next\_action()}$
        \State $x_t \gets \text{Environment.update\_state(}a_t\text{)}$
        \State $N \gets \text{Environment.build\_network(}x_t\text{)}$
        \State $\text{ACC}_{N_t} \gets \text{N.accuracy(Dataset)}$
        \If{$a_t$ is shifting}
            \State $r_t \gets \sigma \cdot \text{ACC}_{N_t}$ 
        \Else
            \State $r_t \gets \text{ACC}_{N_t}$ 
        \EndIf
        
        \State $done \gets \text{Environment.is\_termination()}$ \label{alg:methodology:rl:environments:interaction:termination}
        
        \State $\text{Agent.learn(}x_t, a_t, r_t, done\text{)}$
        
        \If{$done$}
            \State $ \text{Environment.reset\_to\_initial\_state()}$
            \State $done \gets \text{False}$
        \EndIf
    \EndWhile
    \EndProcedure
    
\end{algorithmic}
\label{alg:methodology:rl:environments:interaction}
\end{algorithm}

As mentioned in the beginning of Section~\ref{sec:methodology}, we work with more than one environment. Specifically, we define five environments, each one associated with a different dataset sampled from the meta-dataset collection~\citep{MetaDataset}. The datasets are listed in Table~\ref{tab:methodology:environments:datasets} and they were selected as explained in Appendix~\ref{app:datasets}. All datasets have balanced classes. In order to evaluate the accuracy of a network $N_t$, for any dataset we perform a deterministic 1/3 train-test split and follow the pre-processing that has been initially proposed by the meta-dataset authors so that the images are resized to a shape of $84 \times 84 \times 3$ using bilinear interpolation. 

\begin{table}[ht]
\centering
\begin{tabular}{@{}ccccc@{}}
\toprule
Dataset ID    & Dataset name                              & Usage      & N classes & N observations \\ \midrule
aircraft      & FGVC-Aircraft                             & Validation & 100       & 10000          \\
cu\_birds     & CUB-200-2011                              & Validation & 200       & 11788          \\
dtd           & Describable Textures                      & Train      & 47        & 5640           \\
omniglot      & Omniglot                                  & Train      & 1623      & 32460          \\
vgg\_flower   & VGG Flower                                & Train      & 102       & 8189           \\ \bottomrule
\end{tabular}
\caption{List of datasets considered for the environments. They are sampled from the meta-dataset~\citep{MetaDataset} as explained in Appendix \ref{app:datasets}.}.
\label{tab:methodology:environments:datasets}
\end{table}

\subsubsection{Deep meta-reinforcement learning}\label{sec:methodology:rl:dmrl}

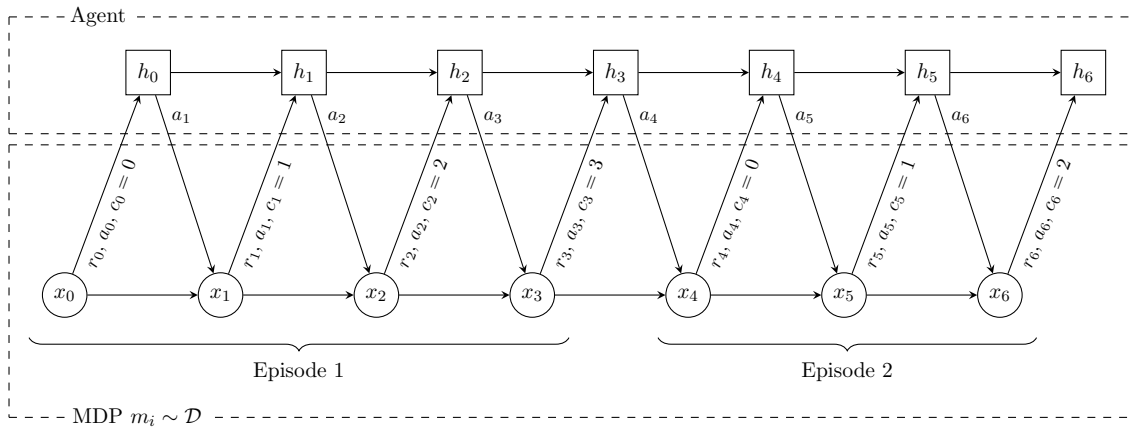
\begin{figure}[ht]
\begin{center}
\begin{tikzpicture}[scale=0.74, every node/.style={scale=0.74}]

\tikzstyle{hidden} = [rectangle, minimum width=0.8cm, minimum height=0.8cm, text centered, draw=black]

\tikzstyle{state} = [circle, radius=0.8, text centered, draw=black]

\tikzstyle{arrow} = [->,>=stealth]

\draw[dashed] (-2.5, 0) -- (-2.5, -2.1); 
\draw[dashed] (-2.5, 0) -- (-1.5, 0); 
\node at (-0.9, 0){Agent}; 
\draw[dashed] (-0.2, 0) -- (17.8, 0); 
\draw[dashed] (-2.5, -2.1) -- (17.8, -2.1); 
\draw[dashed] (17.8, 0) -- (17.8, -2.1); 

\draw[dashed] (-2.5, -2.3) -- (-2.5, -7.2); 
\draw[dashed] (-2.5, -2.3) -- (17.8, -2.3); 
\draw[dashed] (-2.5, -7.2) -- (-1.5, -7.2); 
\node at (-0.2, -7.2){ MDP $m_i \sim \mathcal{D} $}; 
\draw[dashed] (1.2, -7.2) -- (17.8, -7.2); 
\draw[dashed] (17.8, -2.3) -- (17.8, -7.2); 

\node (h0) [hidden] at (0, -1){ $h_0$};

\node (h1) [hidden, right of=h0, xshift=1.8cm]{ $h_1$};

\node (h2) [hidden, right of=h1, xshift=1.8cm]{ $h_2$};

\node (h3) [hidden, right of=h2, xshift=1.8cm]{ $h_3$};

\node (h4) [hidden, right of=h3, xshift=1.8cm]{ $h_4$};

\node (h5) [hidden, right of=h4, xshift=1.8cm]{ $h_5$};

\node (h6) [hidden, right of=h5, xshift=1.8cm]{ $h_6$};

\node (x1) [state, below of=h0, xshift=1.3cm, yshift=-3cm]{ $x_1$};

\node (x0) [state, left of=x1, xshift=-1.8cm]{ $x_0$};

\node (x2) [state, right of=x1, xshift=1.8cm]{ $x_2$};

\node (x3) [state, right of=x2, xshift=1.8cm]{ $x_3$};

\node (x4) [state, right of=x3, xshift=1.8cm]{$x_4$};

\node (x5) [state, right of=x4, xshift=1.8cm]{ $x_5$};

\node (x6) [state, right of=x5, xshift=1.8cm]{$x_6$};

\draw [arrow] (h0) -- (h1);
\draw [arrow] (h1) -- (h2);
\draw [arrow] (h2) -- (h3);
\draw [arrow] (h3) -- (h4);
\draw [arrow] (h4) -- (h5);
\draw [arrow] (h5) -- (h6);

\draw [arrow] (x0) -- (h0) node[near start,below, sloped, xshift=0.4cm] {\small $r_0$, $a_0$, $c_0=0$};

\draw [arrow] (x1) -- (h1) node[near start,below, sloped, xshift=0.4cm] {\small $r_1$, $a_1$, $c_1=1$};

\draw [arrow] (x2) -- (h2) node[near start,below, sloped, xshift=0.4cm] {\small $r_2$, $a_2$, $c_2=2$};

\draw [arrow] (x3) -- (h3) node[near start,below, sloped, xshift=0.4cm] {\small $r_3$, $a_3$, $c_3=3$};

\draw [arrow] (x4) -- (h4) node[near start,below, sloped, xshift=0.4cm] {\small $r_4$, $a_4$, $c_4=0$};

\draw [arrow] (x5) -- (h5) node[near start,below, sloped, xshift=0.4cm] {\small $r_5$, $a_5$, $c_5=1$};

\draw [arrow] (x6) -- (h6) node[near start,below, sloped, xshift=0.4cm] {\small $r_6$, $a_6$, $c_6=2$};

\draw [arrow] (h0) -- (x1) node[near start, right, yshift=0.4cm, xshift=-0.1cm] {\small $a_1$};

\draw [arrow] (h1) -- (x2) node[near start,right, yshift=0.4cm, xshift=-0.1cm] {\small $a_2$};

\draw [arrow] (h2) -- (x3) node[near start,right, yshift=0.4cm, xshift=-0.1cm] {\small $a_3$};

\draw [arrow] (h3) -- (x4) node[near start,right, yshift=0.4cm, xshift=-0.1cm] {\small $a_4$};

\draw [arrow] (h4) -- (x5) node[near start,right, yshift=0.4cm, xshift=-0.1cm] {\small $a_5$};

\draw [arrow] (h5) -- (x6) node[near start,right, yshift=0.4cm, xshift=-0.1cm] {\small $a_6$};

\draw [arrow] (x0) -- (x1);
\draw [arrow] (x1) -- (x2);
\draw [arrow] (x2) -- (x3);
\draw [arrow] (x3) -- (x4);
\draw [arrow] (x4) -- (x5);
\draw [arrow] (x5) -- (x6);

\draw [decorate,decoration={brace,amplitude=5pt,mirror,raise=4pt},xshift=-4pt,yshift=-5pt]
(-2,-5.4) -- (7.7,-5.4) node [black,midway,yshift=-0.8cm]{Episode 1};

\draw [decorate,decoration={brace,amplitude=5pt,mirror,raise=4pt},xshift=-4pt,yshift=-5pt]
(9.3,-5.4) -- (16.1,-5.4) node [black,midway,yshift=-0.8cm]{Episode 2};

\end{tikzpicture}
\caption{A graphic representation, inspired by the RL$^2$ illustration~\citep{RL2}, of our deep meta-reinforcement learning framework. In this example, the trial consists of $t_{max}=6$ time-steps, and the agent is able to complete two episodes of different length. $c_s$ is a counter of the current step in the episode and it gets reset at the start of any new episode. The states $x_0$ and $x_4$ are shown to be different, although in practice the initial state of an episode could always be the same.}
\label{fig:methodology:rl:formal:diag}
\end{center}
\end{figure}

Our deep meta-RL approach, illustrated in Figure~\ref{fig:methodology:rl:formal:diag}, is based on the work of~\citet{LtRL} and~\citet{RL2}. They propose to learn a policy that, in addition to the \textit{state-action} pairs of standard RL, uses the current time-step in the agent-environment interaction (the temporal information) as well as the previous action and reward. In this way, the agent can learn the relation between its past decisions and the current action. However, we introduce a modification in the temporal information, by considering the relative step within an episode instead of the global time-step so that the agent can capture the relation between changes in a neural architecture.

Formally, let $\mathcal{D}$ be a set of Markov Decision Processes (MDPs). Consider an agent embedding a Recurrent Neural Network (RNN) - with internal state $h$ - modeling a policy $\pi$. At the start of a \textit{trial}, a new task $m_i \in \mathcal{D}$ is sampled, and the internal state $h$ is set to zeros (empty network). The agent then executes its action-selection strategy for a certain number $t_{max}$ of discrete time-steps, performing $n$ episodes of maximum length $l$ depending on the environment's rules. At each step $t$ (with $0 \leq t \leq t_{max}$) an action $a_t \in A$ is executed as a function of the observed history $H_t = \{x_0, a_0, r_0, c_0, . . . , x_{t-1}, a_{t-1}, r_{t-1}, c_{t-1}, x_t\}$ (set of states $\{x_s\}_{0 \leq s \leq t}$, actions $\{a_s\}_{0 \leq s < t}$, rewards $\{r_s\}_{0 \leq s < t}$, episode-related steps $\{c_s\}_{0 \leq s \leq l}$) and a reward $r_t$ is obtained. At the very beginning of the trial, the action $a_0$ is sampled at random from a uniform distribution of all actions available, and the state $x_0$ is given by the environment's rules. The RNN's weights are trained to maximize the total discounted reward accumulated during each \textit{trial}. The evaluation consists of resetting $h$ and fixing $\pi$ to run an interaction with a new MDP $m_e \not\in \mathcal{D}$.

\subsubsection{The policy optimization algorithm}\label{sec:methodology:rl:poa}

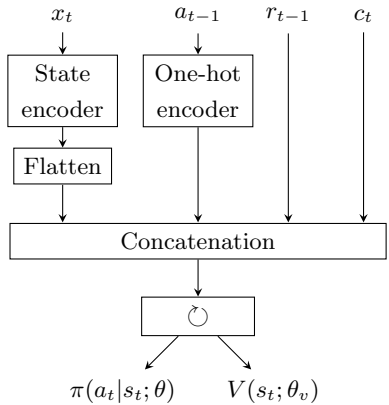
\begin{figure}[ht]
\begin{center}
\begin{tikzpicture}

\tikzstyle{point} = []
\tikzstyle{box} = [rectangle, text centered, draw=black]

\tikzstyle{arrow} = [->,>=stealth]

\node (x) [point] at (0, 0){\footnotesize$x_t$};

\node (enc) [box, below of=x, text width=1.2cm]{\footnotesize State encoder};

\node (flat) [box, below of=enc]{\footnotesize Flatten};

\node (oenc) [box, right of=enc, text width=1.2cm, xshift=0.8cm]{\footnotesize One-hot encoder};

\node (a) [point, above of=oenc]{\footnotesize$a_{t-1}$};

\node (r) [point, right of=a, xshift=0.2cm]{\footnotesize$r_{t-1}$};
\node (c) [point, right of=r]{\footnotesize$c_{t}$};

\node (con) [box, below of=flat, xshift=1.8cm, minimum width=5cm]{\footnotesize Concatenation};

\node (lstm) [box, below of=con, minimum width=1.5cm]{ $\circlearrowright$};

\node (pi) [point, below of=lstm, xshift=-1cm]{\footnotesize $\pi(a_t | s_t; \theta)$};

\node (v) [point, below of=lstm, xshift=1cm]{\footnotesize $V(s_t; \theta_v)$};

\draw [arrow] (x) -- (enc);

\draw [arrow] ($(flat)+(0, -0.25)$) -- ($(con)+(-1.8, 0.25)$);
\draw [arrow] (a) -- (oenc);
\draw [arrow] (oenc) -- ($(con)+(0, 0.25)$);

\draw [arrow] (r) -- ($(con)+(1.2, 0.25)$);
\draw [arrow] (c) -- ($(con)+(2.2, 0.25)$);
\draw [arrow] (enc) -- (flat);

\draw [arrow] (con) -- (lstm);

\draw [arrow] (lstm) -- (pi);
\draw [arrow] (lstm) -- (v);

\end{tikzpicture}
\caption{Illustration of the \textit{meta}-A2C architecture. In our implementation, the ``State encoder" follows the procedure explained in  Section~\ref{sec:methodology:rl:states}, and the recurrent layer is an LSTM with 128 units.}
\label{fig:related:rl:alg:lstm}
\end{center}
\end{figure}

Similarly to~\citet{LtRL}, we make use of the synchronous Advantage Actor-Critic (A2C)~\citep{A2C} with one worker. As it can be observed in Figure~\ref{fig:related:rl:alg:lstm}, the only change in the A2C network is in the input of the recurrent unit, so that the updates of the network's parameters remain unchanged.

Formally, let $t$ be the current time step, $s_t=x_t \cdot a_{t-1} \cdot r_{t-1} \cdot c_t$ a concatenation of inputs, $\pi(a_t|s_t; \theta)$ the policy, $V(s_t; \theta_v)$ the value function, $H$ the entropy, $j \in \mathbb{N}$ the horizon, $\gamma \in (0, 1]$ the discount factor, $\eta$ the regularization coefficient, and $R_t = \sum_{i=0}^{j-1} \gamma^i r_{t+i}$ the total accumulated return from time step $t$. The gradient of the objective function is:
\begin{equation}
    \nabla_{\theta} \log \pi(a_t|s_t;\theta) \underbrace{(R_t - V(s_t; \theta_v))}_\textrm{Advantage estimate} + \underbrace{\eta \nabla_{\theta}H(\pi(s_t; \theta))}_\textrm{Entropy regularization}
\end{equation}\label{eq:methodology:rl:poa:gradient}

As it is usually the case for A2C, the parameters $\theta$ and $\theta_v$ are shared except for the ones in output layers. For a detailed description of the algorithm, we refer to the original paper~\citep{A2C}.

%% file: software.tex
\section{Evaluation framework}\label{sec:software}

The current Neural Architecture Search (NAS) solutions lack a crucial element: an open-source framework for reproducibility and further research. Specifically for NAS with reinforcement learning, it would be desirable to build on a programming interface that allows researchers to explore the effect of different algorithms on the same NAS environment. In an attempt to fill this gap, we have developed the \textsc{nasgym}\footnote{Source code available at: github.com/gomerudo/nas-env}, a python OpenAI Gym~\citep{openaigym} environment that can jointly be used with all the reinforcement learning algorithms exposed in the OpenAI baselines~\citep{openaibaselines}. 

\begin{figure}[ht]
\begin{center}
\begin{tikzpicture}

\tikzstyle{boxA} = [rectangle, dashed, minimum width=2cm, minimum height=1cm, text centered, draw=black, fill=gray!0]
\tikzstyle{boxempty} = [rectangle, minimum height=1cm, text width=3cm, text centered, fill=gray!0]

\tikzstyle{arrow} = [->,>=stealth]

\draw (-4, 0) -- (-4, -4); 
\draw (-4, 0) -- (-3, 0); 
\node at (-2.2, 0){\textsc{nasgym}}; 
\draw (-1.35, 0) -- (5, 0); 
\draw (-4, -4) -- (5, -4); 
\draw (5, 0) -- (5, -4); 

\node (nasenv) [boxA] at (0.75, -1) {\small \textsc{DefaultNasEnvironment}};

\node (datahandler) [boxA, below of=nasenv, yshift=-1cm, xshift=-2cm] {\small \textsc{DatasetHandler}};

\node (dbexperiments) [boxA, below of=nasenv, yshift=-1cm, xshift=2cm] {\small \textsc{DbInterface}};

\node (nscdef) [boxempty, left of=nasenv, xshift=-6cm] {\faFileCodeO \\ \small NSC \\definition file};

\node (config) [boxempty, below of=nscdef, yshift=-1cm] {\faFileCodeO \\ \small Hyper-parameters values};

\node (tfrecords) [boxempty, below of=datahandler, yshift=-1cm] {\faFilesO \\ \small TFRecords files};

\node (db) [boxempty, below of=dbexperiments, yshift=-1.2cm] {\faDatabase \\ \small Database of \\ experiments};

\draw [arrow] (nscdef) -- (nasenv);
\draw [arrow] (config) -- (nasenv.west);
\draw [arrow] (tfrecords) -- (datahandler);
\draw [arrow] (db) -- (dbexperiments);

\end{tikzpicture}
\caption{An sketch of the system built to perform our research. The \textsc{nasgym} package contains a default NAS environment whose states and actions are designed according to the Neural Structure Code (NCS) space defined in a \textit{.yml} file. The hyperparameters for all the machine learning components are defined in a \textit{.ini} file. Internally, the environment makes use of a dataset handler that reads TFRecords files and sends them as inputs to the neural architectures. A simple database of experiments is used to store experiments in a local file, although the logic can be easily be extended to support a more robust database system.}
\label{fig:software:architecture}
\end{center}
\end{figure}
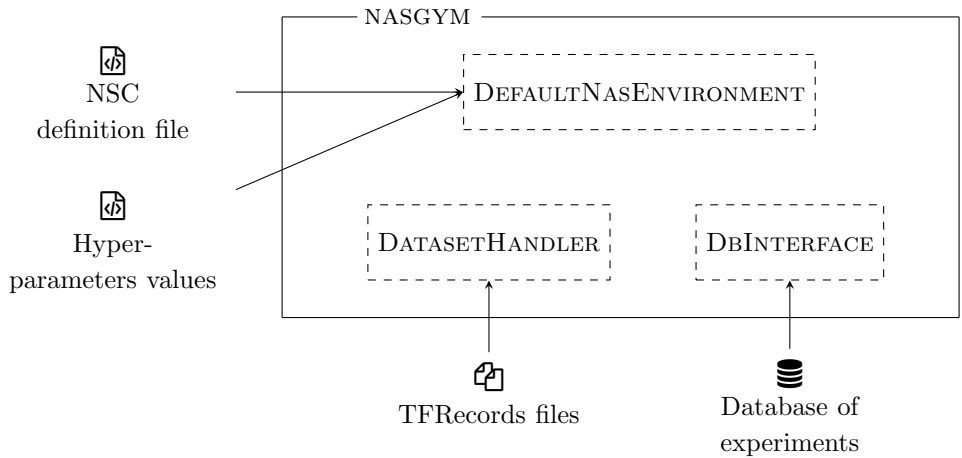

We make use of the object-oriented paradigm to abstract the most essential elements of NAS as a reinforcement learning problem, resulting in a system that can be extended to perform new experiments, as displayed in Figure~\ref{fig:software:architecture}. Although the defaults in the \textsc{nasgym} are the elements in our methodology, the system allows us to easily modify the key components, such as the performance estimation strategy, the action space, or the Neural Structure Code space. We also provide an interface to use a database of experiments that can help to store previously computed rewards, thus reducing the computation time of future trials. All the deep learning components are built with TensorFlow v1.12~\citep{tensorflow}.

Additionally to the \textsc{nasgym}, we implement the meta version of the A2C algorithm on top of the OpenAI baselines\footnote{Source code available at: github.com/gomerudo/openai-baselines}. We believe that this software engineering effort will help to compare, reproduce, and develop future research in NAS.

%% file: experiments.tex
\section{Experiments}\label{sec:experiments}

To evaluate our framework, we conduct three experiments. The first two aim to study the behavior of the agent when challenged to design chain-structured networks, and the third one is intended to observe its behavior in the multi-branch setting. We empirically assess the quality of the networks designed by the agent through episodes, the ability of the agent to adapt to each environment, and the runtimes of the training trials.

\subsection{Chain-structured networks} \label{sec:experiments:chain}

\textbf{Experiment 1: evolution during training.} The agent learns from the three train environments listed in Table~\ref{tab:methodology:environments:datasets}, using deep meta-RL. It starts in the \textit{omniglot} environment, continues in \textit{vgg\_flower}, and finishes in \textit{dtd} so that it faces increasingly harder classification tasks (see Appendix~\ref{app:datasets}), and the policy learned in one environment is reused in the next one. The agent interacts with each environment for $t_{max}=8000$, $t_{max}=7000$, and $t_{max}=7000$, respectively so that the agent spends more time in the first environment to develop its initial knowledge.
We compare against two baselines: random search and \textsc{DeepQN} with experience replay, where the agent learns a new policy on each environment (i.e., it does not re-use the policy between trials) for $t_{max}=6500, 5500, \text{ and }7000$, respectively. Due to resources and time constraints, all $t_{max}$ values were empirically selected according to the behaviour of the rewards (see Section~\ref{sec:results}). The most relevant hyper-paremeters are set as follows:

\begin{itemize}
    \setlength\itemsep{0em}
    \setlength\parskip{0pt}
    \item[-] \textsc{Environment}
        \begin{itemize}
            \setlength\itemsep{0em}
            \item[-]  $d=10$. The maximum depth of a neural architecture.
            \item[-] $\tau=10$. The maximum length of an episode.
        \end{itemize}
    \item[-] \textsc{A2C hyper-parameters}
    \begin{itemize}
        \setlength\itemsep{0em}
        \setlength\parskip{0pt}
        \item[-] $j=5$. The number of steps to perform before updating the A2C parameters (see Equation~\ref{eq:methodology:rl:poa:gradient}). We set the value to half the maximum depth of the networks to allow the agent to learn before the termination of an episode.
        \item[-] $\gamma=0.9$. The discount factor for the past actions.
        \item[-] $\eta=0.01$. The default in the OpenAI baselines~\citep{openaibaselines}.
        \item[-] $\alpha=0.001$. The A2C learning rate set as in \textit{Learning to reinforcement learn}~\citep{LtRL}.
    \end{itemize}
    \item[-] \textsc{DeepQN}
    \begin{itemize}
        \setlength\itemsep{0em}
        \setlength\parskip{0pt}
        \item[-] Experience $\text{buffer size}=\frac{t_{max}}{2}$. 
        \item[-] Target model's $\text{batch size} = 20$.
        \item[-] $\epsilon$ with linear decay from 1.0 to 0.1. The parameter controlling the exploration of the agent.
        \item[-] $\alpha=0.0005$. The default learning rate in the OpenAI baselines~\citep{openaibaselines}.
    \end{itemize}
    \item[-] \textsc{Training of the sampled networks}
    \begin{itemize}
        \setlength\itemsep{0em}
        \setlength\parskip{0pt}
        \item[-] $\text{batch size}=128$.
        \item[-] $\text{epochs}=12$. The value used in BlockQNN~\citep{BlockQNN}.
    \end{itemize}
\end{itemize}

\textbf{Experiment 2: evaluation of the policy.} We fix the policy obtained in Experiment 1. The agent interacts with the evaluation environments, \textit{aircraft} and \textit{cu\_birds}, and deploys its decision-making strategy to design a neural architecture for each dataset. The interaction runs for $t_{max}=2000$ to study the performance of the policy in short evaluation trials. At the end of the interaction, we select the best two architectures per environment (i.e., the ones with the highest reward) and train them on the same datasets but applying a more intensive procedure as follows. First, we augment the capacity of the architectures by changing the number of filters in the convolution layers according to the layer's depth; i.e., $\text{number of units}=2^{4+i}$ with $i$ being the current count of convolutions while building the network  (e.g., $\text{number of units}=32 \rightarrow 64 \rightarrow 128$). Second, we stack the prediction module as described in Section~\ref{sec:methodology:nas:pss}, but we increase the number of units in the first dense layer to 4096, we use a learning rate with exponential decay, and we train the network for 100 epochs. Since the datasets that we use are resized to a shape of 84x84x3, it is not fair to compare our resulting accuracy values with those of state-of-the-art architectures that assume a higher order of shape~\citep{FineGrained2, FineGrained3, FineGrainedResults}, and neither is to train our networks (which are designed for a given input size) with bigger images. Hence, based on the baselines of~\citet{FineGrainedResults}, we use a VGG-19 network~\citep{VGGPaper} with only two blocks as our baseline on both datasets.

\subsection{Multi-branch networks} \label{sec:experiments:multibranch}

\textbf{Experiment 3: training on a more complex environment.} In this experiment, we extend the search space to multi-branch architectures. We consider the \textit{omniglot} environment only. The goal here is to observe the ability of the agent to design multi-branch networks through time; i.e., the number of multi-branch structures generated through training. The interaction runs for $t_{max}=20000$ time-steps because more exploration is required due to the larger action space. The hyper-parameters are the same as in Experiment 1, except that $\tau=20$ and $j=10$ because the trajectories are longer due to the shifting of the pointers controlling the predecessors, and $\text{batch size}=64$ because the concatenation operation can generate networks that require more space in memory. We train the agent from scratch two times varying the parameter $\sigma \in [0.0, 0.1]$ (see Section~\ref{sec:methodology:rl:environments}) to study its effect encouraging the generation of multi-branch structures.

%% file: results.tex
\section{Results}\label{sec:results}

\subsection*{Experiment 1: evolution during training}

\begin{figure}[ht]
\centering
\begin{subfigure}{.33\textwidth}
  \centering
      \includegraphics[width=\linewidth]{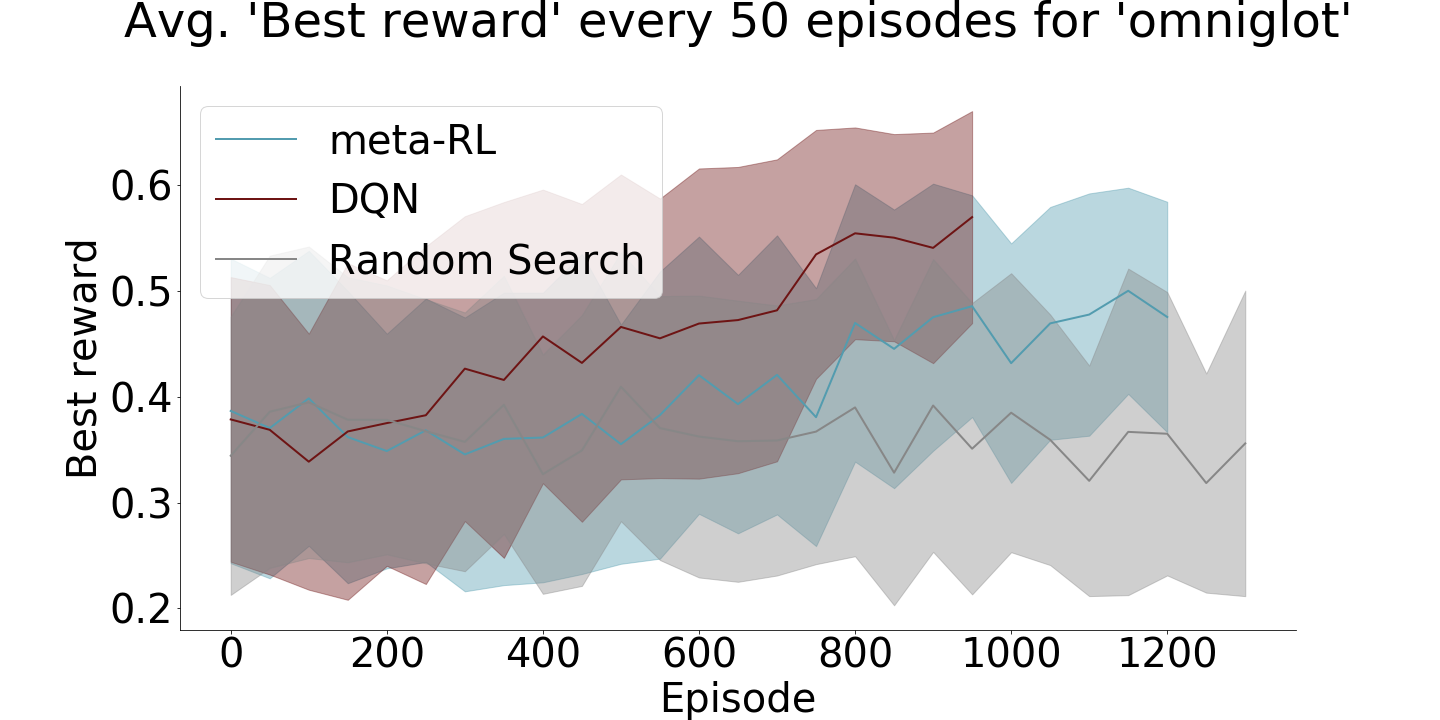}
  \caption{}
  \label{fig:results:exp1:evolution:a}
\end{subfigure}%
\begin{subfigure}{.33\textwidth}
  \centering
      \includegraphics[width=\linewidth]{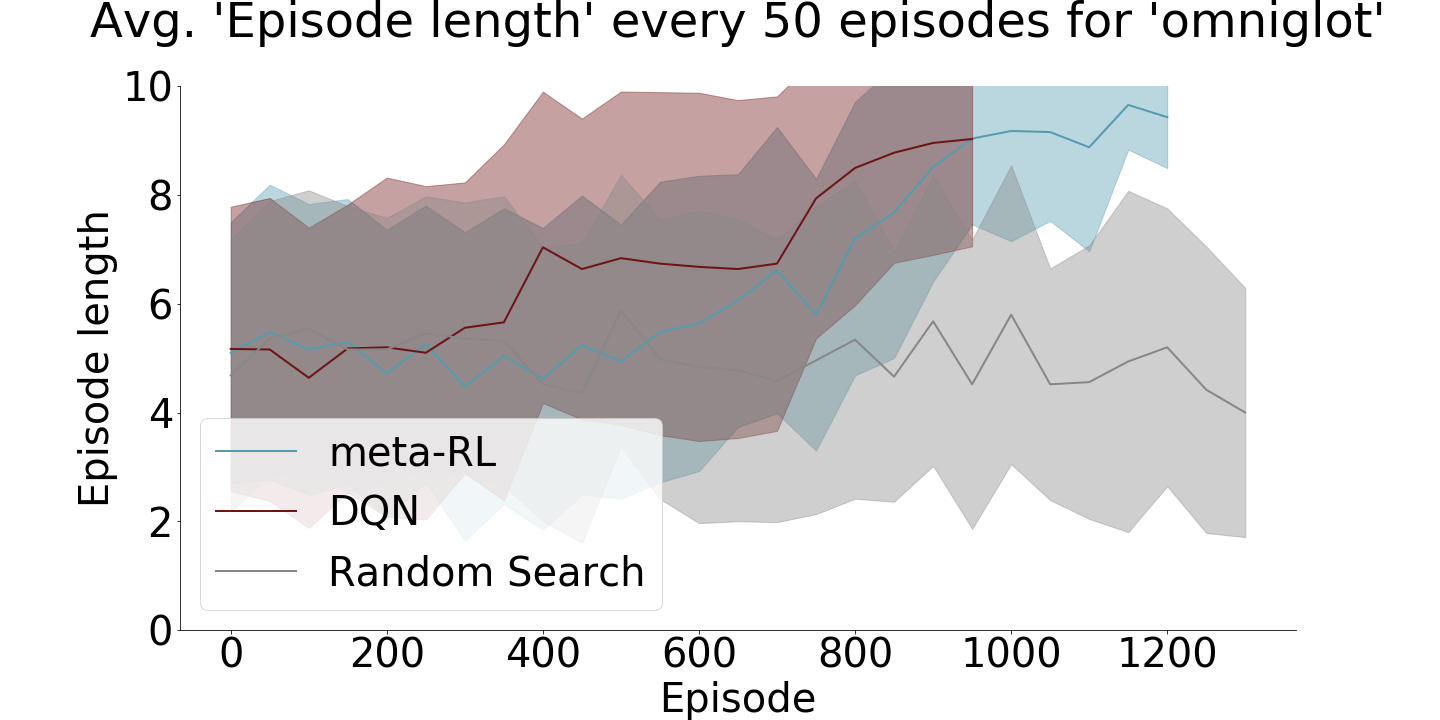}
  \caption{}
  \label{fig:results:exp1:evolution:b}
\end{subfigure}%
\begin{subfigure}{.33\textwidth}
  \centering
      \includegraphics[width=\linewidth]{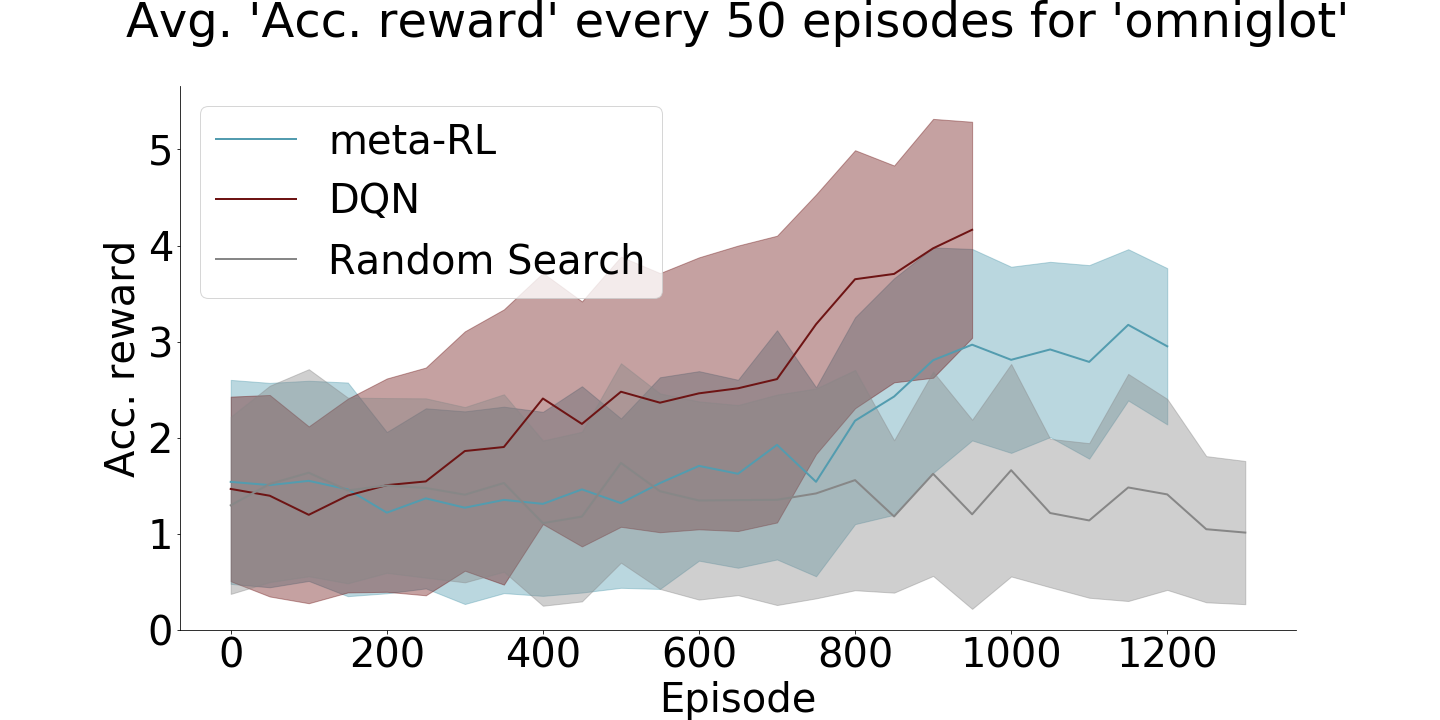}
  \caption{}
\label{fig:results:exp1:evolution:c}
\end{subfigure}
\begin{subfigure}{.33\textwidth}
  \centering
      \includegraphics[width=\linewidth]{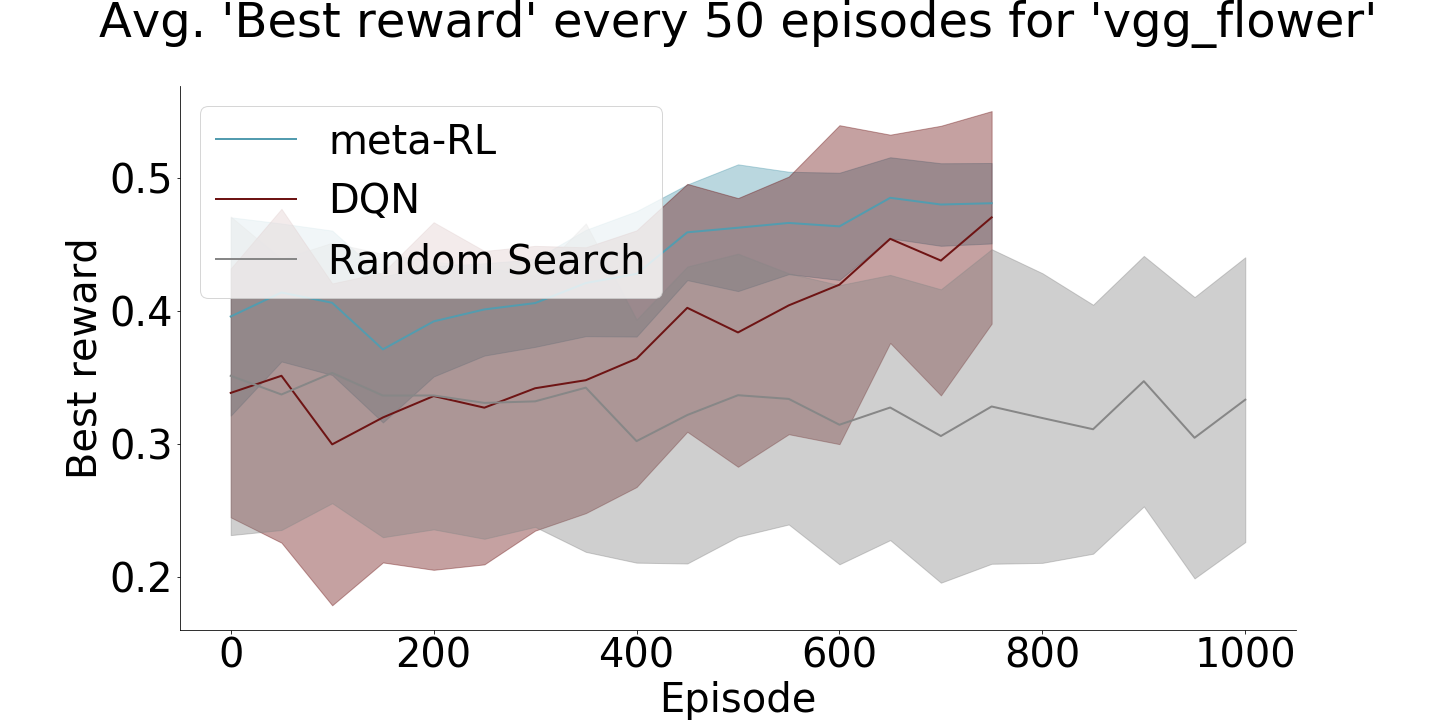}
  \caption{} 
\label{fig:results:exp1:evolution:d}
\end{subfigure}%
\begin{subfigure}{.33\textwidth}
  \centering
      \includegraphics[width=\linewidth]{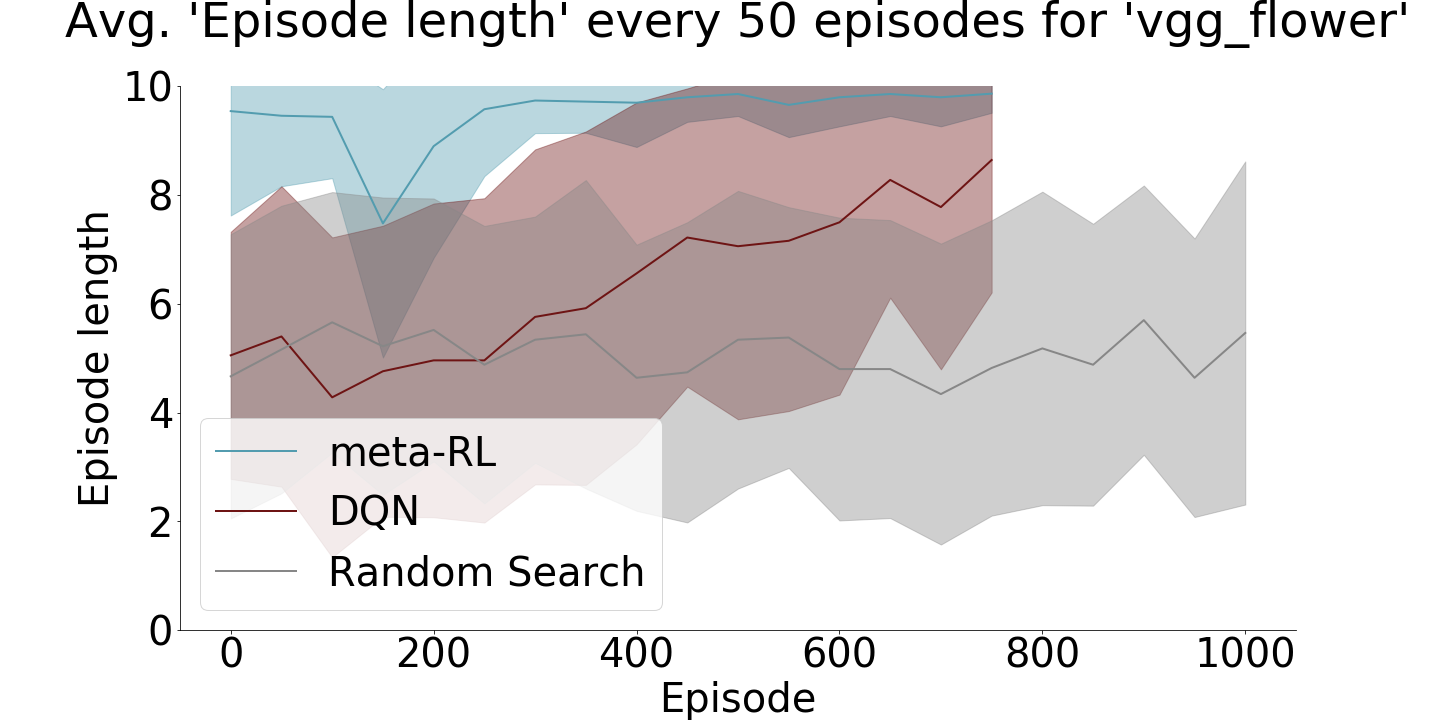}
  \caption{}
\label{fig:results:exp1:evolution:e}
\end{subfigure}%
\begin{subfigure}{.33\textwidth}
  \centering
      \includegraphics[width=\linewidth]{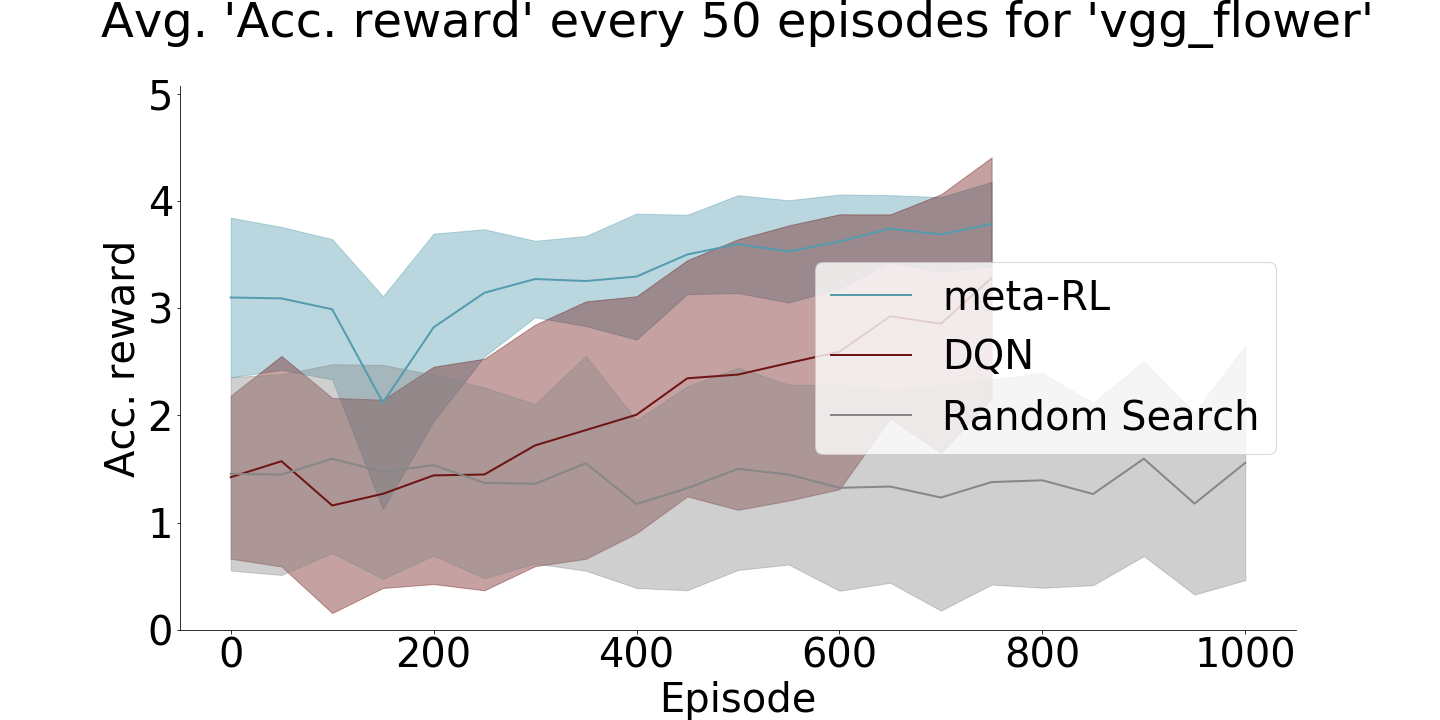}
  \caption{} 
\label{fig:results:exp1:evolution:f}
\end{subfigure}
\begin{subfigure}{.33\textwidth}
  \centering
      \includegraphics[width=\linewidth]{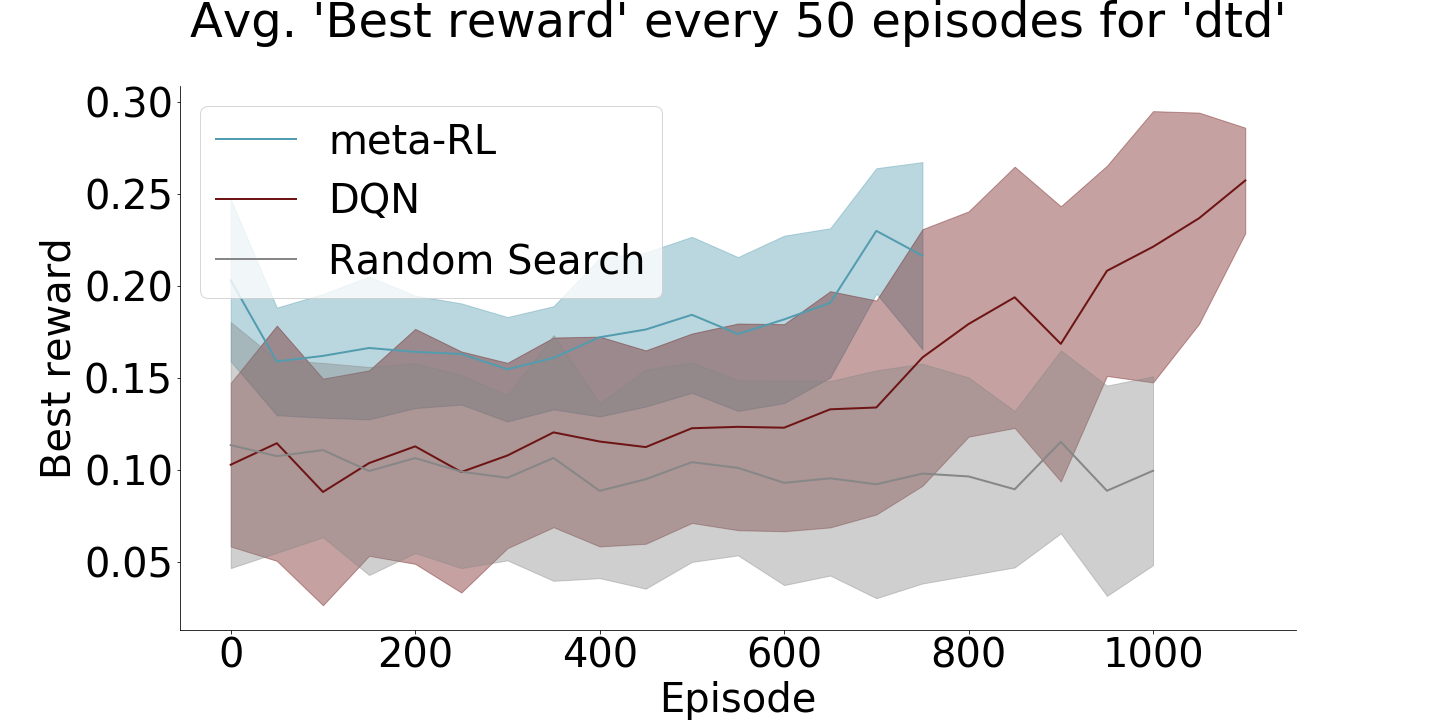}
  \caption{}
\label{fig:results:exp1:evolution:g}
\end{subfigure}%
\begin{subfigure}{.33\textwidth}
  \centering
      \includegraphics[width=\linewidth]{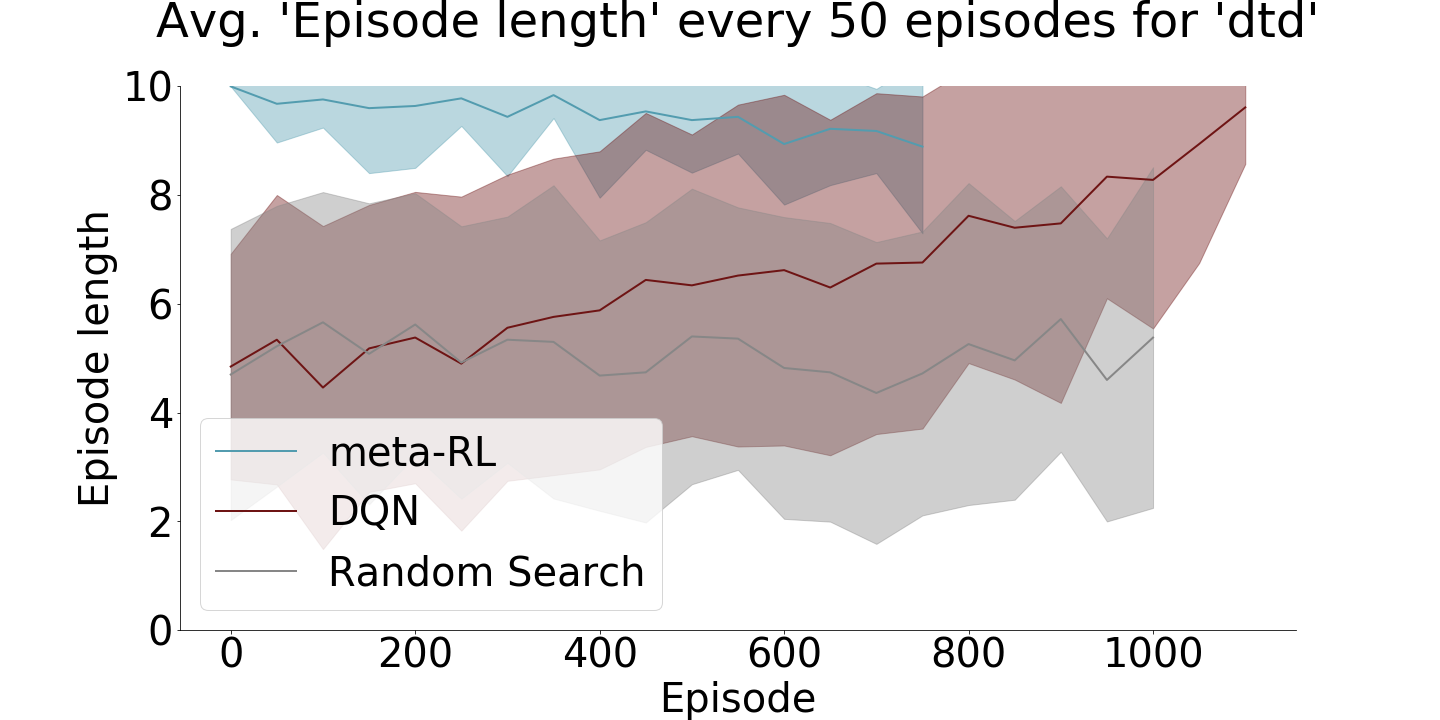}
  \caption{} 
\label{fig:results:exp1:evolution:h}
\end{subfigure}%
\begin{subfigure}{.33\textwidth}
  \centering
      \includegraphics[width=\linewidth]{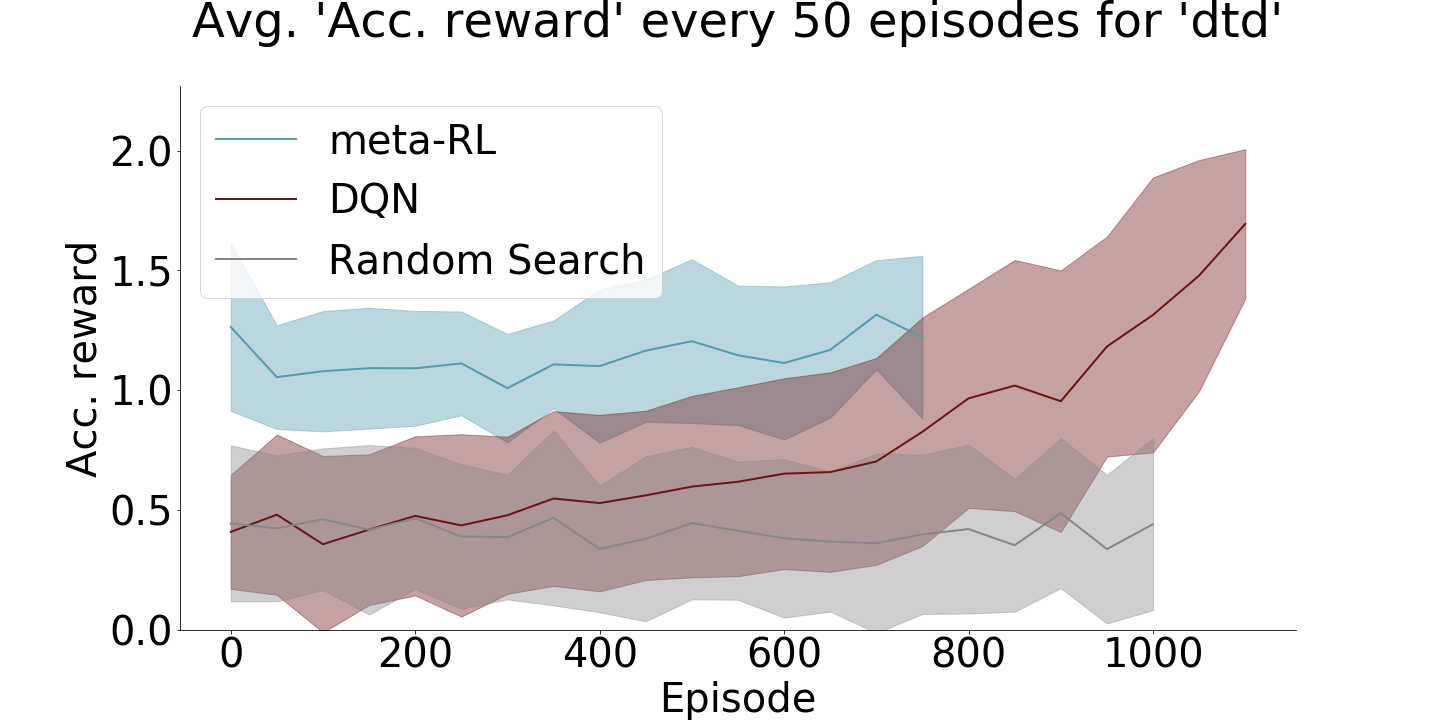}
  \caption{} 
\label{fig:results:exp1:evolution:i}
\end{subfigure}
\caption{Evolution of training episodes through time from different perspectives, showing the means and $\pm 1$ standard deviations for every 50 episodes. Since the different techniques can build networks of different depths per episode, the number of episodes executed per environment may differ.}
\label{fig:results:exp1:evolution}
\vspace{-0.5cm}
\end{figure}

Figure~\ref{fig:results:exp1:evolution} shows the evolution of the \textit{best reward} and the \textit{accumulated reward} per episode (representing the quality of the neural architectures), as well as the \textit{episode length} (in a chain-structured network this represents the number of layers). We observe that, in the first environment (\textit{omniglot}), our deep meta-RL agent performs worse than \textsc{DeepQN}. Nevertheless, in the second and third environments (\textit{vgg\_flower}, \textit{dtd}), the agent performs better than the baselines from the very first steps, and more consistently through all episodes (showing less variance). \textsc{DeepQN} only catches up after many more episodes, although it exhibits a faster learning curve, which we ascribe to the linear exploration that makes it to end exploration sooner.

\begin{figure}[ht]
\centering
    \includegraphics[width=0.45\linewidth]{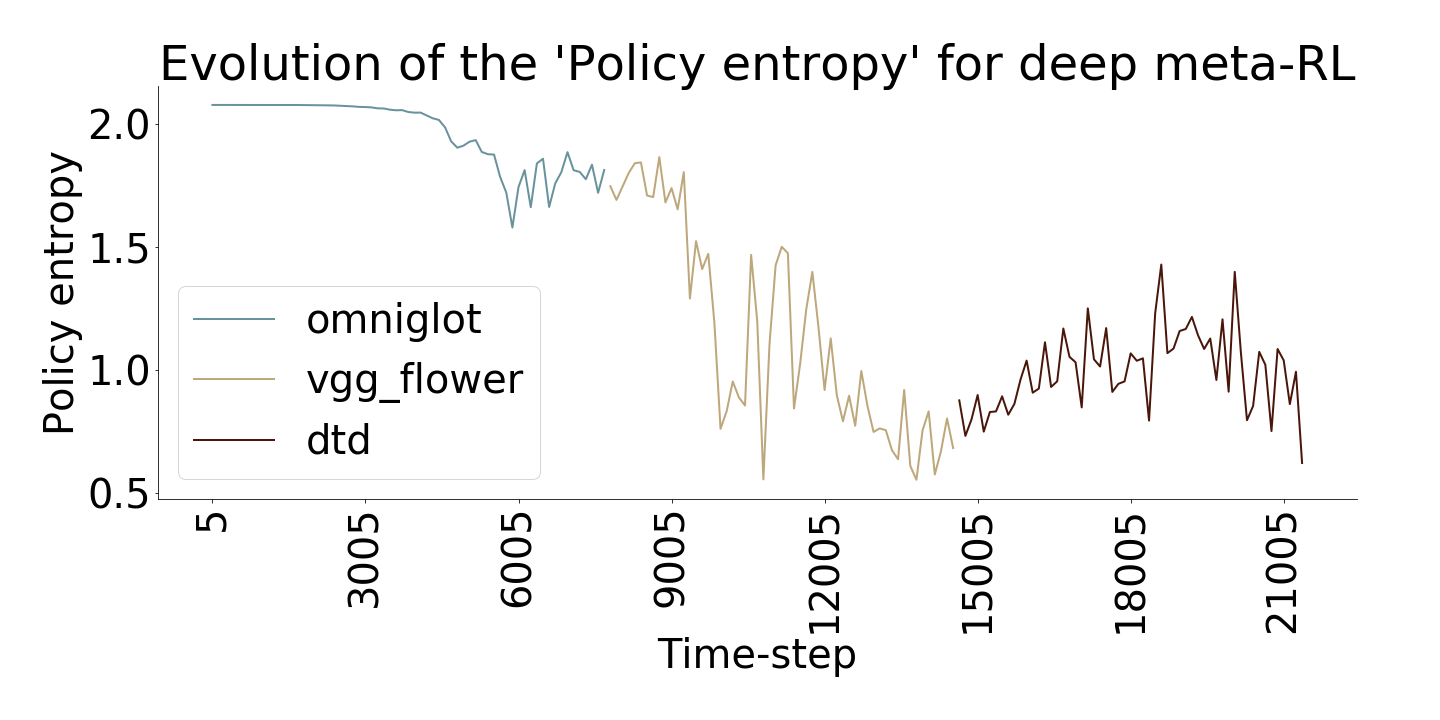}
\caption{Policy entropy through environments in Experiment 1.}
\label{fig:results:entropy}
\end{figure}

Figure~\ref{fig:results:entropy} shows the entropy of the policy during training over the different environments, which in the A2C algorithm is related to the level of exploration by the agent (more exploration leads to high entropy). In the first environment, our agent explores the environment for a significant number of time-steps, which translates to the slow increase observed in Figure~\ref{fig:results:exp1:evolution:a}. In the second environment, the exploration drops down quickly, except for a short period with increased exploration (time-steps 9005 to 12005). In the last and hardest environment, the agent re-explores the environment to adapt its strategy, leading to a reduction of the episode length (depth of the networks) and, consequently, the accumulated reward does not appear to increase due to the shorter episodes. We believe that exploration causes the drops in episode length in \textit{vgg\_flower} and \textit{dtd} (Figures~\ref{fig:results:exp1:evolution:e} and~\ref{fig:results:exp1:evolution:h}).

\begin{figure}[ht]
\centering
    \includegraphics[width=0.55\linewidth]{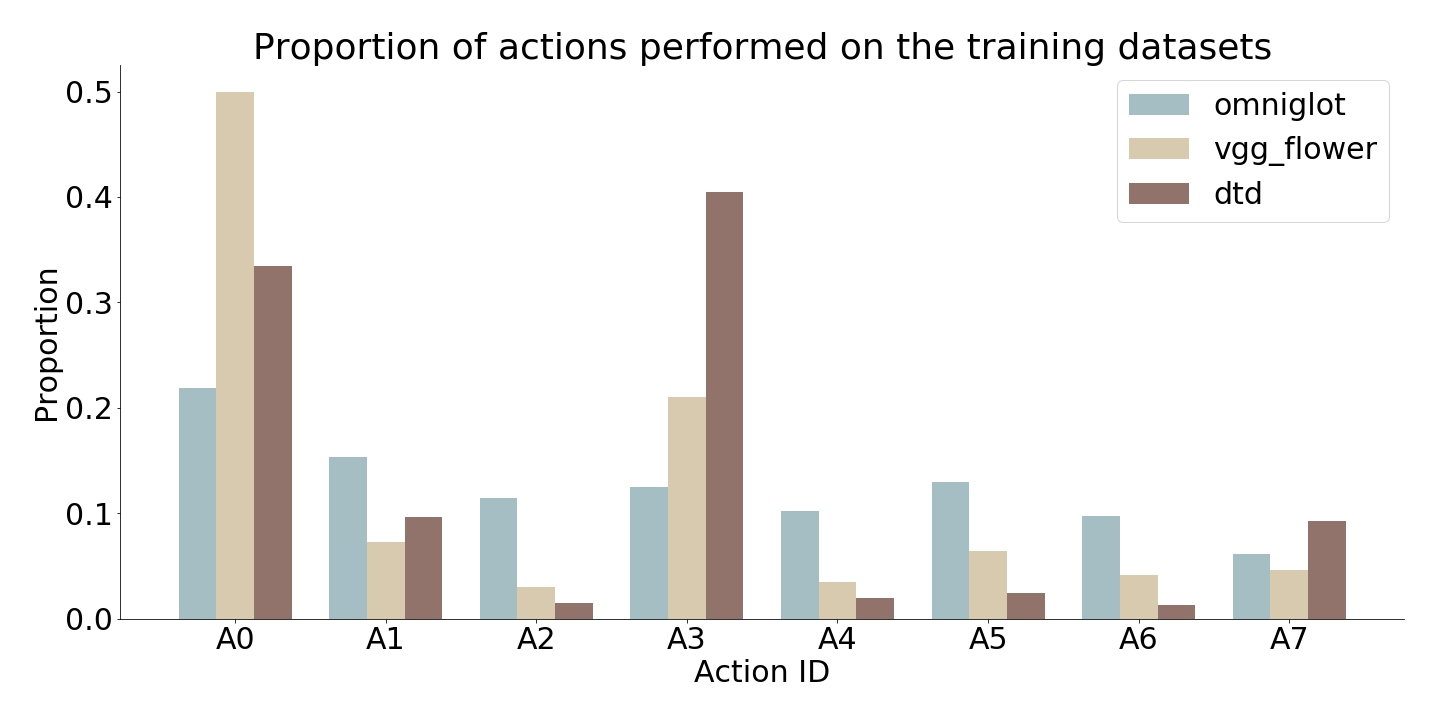}
\caption{Proportion of actions performed by the agent per dataset in Experiment 1. The labels in the x-axis match the IDs in Table~\ref{tab:methodology:rl:as}.}
\label{fig:results:exp1:actions}
\end{figure}

In Figure~\ref{fig:results:exp1:actions}, the proportion of the actions performed by the agent during training is shown. We see that it deployed different strategies per environment. Specifically, we note the changes in proportion for actions \textsc{A0} (\textit{convolution} with a kernel of size 1),  \textsc{A3} (\textit{max-pooling} with pool size of 2), and \textsc{A7} (the \textit{terminal} state) when the environment switched from \textit{vgg\_flower} to \textit{dtd}, suggesting that the agent preferred different layers and depth according to the dataset.

Finally, Table~\ref{tab:results:times} shows the running times per environment for each RL algorithm. Here, we do not observe significant differences considering that once transferred, the policy of the deep meta-RL agent designs deeper and more costly networks, as observed in Figures~\ref{fig:results:exp1:evolution:e} and~\ref{fig:results:exp1:evolution:h}.

\begin{table}[ht]
\centering
\begin{tabular}{@{}ccc@{}}
\toprule
Dataset     & Deep meta-RL & DQN         \\ \midrule
omniglot    & 11 days 9h   & 6 days 14h    \\
vgg\_flower & 7 days 23h   & 5 days 15h     \\
dtd         & 6 days 17h   & 6 days 4h      \\ \midrule
Total       & 26 days 1h   & 18 days 9h    \\ \bottomrule
\end{tabular}
\caption{Running times per dataset during training. All experiments ran on a single NVIDIA Tesla K40m GPU.}
\label{tab:results:times}
\end{table}


\subsection*{Experiment 2: evaluation of the policy}

\begin{figure}[ht]
\centering
\begin{subfigure}{.33\textwidth}
  \centering
      \includegraphics[width=\linewidth]{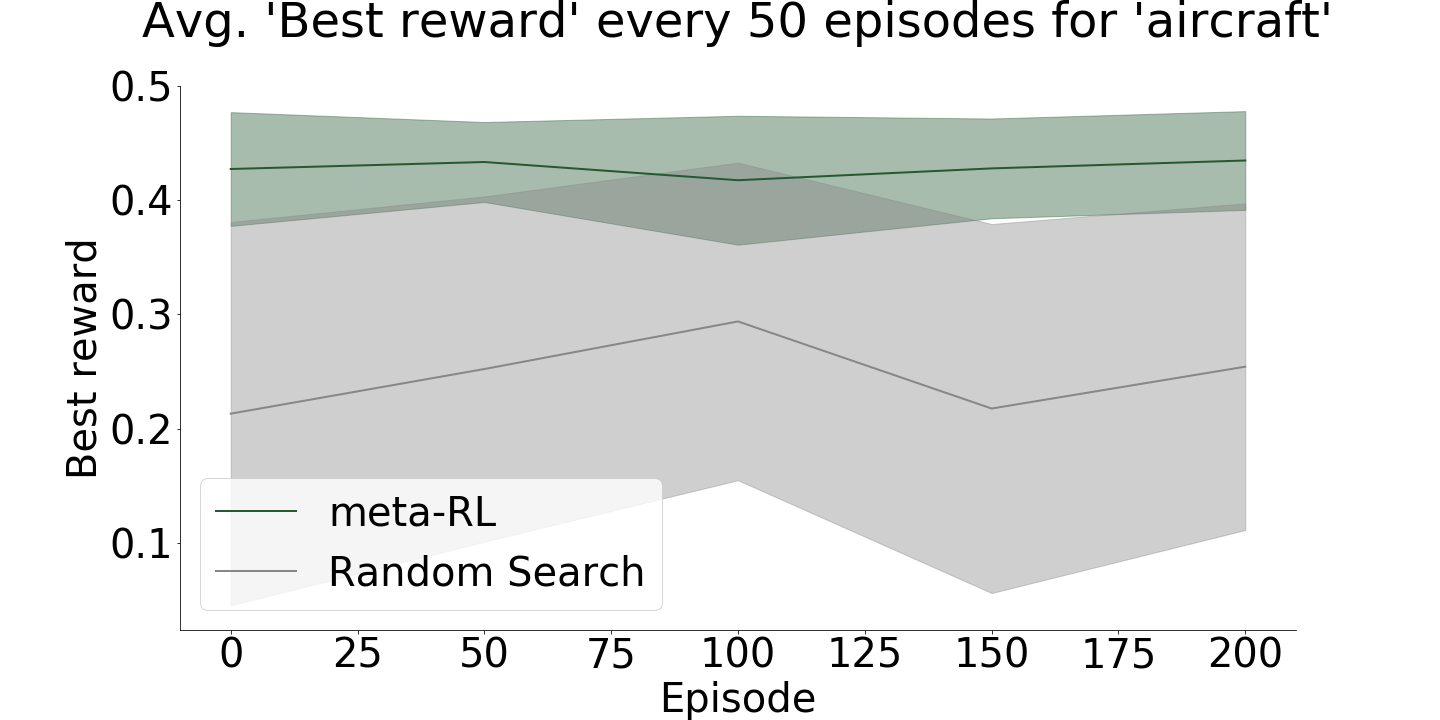}
  \caption{}
  \label{fig:results:exp2:evolution:a}
\end{subfigure}%
\begin{subfigure}{.33\textwidth}
  \centering
      \includegraphics[width=\linewidth]{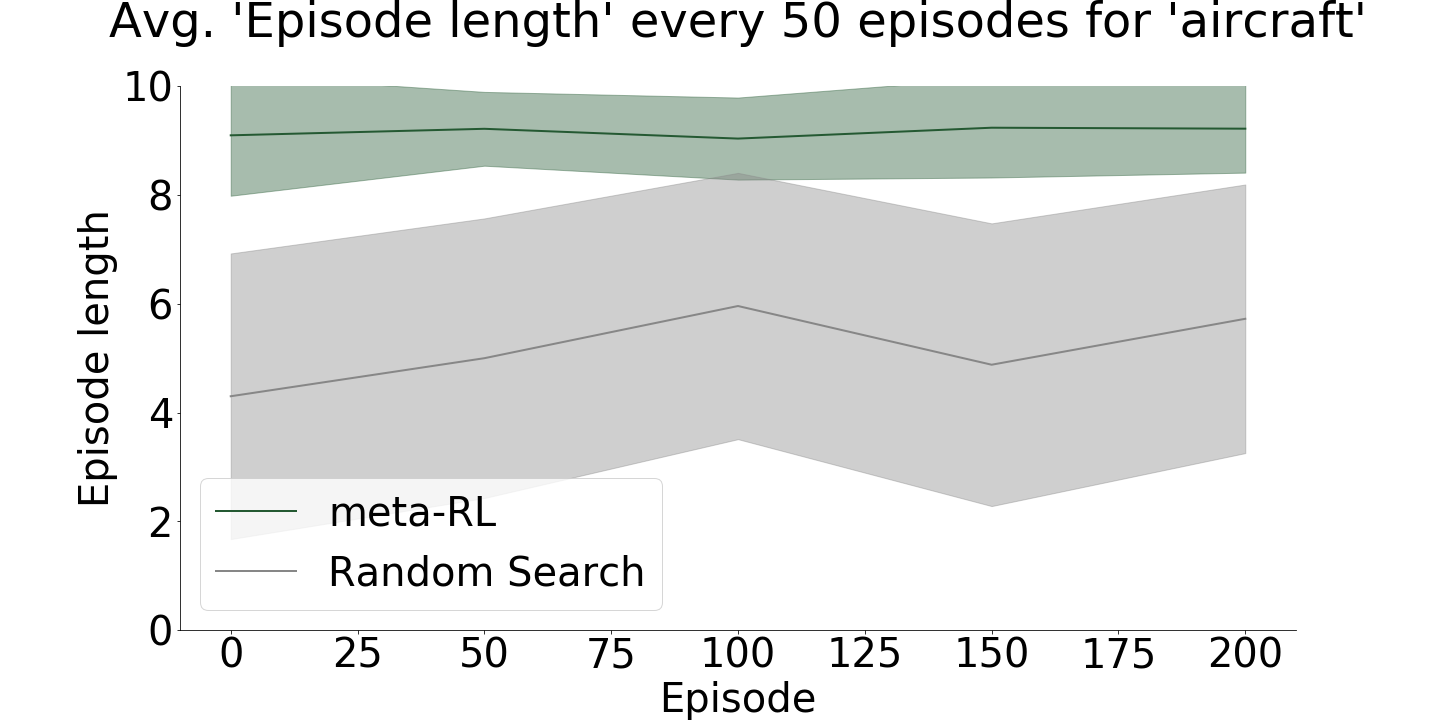}
  \caption{}
  \label{fig:results:exp2:evolution:b}
\end{subfigure}%
\begin{subfigure}{.33\textwidth}
  \centering
      \includegraphics[width=\linewidth]{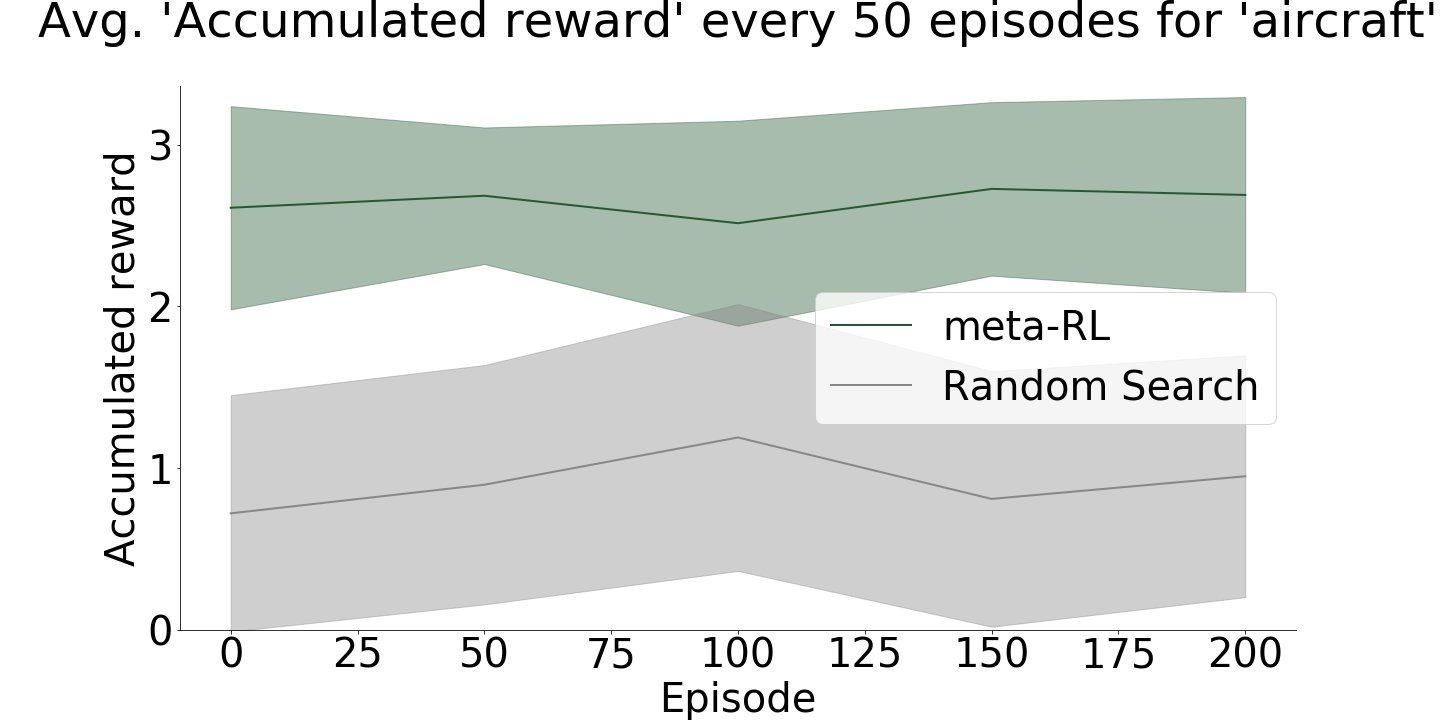}
  \caption{}
\label{fig:results:exp2:evolution:c}
\end{subfigure}
\begin{subfigure}{.33\textwidth}
  \centering
      \includegraphics[width=\linewidth]{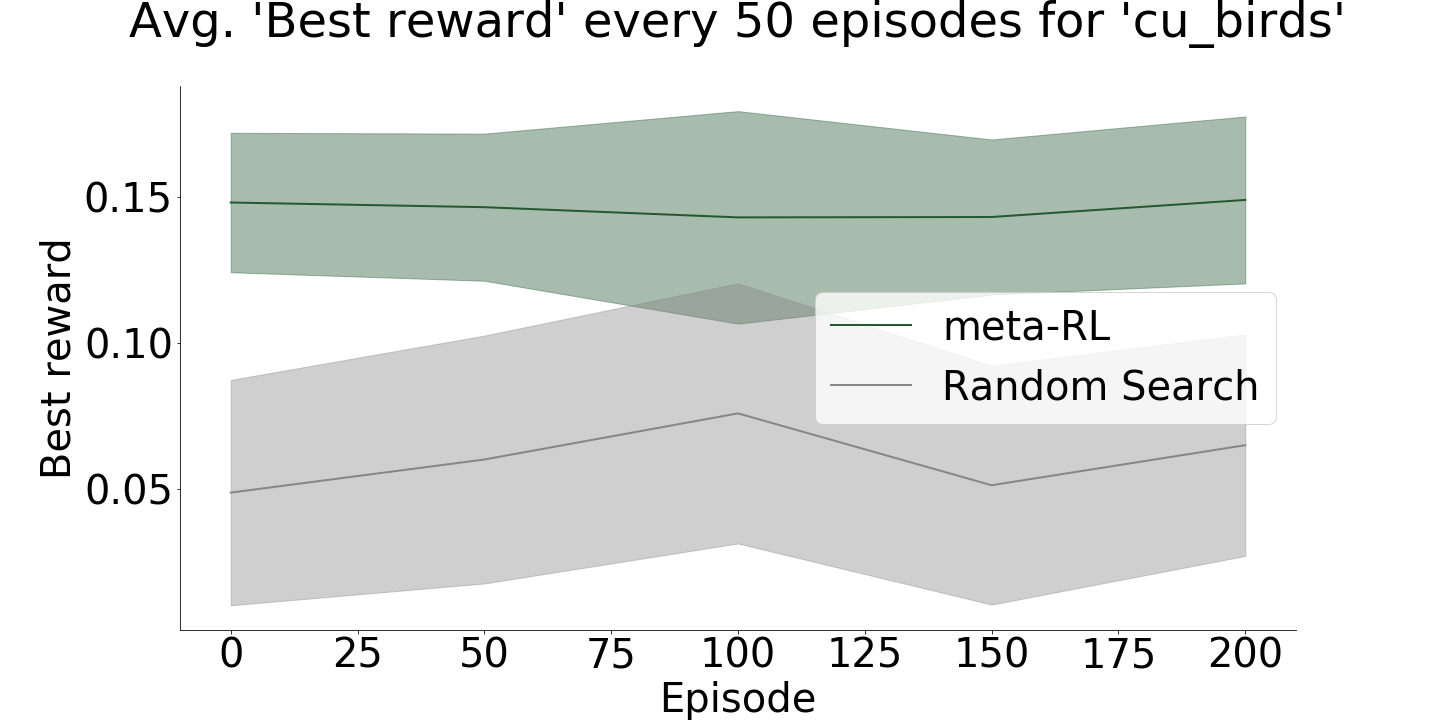}
  \caption{} 
\label{fig:results:exp2:evolution:d}
\end{subfigure}%
\begin{subfigure}{.33\textwidth}
  \centering
      \includegraphics[width=\linewidth]{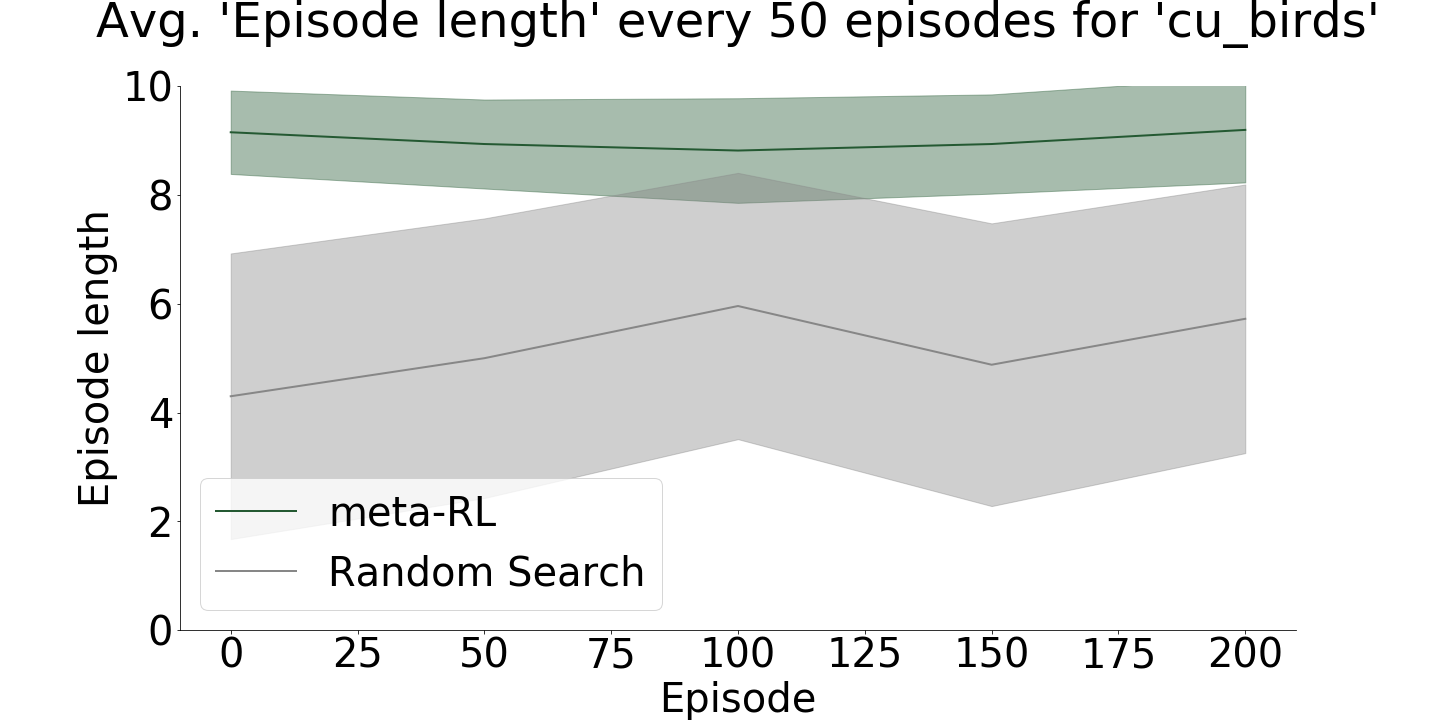}
  \caption{}
\label{fig:results:exp2:evolution:e}
\end{subfigure}%
\begin{subfigure}{.33\textwidth}
  \centering
      \includegraphics[width=\linewidth]{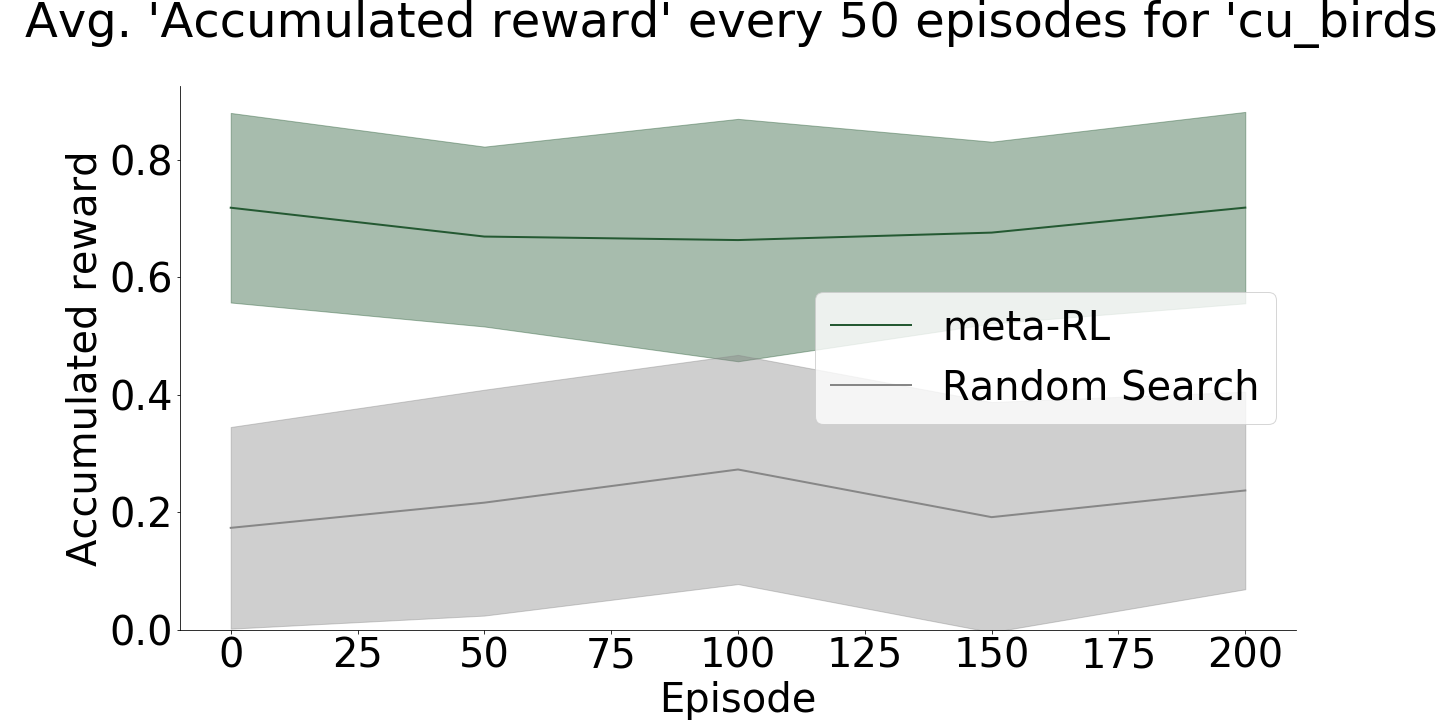}
  \caption{} 
\label{fig:results:exp2:evolution:f}
\end{subfigure}
\caption{Evolution of evaluation episodes through time from different perspectives, showing the means and $\pm 1$ standard deviations for every 50 episodes.}
\label{fig:results:exp2:evolution}
\vspace{-0.5cm}
\end{figure}

\begin{figure}[ht]
\centering
    \includegraphics[width=0.7\linewidth]{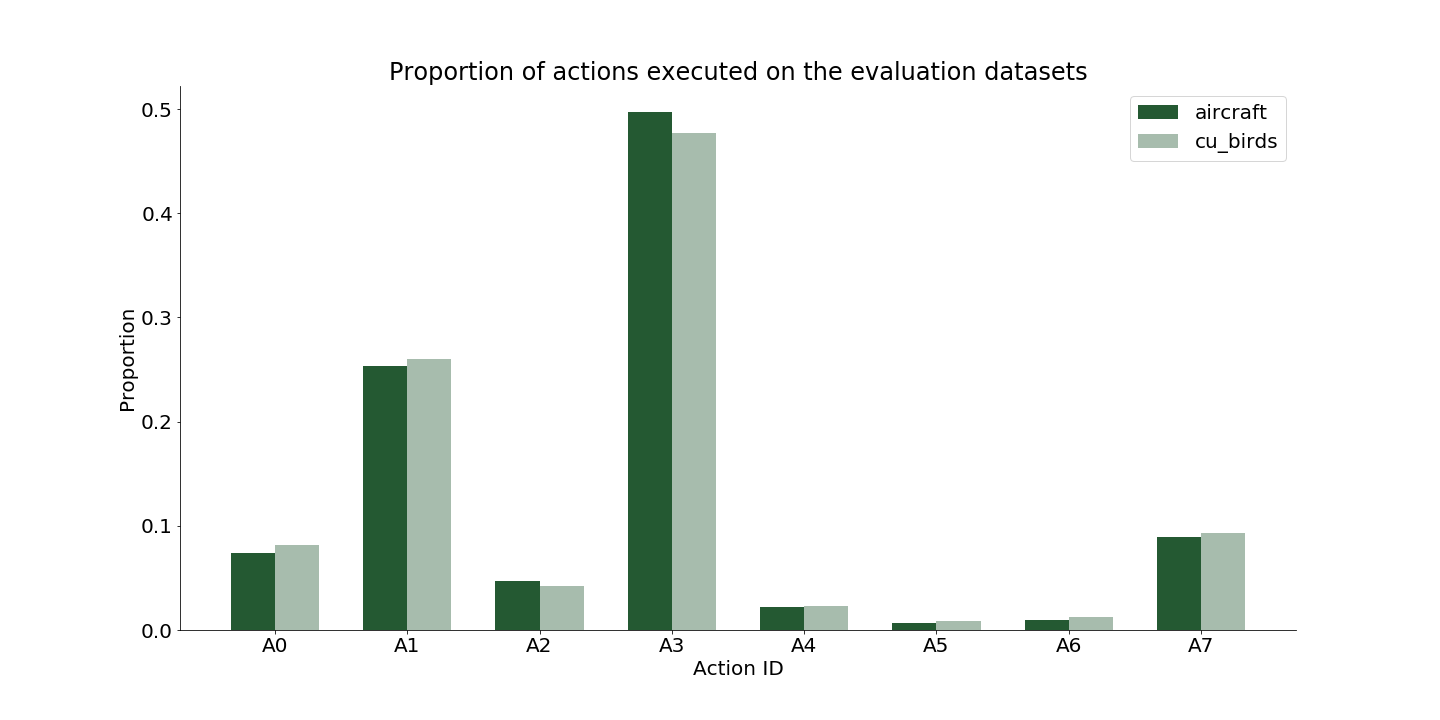}
\caption{Proportion of actions per dataset during evaluation. The labels in the x-axis match the IDs in Table~\ref{tab:methodology:rl:as}.}
\label{fig:results:exp2:actions}
\vspace{-0.5cm}
\end{figure}

The results of replaying the learned policy on completely new datasets are displayed in Figure~\ref{fig:results:exp2:evolution}, and the corresponding runtimes are listed in Table~\ref{tab:results:exp2:times}. They show that the agent immediately finds a good solution (with a deep network), and rewards remain consistent; however, it does not improve over time, which warrants further study (see Section~\ref{sec:conclusions}). Moreover, the strategy deployed by the agent is not different on each dataset, as it is observed in Figure~\ref{fig:results:exp2:actions}. We confirm that the strategies are not significantly different by performing a Wilcoxon signed-rank test with the null hypothesis that the two related paired samples come from the same distribution. The output is a $\text{p-value}=0.48$ with 95\% confidence.

\begin{table}[ht]
\centering
\begin{tabular}{@{}cc@{}}
\toprule
     Dataset             & Runtime \\ \midrule
aircraft  & 2 days 6h   \\
cu\_birds & 2 days 22h  \\ \midrule
Total                    & 5 days 4h  \\ \bottomrule
\end{tabular}
\caption{Running times for the evaluation of the deep meta-RL agent. All experiments ran on a single NVIDIA Tesla K40m GPU.}
\label{tab:results:exp2:times}
\end{table}

As we mentioned in Section~\ref{sec:experiments}, another result of interest is the performance of the best networks designed by the agent when they follow more intensive training. Table~\ref{tab:results:exp2:acc} shows the accuracy values obtained. We note that the networks achieve low accuracy, in the majority of the cases worse than random guessing. An important observation is that these low values can be a consequence of the relaxation made in the shape of the images. Whereas state-of-the-art architectures on both \textit{aircraft} and \textit{cu\_birds} work with shapes greater than $200 \times 200$, we use a smaller version of $84 \times 84$ that might lead to loss of information. Moreover, state-of-the-art results for these datasets are usually obtained after data augmentation and use deeper and more complex networks with multi-branch structures~\citep{FineGrained2, FineGrained3, FineGrainedResults}. However, in this experiment, we do not consider any of the latter aspects since we work under resource constraints that force us to make relaxations, and thus a lower accuracy can be expected.

Despite the low values, the architectures for the two datasets designed by the deep meta-RL agent outperformed by a significant amount the shortened version of VGG19. This shows that by using the learned policy it is possible to find better architectures than one inspired by state-of-the-art networks. A final observation is that the best architecture found by the agent during training did not become the best final network, thus exhibiting that early-stop can underestimate the long-term performance of the networks, which also warrants future work. 

\begin{table}[ht]
\centering
\begin{tabular}{@{}cccc@{}}
\toprule
Dataset   & Deep meta-RL (1st) & Deep meta-RL (2nd)          & VGG19-like                   \\ \midrule
aircraft  & 49.18 $\pm$ 1.2  & \textbf{50.11 $\pm$ 1.02} & 30.85 $\pm$ 10.82 \\
cu\_birds & 23.97 $\pm$ 1.28 & \textbf{24.24 $\pm$ 0.90} & 6.66 $\pm$ 1.98             \\ \bottomrule
\end{tabular}
\caption{Accuracy values of the best architectures after a more intense training. Every reported accuracy value is the mean $\pm$ 2 standard deviations of five independent trainings. For the sake of completeness, we show the designed networks in Appendix~\ref{app:networks}}
\label{tab:results:exp2:acc}
\end{table}

\subsection*{Experiment 3: training on a more complex environment}

\begin{figure}[ht]
\centering
\begin{subfigure}{.40\textwidth}
  \centering
      \includegraphics[width=\linewidth]{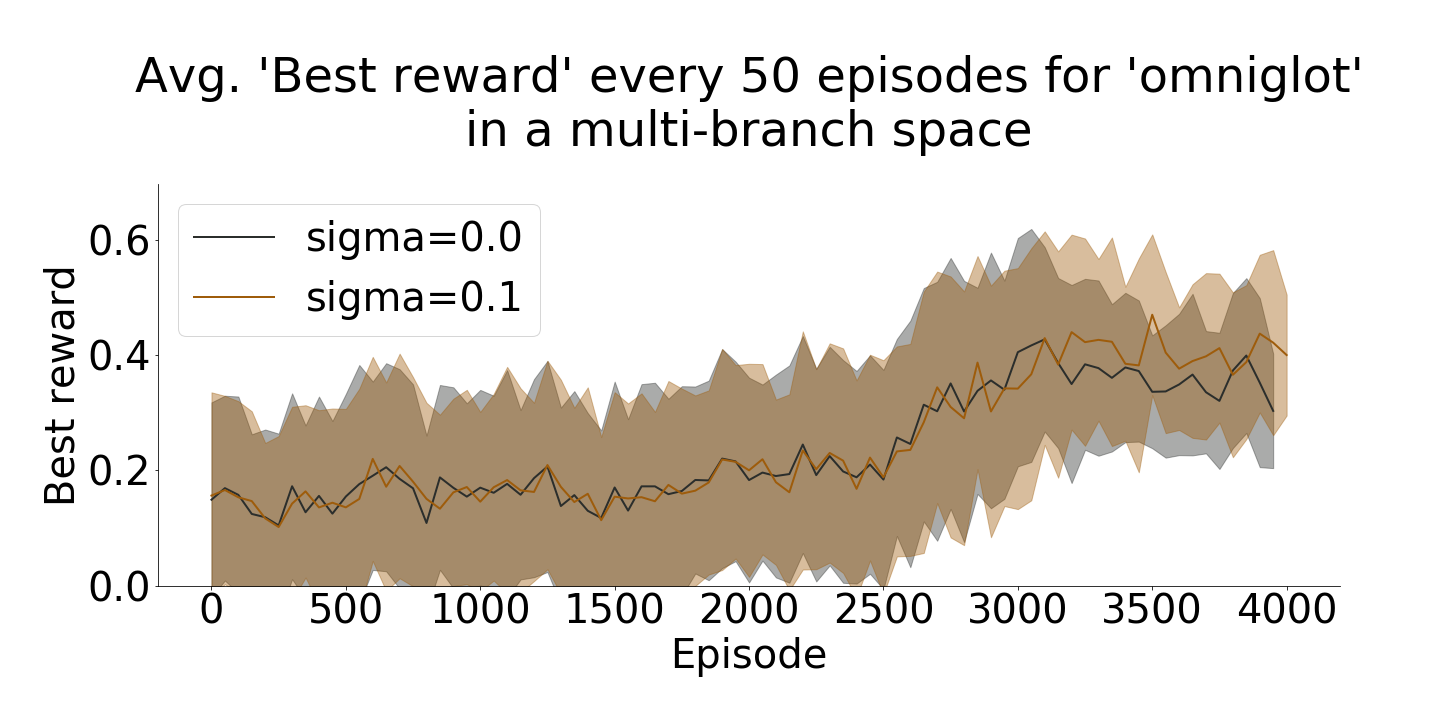}
  \caption{}
  \label{fig:results:exp3:evolution:a}
\end{subfigure}%
\begin{subfigure}{.40\textwidth}
  \centering
      \includegraphics[width=\linewidth]{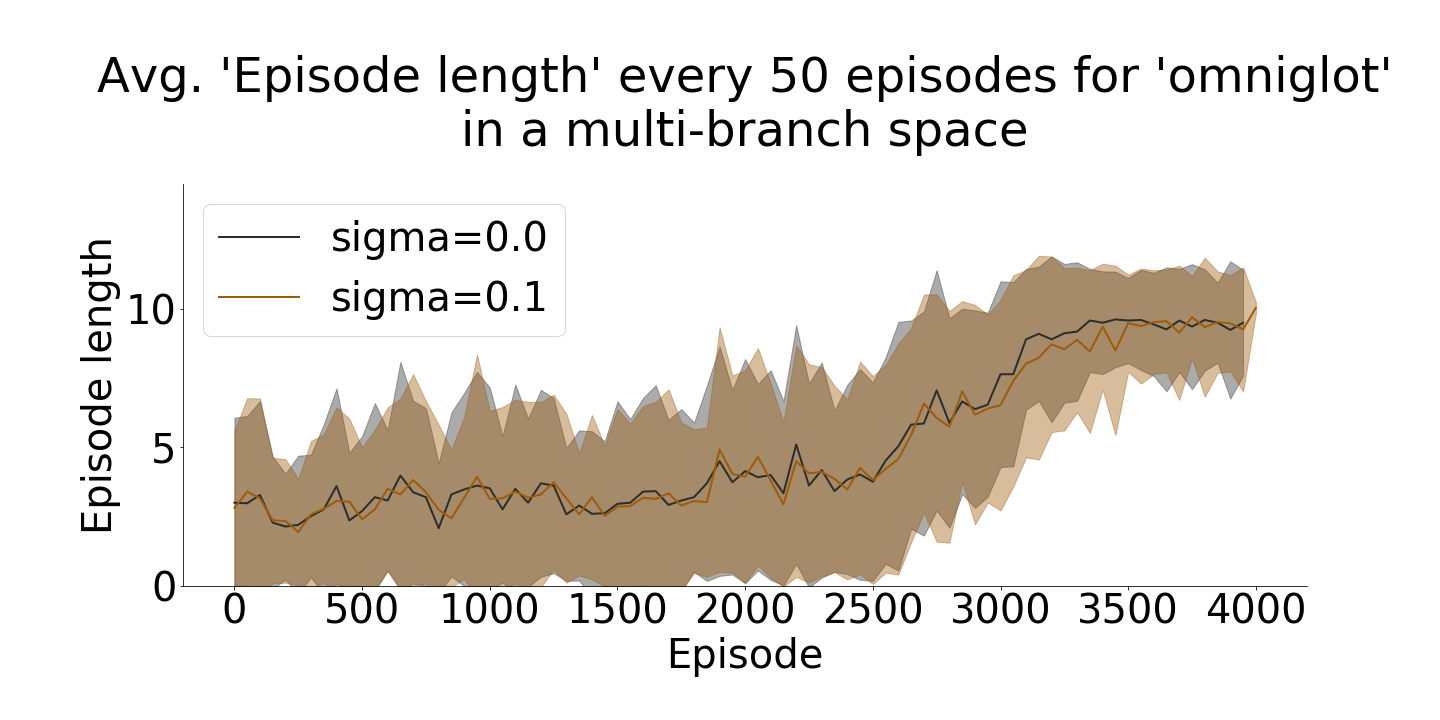}
  \caption{}
  \label{fig:results:exp3:evolution:b}
\end{subfigure}
\begin{subfigure}{.40\textwidth}
  \centering
      \includegraphics[width=\linewidth]{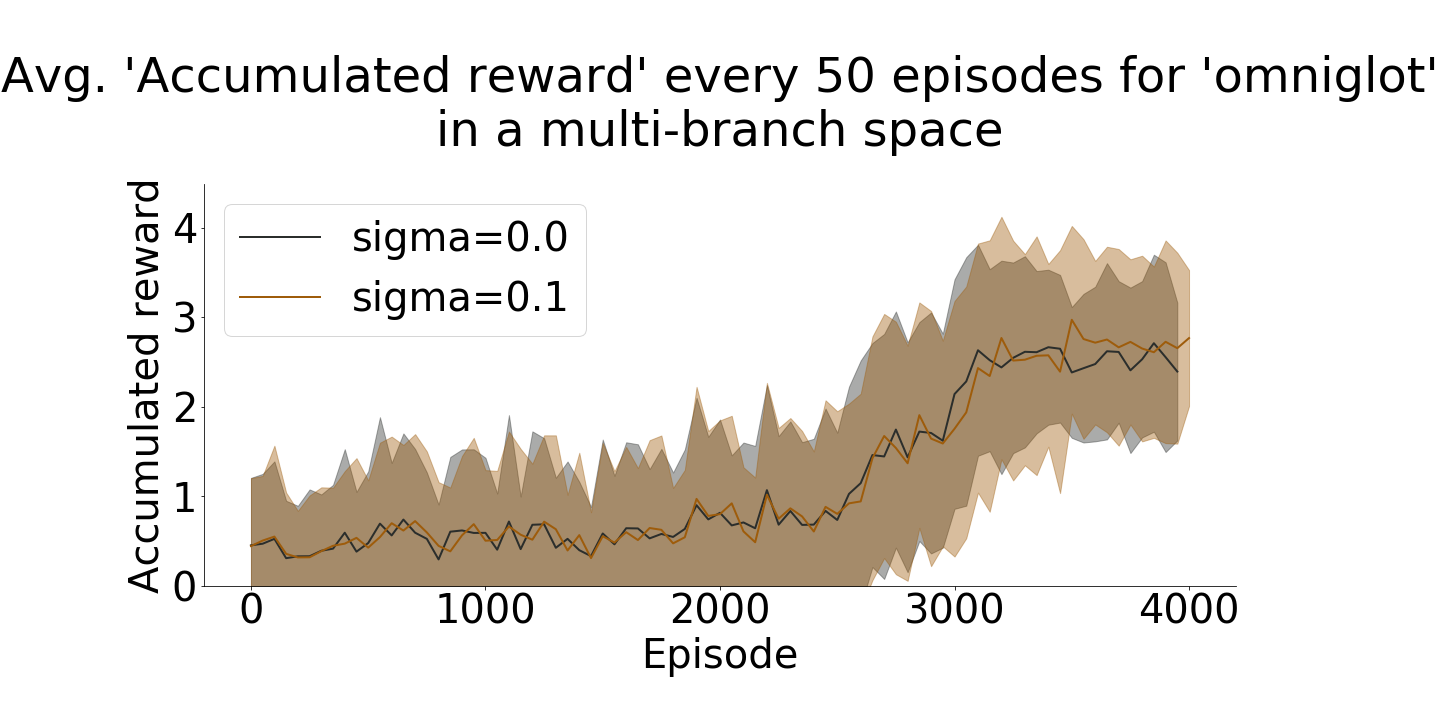}
  \caption{}
\label{fig:results:exp3:evolution:c}
\end{subfigure}
\caption{Evolution of training episodes for the multi-branch experiment from different perspectives. The plots show the means and $\pm 1$ standard deviations for every 50 episodes.}
\label{fig:results:exp3:evolution}
\vspace{-0.5cm}
\end{figure}

Figure~\ref{fig:results:exp3:evolution} shows the evolution of the \textit{best reward}, \textit{episode length}, and \textit{accumulated reward} during the multi-branch experiment on \textit{omniglot}. We do not observe differences in the behavior of the agent when using different $\sigma$ values, but we note that it took longer to output meaningful rewards (around episode 3000) when compared to Experiment 1, causing extended runtimes as shown in Table~\ref{tab:results:exp3:times}.

\begin{table}[ht]
\centering
\begin{tabular}{@{}cc@{}}
\toprule
                  & Runtime \\ \midrule
$\sigma = 0.0$ & 13 days 9h   \\
$\sigma = 0.1$ & 15 days 14h  \\ \midrule
Total                    & 28 days 23h  \\ \bottomrule
\end{tabular}
\caption{Runtimes for the training of the deep meta-RL agent in the multi-branch search space for \textit{omniglot}. All experiments ran on a single NVIDIA Tesla K40m GPU.}
\label{tab:results:exp3:times}
\end{table}

\begin{figure}[ht]
\centering
\begin{subfigure}{.53\textwidth}
  \centering
      \includegraphics[width=\linewidth]{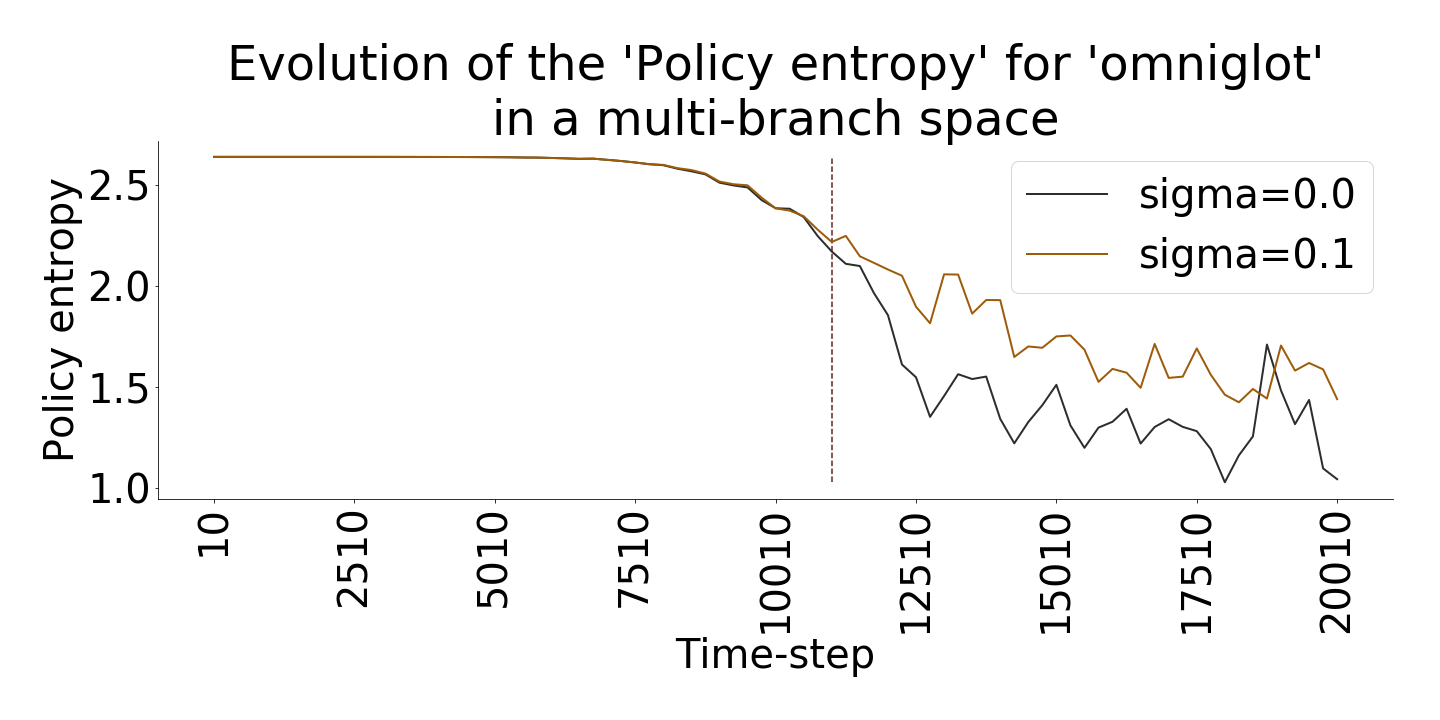}
  \caption{}
  \label{fig:results:exp3:exploration:a}
\end{subfigure}%
\begin{subfigure}{.44\textwidth}
  \centering
      \includegraphics[width=\linewidth]{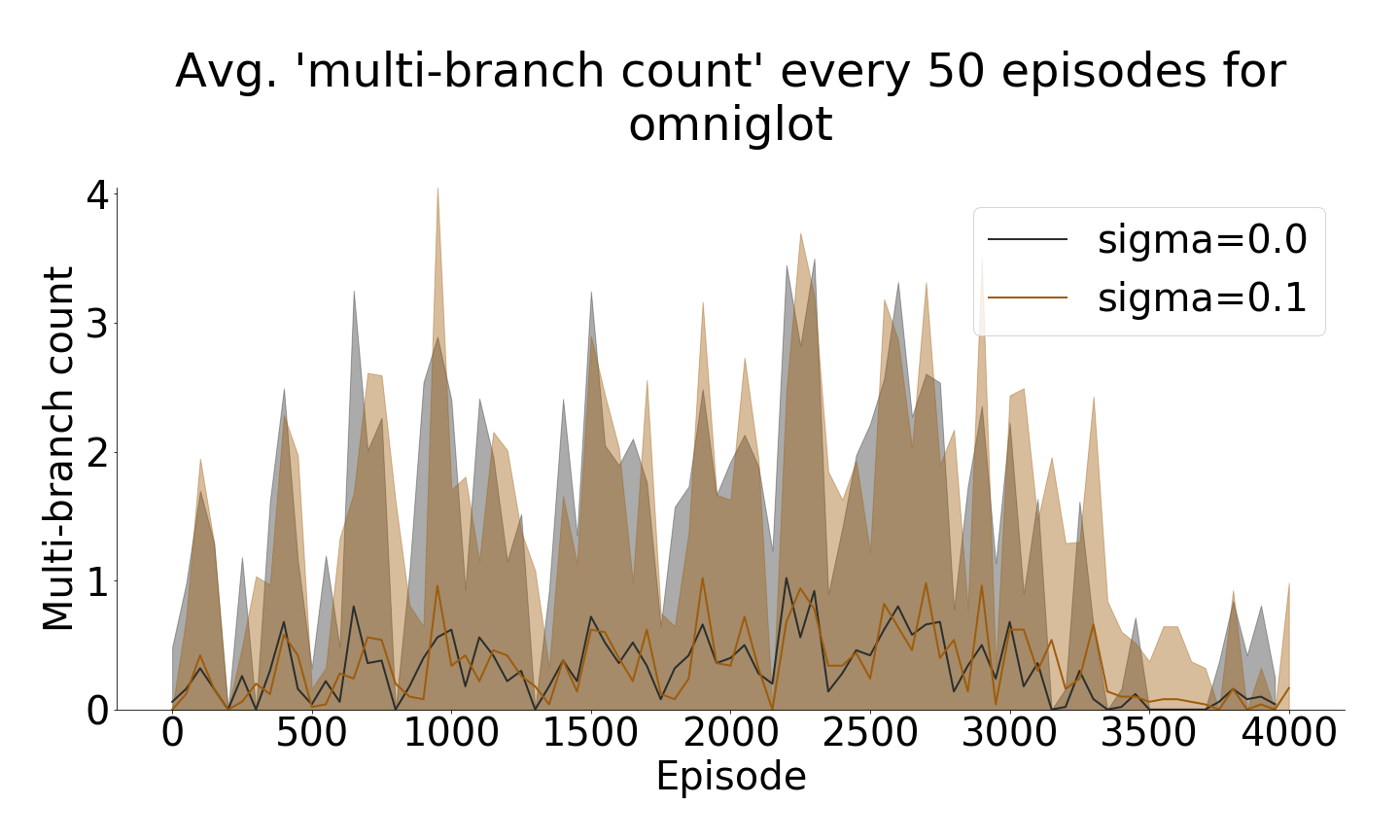}
  \caption{}
  \label{fig:results:exp3:exploration:b}
\end{subfigure}
\caption{The exploration of the agent through time. (a) The entropy of the policy through time-steps. The vertical line cuts the horizontal axis at the time-step where the episde 3000 starts. (b) The count of multi-branch structures explored by the agent, showing the mean $\pm$ 1 standard deviation every 50 episodes.\\}
\label{fig:results:exp3:exploration}
\vspace{-0.5cm}
\end{figure}

\begin{figure}[ht]
\centering
    \includegraphics[width=0.65\linewidth]{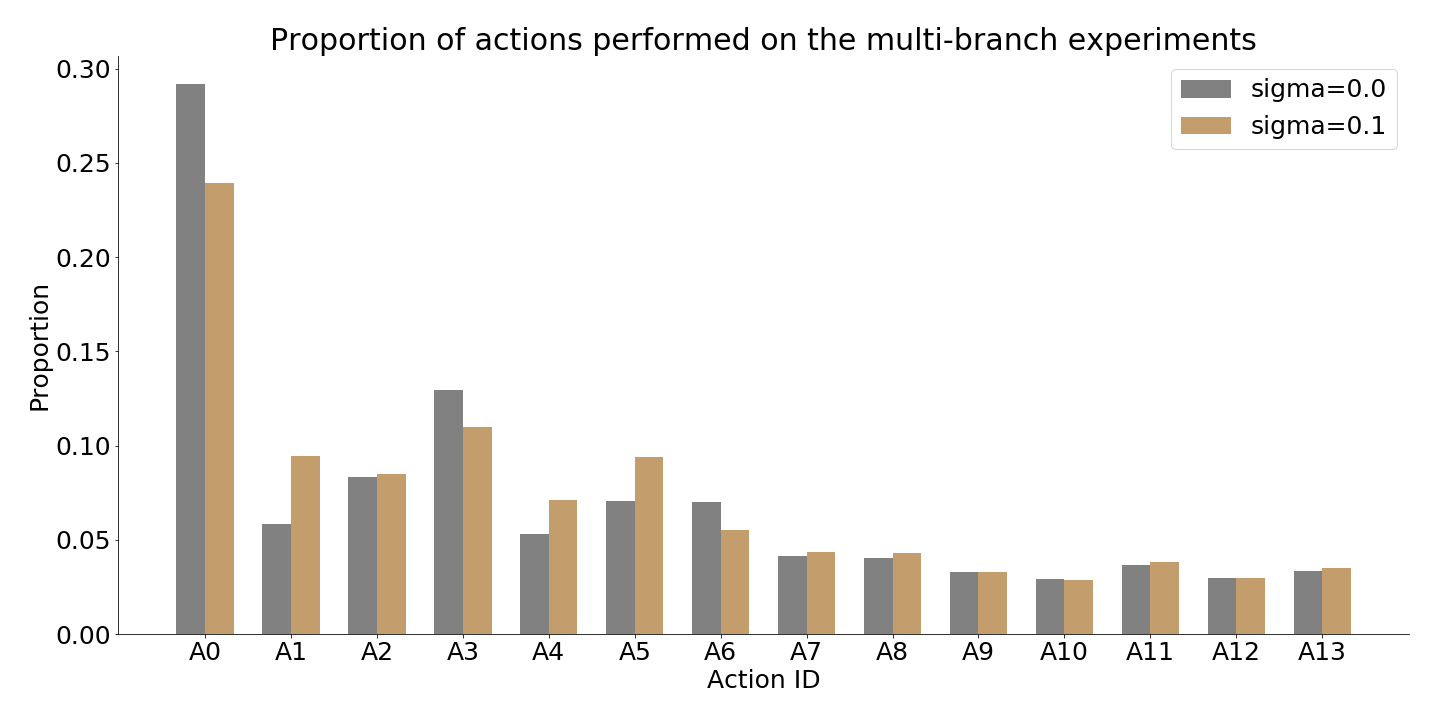}
\caption{Proportion of actions taken by the agent in the multi-branch experiments. The labels in the x-axis match the IDs in Table~\ref{tab:methodology:rl:as}.}
\label{fig:results:exp3:actions}
\end{figure}

However, we stated in Section~\ref{sec:experiments:multibranch} that our main interest in this experiment is to study whether or not the agent can explore multi-branch structures. Figure~\ref{fig:results:exp3:exploration:a} shows the entropy of the policy through time-steps, and Figure~\ref{fig:results:exp3:exploration:b} the count of multi-branch structures through episodes. We note that during exploration the appearance of multi-branch structures is more likely, and after episode 3000 (represented by the vertical line in Figure~\ref{fig:results:exp3:exploration:a}), when exploration drops down, the multi-branch structures become less frequent. Furthermore, we found that the proportion of multi-branch vs.~chain-structured networks is only 1:10, meaning that the agent did not explore multi-branch structures aggressively, and settled for chain-structured networks instead. The latter is supported by the proportion of actions displayed in Figure~\ref{fig:results:exp3:actions}, where the actions \textsc{A8-13} (related to multi-branch structures) are the least frequent.

We believe that a multi-branch space requires us to handle differently how the predecessors of the NSC vectors are selected (see Section~\ref{sec:methodology:rl:as}). Some alternatives are: defining heuristics to chose the predecessors instead of using the shifting operations, assigning other rewards to the actions related to the predecessors, and modifying the hyper-parameters of the A2C network to encourage more exploration of the agent in the beginning of the deep meta-RL training.

%% file: conclusions.tex
\section{Conclusions and future work}\label{sec:conclusions}

In this work, we presented the first application of deep meta-RL in the NAS setting. Firstly, we investigated the advantages of deep meta-RL against standard RL on the relatively simple scenario of chain-structured architectures. Despite resource limitations (1 GPU only), we observed that a policy learned using deep meta-RL can be transferred to other environments and quickly designs architectures with higher and more consistent accuracy than standard RL. Nevertheless, standard RL outperforms meta-RL when both learn a policy from scratch. We also note that the meta-RL agent exhibited adaptive behavior during the training, changing its strategy according to the dataset in question.  Secondly, we analyzed the adaptability of the agent during evaluation (i.e., when the policy's weights are fixed) and the quality of the networks that it designs for previously unseen datasets. In our experiment, the agent was not able to adapt its strategy to different environments, but the performance of the networks delivered was better than the performance of a human-designed network, showing that the knowledge developed by our agent in the training environments is meaningful in others. Thirdly, we extended our approach to a more complex NAS scenario with a multi-branch search space. In this setting, the meta-RL agent was not able to deeply explore the multi-branch structures and settled for chain-structured networks instead.

We conclude that deep meta-RL does provide an advantage over standard RL when transferring is enabled, and it can effectively adapt its strategy to different environments during training. Moreover, the policy learned can be used to deliver meaningful and well-performing architectures on previously unseen environments without further training. We believe that it is possible to strengthen our deep-meta RL framework in future work. Specifically, we propose to investigate the following aspects under more powerful computational resources:

\begin{itemize}
    \setlength\itemsep{0em}
    \setlength\parskip{0pt}
    \item[-] \textit{Hyper-parameter tunning of the A2C components}. In Experiment 1, we observed that the learning progress of the meta-RL agent is slow. We also noticed a long exploration window in the first environment. In order to improve these aspects, we propose to tune the hyper-parameters according to the next intuitions:

        \begin{itemize}
            \setlength\itemsep{0em}
            \item[-]  $j$: the parameter controlling the number of steps before a learning update. We suggest reducing this value to speed up learning.
            \item[-] $\eta$: the entropy regularization. Experiments varying the range of this hyper-parameter are required to observe its impact on the learning curve. Also, different values could be used depending on the hardness of the environments.
            \item[-] $\alpha$: the learning rate. We suggest exploring decay functions for the learning rate to encourage faster learning after exploration.
        \end{itemize}
    
    \item[-] \textit{Duration of the agent-environment interaction}. In Experiment 2, the policy did not exhibit adaptive behavior. A possible explanation is that the training trials were relatively short when compared to other reinforcement learning applications. Training the agent for longer trials could help improve the adaptation of the policy during evaluation.
    
    \item[-] \textit{The action space in the multi-branch setting}. In Experiment 3, we observed that the agent was not able to explore the multi-branch space sufficiently and settled for chain-structured networks instead. Although hyper-parameter tuning could also help encourage exploration of the multi-branching actions, we believe that redefining the actions is a more suitable area of improvement. In that line, we recommend exploring heuristics based on the number of connections to select the predecessors.

    \item[-] \textit{The datasets and the performance estimation strategy}. In all experiments, we observed a low accuracy of the networks in the datasets. Since we worked with constrained resources, we applied relaxations to the datasets and the performance estimation strategy to reduce the computational cost, which could have affected the accuracy of the networks. Future work can focus on designing a different set of environments with images with a smaller size, optimizing the performance estimation strategy per dataset, and investigating alternatives to reduce the cost of computing the rewards associated to the networks.
    
    \item[-] \textit{Transforming other standard RL algorithms to a meta-RL version}. The transformation of the A2C algorithm to a meta-RL implementation required to change the input passed to the policy and to rely on a recurrent unit to learn the temporal dependencies between actions. This transformation is possible on other standard RL algorithms, which would help study different meta-RL approaches to NAS.
    
    \item[-] \textit{Benchmarking of other RL on the same NAS environments}. In Section~\ref{sec:software}, we introduced the system developed to conduct our experiments, which allows to easily play other RL algorithms from the OpenAI baselines on the same NAS environments. We believe that this system will help to encourage research in these directions so that the benefits of different RL algorithms on NAS can be studied in detail.
\end{itemize}

%% file: ack.tex

\acks{To the \textit{SURF} cooperative for kindly providing the required computational resources. To Jane Wang for her valuable feedback to this work.
}

%% file: appendix_datasets.tex
\section{Selection of the datasets}\label{app:datasets}

The deep meta-reinforcement learning framework that we implement requires a set of environments associated to image classification tasks. In order to design these environments, we rely on the meta-dataset~\citep{MetaDataset}, a collection of 10 datasets with a concrete sampling procedure designed for meta-learning in few-shot learning image classification. In our setting, the datasets are intended for standard image classification, thus we redefine the sampling strategy. Our interest is in using small but yet challenging datasets that allow us to save computational resources without making the Neural Architecture Search (NAS) trivial.

\begin{figure}[ht]
\centering
\begin{subfigure}{.5\textwidth}
  \centering
      \includegraphics[width=0.8\linewidth]{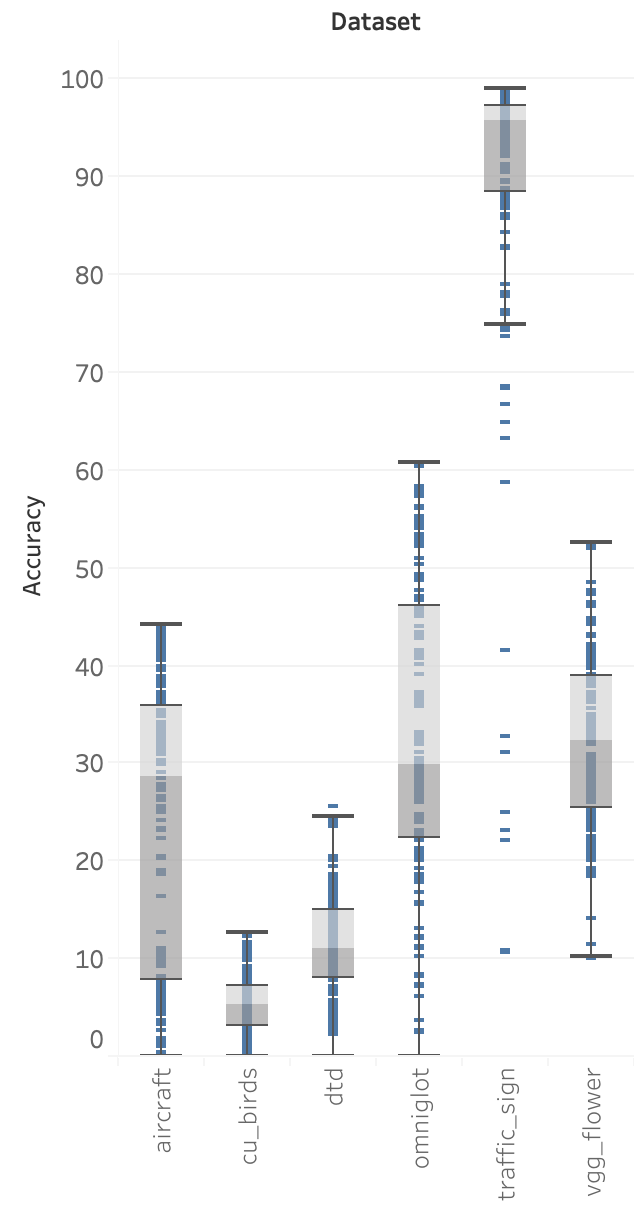}
  \caption{}
  \label{fig:appA:boxplot}
\end{subfigure}%
\begin{subfigure}{.5\textwidth}
  \centering
      \includegraphics[width=0.9\linewidth]{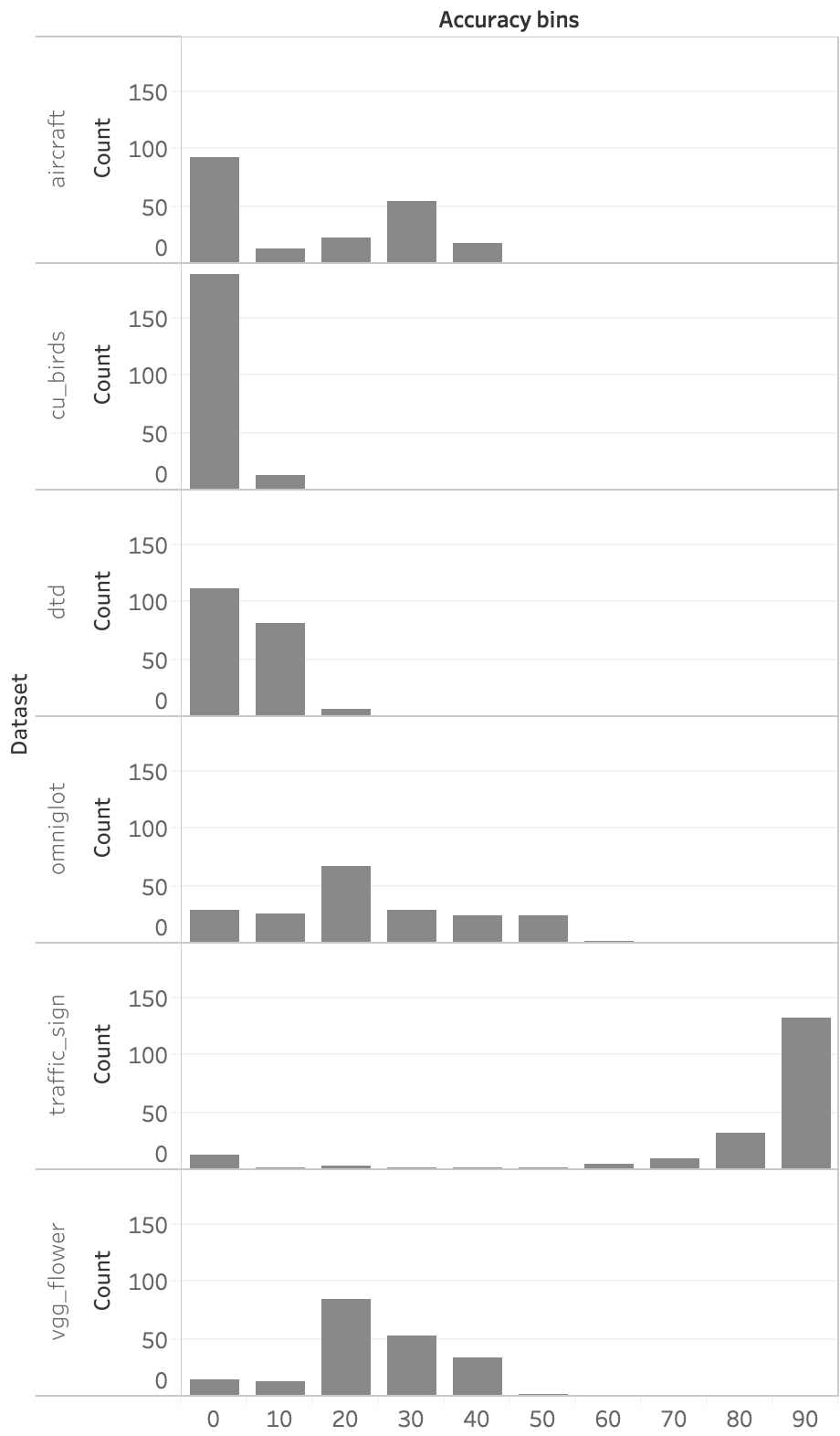}
  \caption{}
  \label{fig:appA:histogram}
\end{subfigure}
\caption{Different visualizations of the early-stop accuracy values obtained to study the hardness of the datasets.}
\label{fig:appA:trialstats}
\end{figure}

In Table~\ref{tab:appA:metadataset} the original datasets in the collection are listed. We select the ones that are smaller than CIFAR-10 (60K observations), which is the reference for NAS. The datasets satisfying the criterion are \textit{aircraft}, \textit{cu\_birds}, \textit{dtd}, \textit{omniglot}, \textit{traffic\_sign} and \textit{vgg\_flower}. We want to evaluate the hardness of these six datasets to define a sampling procedure from the collection, and thus we perform a short and individual deep meta-reinforcement learning trial with $t_{max}=200$ for each dataset. Since at the beginning of the trial the agent does not develop any significant knowledge, its sampling of architectures is random. In Figure~\ref{fig:appA:trialstats} the boxplot and barplot of the obtained accuracy values are presented, and in Table~\ref{tab:appA:times} the running time per experiment is shown.

A simple exploratory analysis suggests three types of datasets: a ``trivial" dataset with high accuracy values with simple networks (\textit{traffic\_sign}), two ``hard" datasets with low accuracy values (all values below 30\%: \textit{dtd} and \textit{cu\_birds}), and three ``medium" datasets with more diversity of accuracy values (median around 30\% and broader interquartile range: \textit{aircraft}, \textit{omniglot}, \textit{vgg\_flower}). On the other hand, for the running times, we can observe that \textit{aircraft} and \textit{cu\_birds} result in the most expensive runs. Considering the computation time, and the hardness of the classification tasks, we defined the sampling presented in Table~\ref{tab:methodology:environments:datasets}. Our training datasets have different levels of hardness and reported the least costly runs.

\begin{table}[ht]
\centering
\begin{tabular}{cccc}
\hline
Dataset ID    & Dataset name                              & N classes & N observations \\ \hline
aircraft      & FGVC-Aircraft                             & 100       & 10000          \\
cu\_birds     & CUB-200-2011                              & 200       & 11788          \\
dtd           & Describable Textures                      & 47        & 5640           \\
fungi         & FGVCx Fungi                               & 1394      & 89760          \\
ilsvrc\_2012  & ImageNet                                  & 1000      & 1280764        \\
mscoco        & Common Objects in Context                 & 80        & 330000         \\
omniglot      & Omniglot                                  & 1623      & 32460          \\
quickdraw     & Quick, Draw!                              & 345       & 50426266       \\
traffic\_sign & German Traffic Sign Recognition Benchmark & 43        & 39209          \\
vgg\_flower   & VGG Flower                                & 102       & 8189           \\ \hline
\end{tabular}
\caption{The original meta-dataset~\citep{MetaDataset} with the number of classes and observations after conversion with the official source code.}
\label{tab:appA:metadataset}
\end{table}

\begin{table}[ht]
\centering
\begin{tabular}{cc}
\hline
Dataset ID    & Time   \\ \hline
aircraft      & 9h49m  \\
cu\_birds     & 16h20m \\
dtd           & 5h38m  \\
omniglot      & 3h38m  \\
traffic\_sign & 4h33m  \\
vgg\_flower   & 4h56m  \\ \hline
\end{tabular}
\caption{Running times of a deep meta-RL trial with $t_{max}=200$, used to study the hardness and cost of each dataset.}
\label{tab:appA:times}
\end{table}

%% file: appendix_networks.tex
\section{Networks designed by the deep meta-RL agent during training and evaluation}\label{app:networks}

\begin{figure}[ht]
\centering
\begin{subfigure}{.40\textwidth}
\smallskip
\smallskip
\smallskip
\smallskip
\smallskip
\smallskip
\smallskip
\begin{center}
\begin{tikzpicture}

\tikzstyle{input} = [rectangle, minimum width=2cm, minimum height=0.8cm, text centered, draw=black, fill=gray!0, text width=2cm]

\tikzstyle{convolution} = [rectangle, minimum width=2cm, minimum height=0.8cm, text centered, draw=black, fill=gray!10, text width=2.5cm]

\tikzstyle{maxpool} = [rectangle, minimum width=2cm, minimum height=0.8cm, text centered, draw=black, fill=gray!40, text width=2.5cm]

\tikzstyle{avgpool} = [rectangle, minimum width=2cm, minimum height=0.8cm, text centered, draw=black, fill=gray!40, text width=2.5cm]

\tikzstyle{concat} = [rectangle, minimum width=2cm, minimum height=0.8cm, text centered, draw=black, fill=gray!80, text width=2.5cm]

\tikzstyle{arrow} = [->,>=stealth]

\node (l0) [input] at (0,0) {\footnotesize Input (84x84)};

\node (l1) [convolution, below of=l0, yshift=-0.5cm] {\footnotesize Convolution k=3 (82x82)};

\node (l2) [convolution, below of=l1, yshift=-0.5cm] {\footnotesize Convolution k=5 (78x78)};

\node (l3) [maxpool, below of=l2, yshift=-0.5cm] {\footnotesize MaxPooling p=2 (39x39)};

\node (l4) [maxpool, below of=l3, yshift=-0.5cm] {\footnotesize MaxPooling p=2 (19x19)};

\draw [arrow] (l0) -- (l1);
\draw [arrow] (l1) -- (l2);
\draw [arrow] (l2) -- (l3);
\draw [arrow] (l3) -- (l4);

\end{tikzpicture}
\smallskip
\smallskip
\smallskip
\smallskip
\smallskip
\smallskip
\smallskip
\caption{}
\label{fig:app:networks:aircraft:a}
\end{center}
\end{subfigure}%
\begin{subfigure}{.40\textwidth}
\begin{center}
\begin{tikzpicture}

\tikzstyle{input} = [rectangle, minimum width=2cm, minimum height=0.8cm, text centered, draw=black, fill=gray!0, text width=2cm]

\tikzstyle{convolution} = [rectangle, minimum width=2cm, minimum height=0.8cm, text centered, draw=black, fill=gray!10, text width=2.5cm]

\tikzstyle{maxpool} = [rectangle, minimum width=2cm, minimum height=0.8cm, text centered, draw=black, fill=gray!40, text width=2.5cm]

\tikzstyle{avgpool} = [rectangle, minimum width=2cm, minimum height=0.8cm, text centered, draw=black, fill=gray!40, text width=2.5cm]

\tikzstyle{concat} = [rectangle, minimum width=2cm, minimum height=0.8cm, text centered, draw=black, fill=gray!80, text width=2.5cm]

\tikzstyle{arrow} = [->,>=stealth]

\node (l0) [input] at (0,0) {\footnotesize Input (84x84)};

\node (l1) [convolution, below of=l0, yshift=-0.5cm] {\footnotesize Convolution k=1 (84x84)};

\node (l2) [convolution, below of=l1, yshift=-0.5cm] {\footnotesize Convolution k=3 (82x82)};

\node (l3) [convolution, below of=l2, yshift=-0.5cm] {\footnotesize Convolution k=3 (80x80)};

\node (l4) [maxpool, below of=l3, yshift=-0.5cm] {\footnotesize MaxPooling p=2 (40x40)};

\node (l5) [maxpool, below of=l4, yshift=-0.5cm] {\footnotesize MaxPooling p=3 (13x13)};

\draw [arrow] (l0) -- (l1);
\draw [arrow] (l1) -- (l2);
\draw [arrow] (l2) -- (l3);
\draw [arrow] (l3) -- (l4);
\draw [arrow] (l4) -- (l5);

\end{tikzpicture}
\small
\caption{}
\label{fig:app:networks:aircraft:b}
\end{center}
\end{subfigure}
\caption{Best architectures designed for \textit{aircraft} during evaluation of the policy. (a) The best architecture with early-stop accuracy of 48.22. (b) The second-best architecture with early-stop accuracy of 47.95}
\label{fig:app:networks:aircraft}
\vspace{-0.5cm}
\end{figure}
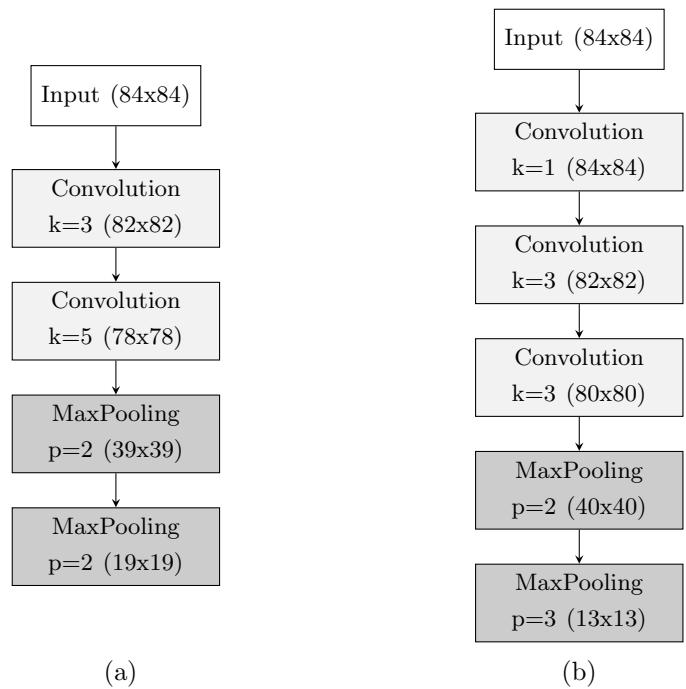

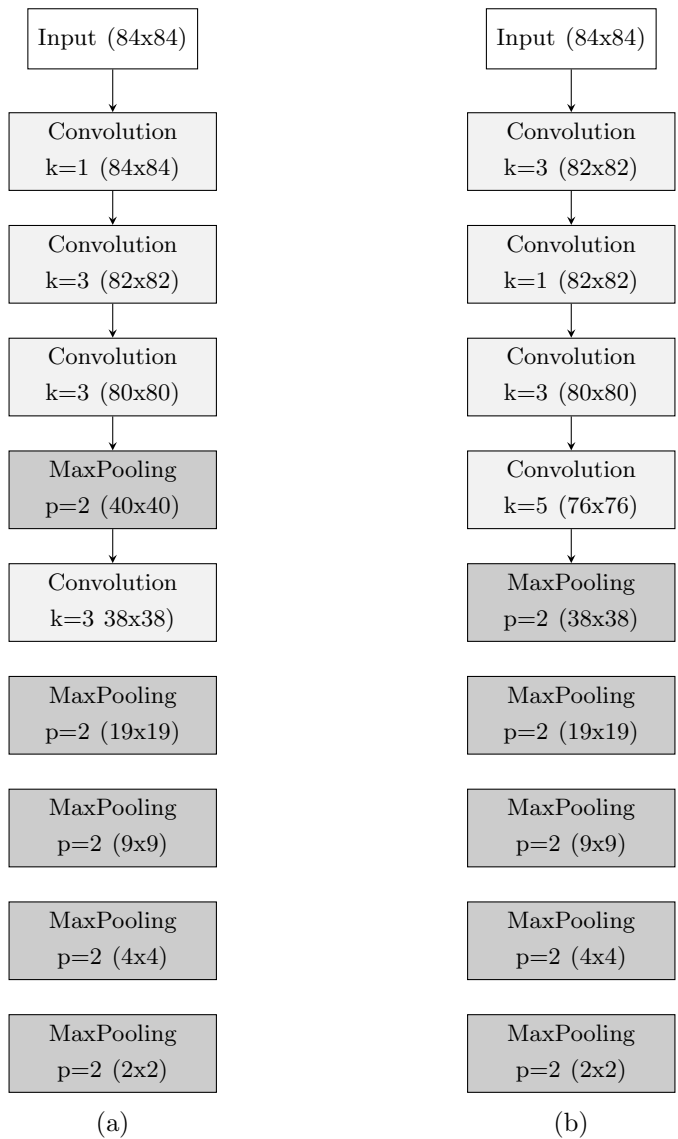
\begin{figure}[ht]
\centering
\begin{subfigure}{.40\textwidth}
\begin{center}
\begin{tikzpicture}

\tikzstyle{input} = [rectangle, minimum width=2cm, minimum height=0.8cm, text centered, draw=black, fill=gray!0, text width=2cm]

\tikzstyle{convolution} = [rectangle, minimum width=2cm, minimum height=0.8cm, text centered, draw=black, fill=gray!10, text width=2.5cm]

\tikzstyle{maxpool} = [rectangle, minimum width=2cm, minimum height=0.8cm, text centered, draw=black, fill=gray!40, text width=2.5cm]

\tikzstyle{avgpool} = [rectangle, minimum width=2cm, minimum height=0.8cm, text centered, draw=black, fill=gray!40, text width=2.5cm]

\tikzstyle{concat} = [rectangle, minimum width=2cm, minimum height=0.8cm, text centered, draw=black, fill=gray!80, text width=2.5cm]

\tikzstyle{arrow} = [->,>=stealth]

\node (l0) [input] at (0,0) {\footnotesize Input (84x84)};

\node (l1) [convolution, below of=l0, yshift=-0.5cm] {\footnotesize Convolution k=1 (84x84)};

\node (l2) [convolution, below of=l1, yshift=-0.5cm] {\footnotesize Convolution k=3 (82x82)};

\node (l3) [convolution, below of=l2, yshift=-0.5cm] {\footnotesize Convolution k=3 (80x80)};

\node (l4) [maxpool, below of=l3, yshift=-0.5cm] {\footnotesize MaxPooling p=2 (40x40)};

\node (l5) [convolution, below of=l4, yshift=-0.5cm] {\footnotesize Convolution k=3 38x38)};

\node (l6) [maxpool, below of=l5, yshift=-0.5cm] {\footnotesize MaxPooling p=2 (19x19)};

\node (l7) [maxpool, below of=l6, yshift=-0.5cm] {\footnotesize MaxPooling p=2 (9x9)};

\node (l8) [maxpool, below of=l7, yshift=-0.5cm] {\footnotesize MaxPooling p=2 (4x4)};

\node (l9) [maxpool, below of=l8, yshift=-0.5cm] {\footnotesize MaxPooling p=2 (2x2)};

\draw [arrow] (l0) -- (l1);
\draw [arrow] (l1) -- (l2);
\draw [arrow] (l2) -- (l3);
\draw [arrow] (l3) -- (l4);
\draw [arrow] (l4) -- (l5);

\end{tikzpicture}
\caption{}
\label{fig:app:networks:cubirds:a}
\end{center}
\end{subfigure}%
\begin{subfigure}{.40\textwidth}
\begin{center}
\begin{tikzpicture}

\tikzstyle{input} = [rectangle, minimum width=2cm, minimum height=0.8cm, text centered, draw=black, fill=gray!0, text width=2cm]

\tikzstyle{convolution} = [rectangle, minimum width=2cm, minimum height=0.8cm, text centered, draw=black, fill=gray!10, text width=2.5cm]

\tikzstyle{maxpool} = [rectangle, minimum width=2cm, minimum height=0.8cm, text centered, draw=black, fill=gray!40, text width=2.5cm]

\tikzstyle{avgpool} = [rectangle, minimum width=2cm, minimum height=0.8cm, text centered, draw=black, fill=gray!40, text width=2.5cm]

\tikzstyle{concat} = [rectangle, minimum width=2cm, minimum height=0.8cm, text centered, draw=black, fill=gray!80, text width=2.5cm]

\tikzstyle{arrow} = [->,>=stealth]

\node (l0) [input] at (0,0) {\footnotesize Input (84x84)};

\node (l1) [convolution, below of=l0, yshift=-0.5cm] {\footnotesize Convolution k=3 (82x82)};

\node (l2) [convolution, below of=l1, yshift=-0.5cm] {\footnotesize Convolution k=1 (82x82)};

\node (l3) [convolution, below of=l2, yshift=-0.5cm] {\footnotesize Convolution k=3 (80x80)};

\node (l4) [convolution, below of=l3, yshift=-0.5cm] {\footnotesize Convolution k=5 (76x76)};

\node (l5) [maxpool, below of=l4, yshift=-0.5cm] {\footnotesize MaxPooling p=2 (38x38)};

\node (l6) [maxpool, below of=l5, yshift=-0.5cm] {\footnotesize MaxPooling p=2 (19x19)};

\node (l7) [maxpool, below of=l6, yshift=-0.5cm] {\footnotesize MaxPooling p=2 (9x9)};

\node (l8) [maxpool, below of=l7, yshift=-0.5cm] {\footnotesize MaxPooling p=2 (4x4)};

\node (l9) [maxpool, below of=l8, yshift=-0.5cm] {\footnotesize MaxPooling p=2 (2x2)};

\draw [arrow] (l0) -- (l1);
\draw [arrow] (l1) -- (l2);
\draw [arrow] (l2) -- (l3);
\draw [arrow] (l3) -- (l4);
\draw [arrow] (l4) -- (l5);

\end{tikzpicture}
\small
\caption{}
\label{fig:app:networks:cubirds:b}
\end{center}
\end{subfigure}
\caption{Best architectures designed for \textit{cu\_birds} during evaluation of the policy. (a) The best architecture with early-stop accuracy of 19.22. (b) The second-best architecture with early-stop accuracy of 19.06}
\label{fig:app:networks:cubirds}
\vspace{-0.5cm}
\end{figure}

\begin{figure}[ht]
\centering
\begin{subfigure}{.40\textwidth}
\begin{center}
\begin{tikzpicture}

\tikzstyle{input} = [rectangle, minimum width=2cm, minimum height=0.8cm, text centered, draw=black, fill=gray!0, text width=2cm]

\tikzstyle{convolution} = [rectangle, minimum width=2cm, minimum height=0.8cm, text centered, draw=black, fill=gray!10, text width=2.5cm]

\tikzstyle{maxpool} = [rectangle, minimum width=2cm, minimum height=0.8cm, text centered, draw=black, fill=gray!40, text width=2.5cm]

\tikzstyle{avgpool} = [rectangle, minimum width=2cm, minimum height=0.8cm, text centered, draw=black, fill=gray!40, text width=2.5cm]

\tikzstyle{concat} = [rectangle, minimum width=2cm, minimum height=0.8cm, text centered, draw=black, fill=gray!80, text width=2.5cm]

\tikzstyle{arrow} = [->,>=stealth]

\node (l0) [input] at (0,0) {\footnotesize Input (84x84)};

\node (l1) [convolution, below of=l0, yshift=-0.5cm] {\footnotesize Convolution k=5 (80x80)};

\node (l2) [convolution, below of=l1, yshift=-0.5cm] {\footnotesize Convolution k=5 (76x76)};

\node (l3) [avgpool, below of=l2, yshift=-0.5cm] {\footnotesize Avgooling p=3 (25x25)};

\node (l4) [convolution, below of=l3, yshift=-0.5cm] {\footnotesize Convolution k=5 (21x21)};

\node (l5) [maxpool, below of=l4, yshift=-0.5cm, xshift=-2cm] {\footnotesize MaxPooling p=2 (10x10)};

\node (l6) [maxpool, below of=l4, yshift=-0.5cm, xshift=2cm] {\footnotesize MaxPooling p=2 (10x10)};

\node (l7) [maxpool, below of=l5, yshift=-0.5cm] {\footnotesize MaxPooling p=2 (5x5)};

\node (l8) [convolution, below of=l6, yshift=-0.5cm] {\footnotesize Convolution k=5 (6x6)};

\node (l9) [maxpool, below of=l7, yshift=-0.5cm] {\footnotesize MaxPooling p=2 (2x2)};

\node (l10) [avgpool, below of=l8, yshift=-0.5cm] {\footnotesize AvgPooling p=3 (3x3)};

\node (l11) [concat, below of=l10, yshift=-0.5cm, xshift=-2cm] {\footnotesize Concat (3x3)};

\draw [arrow] (l0) -- (l1);
\draw [arrow] (l1) -- (l2);
\draw [arrow] (l2) -- (l3);
\draw [arrow] (l3) -- (l4);
\draw [arrow] (l4) -- (l5);
\draw [arrow] (l4) -- (l6);
\draw [arrow] (l5) -- (l7);
\draw [arrow] (l6) -- (l8);
\draw [arrow] (l7) -- (l9);
\draw [arrow] (l8) -- (l10);
\draw [arrow] (l9) -- (l11);
\draw [arrow] (l10) -- (l11);

\end{tikzpicture}
\caption{}
\label{fig:app:networks:multibranch:a}
\end{center}
\end{subfigure}%
\begin{subfigure}{.40\textwidth}
\begin{center}
\begin{tikzpicture}

\tikzstyle{input} = [rectangle, minimum width=2cm, minimum height=0.8cm, text centered, draw=black, fill=gray!0, text width=2cm]

\tikzstyle{convolution} = [rectangle, minimum width=2cm, minimum height=0.8cm, text centered, draw=black, fill=gray!10, text width=2.5cm]

\tikzstyle{maxpool} = [rectangle, minimum width=2cm, minimum height=0.8cm, text centered, draw=black, fill=gray!40, text width=2.5cm]

\tikzstyle{avgpool} = [rectangle, minimum width=2cm, minimum height=0.8cm, text centered, draw=black, fill=gray!40, text width=2.5cm]

\tikzstyle{concat} = [rectangle, minimum width=2cm, minimum height=0.8cm, text centered, draw=black, fill=gray!80, text width=2.5cm]

\tikzstyle{arrow} = [->,>=stealth]

\node (l0) [input] at (0,0) {\footnotesize Input (84x84)};

\node (l1) [avgpool, below of=l0, yshift=-0.5cm] {\footnotesize AvgPooling p=2 (42x42)};

\node (l2) [convolution, below of=l1, yshift=-0.5cm] {\footnotesize Convolution k=5 (38x38)};

\node (l3) [convolution, below of=l2, yshift=-0.5cm] {\footnotesize Convolution k=3 (36x36)};

\node (l4) [maxpool, below of=l3, yshift=-0.5cm] {\footnotesize MaxPooling p=2 (18x18)};

\node (l5) [convolution, below of=l4, yshift=-0.5cm] {\footnotesize Convolution k=3 (16x16)};

\node (l6) [maxpool, below of=l5, yshift=-0.5cm] {\footnotesize MaxPooling p=2 (8x8)};

\node (l7) [maxpool, below of=l6, yshift=-0.5cm] {\footnotesize MaxPooling p=2 (4x4)};

\node (l8) [convolution, below of=l7, yshift=-0.5cm] {\footnotesize Convolution k=1 (4x4)};

\draw [arrow] (l0) -- (l1);
\draw [arrow] (l1) -- (l2);
\draw [arrow] (l2) -- (l3);
\draw [arrow] (l3) -- (l4);
\draw [arrow] (l4) -- (l5);
\draw [arrow] (l5) -- (l6);
\draw [arrow] (l6) -- (l7);
\draw [arrow] (l7) -- (l8);

\end{tikzpicture}
\small
\caption{}
\label{fig:app:networks:multibranch:b}
\end{center}
\end{subfigure}
\caption{Best architectures designed for during the experiment in a multi-branch search space. (a) The best architecture when $\sigma=0.0$, with early-stop accuracy of 66.10. (b) The best architecture when $\sigma=0.1$, with early-stop accuracy of 66.45}
\label{fig:app:networks:multibranch}
\vspace{-0.5cm}
\end{figure}
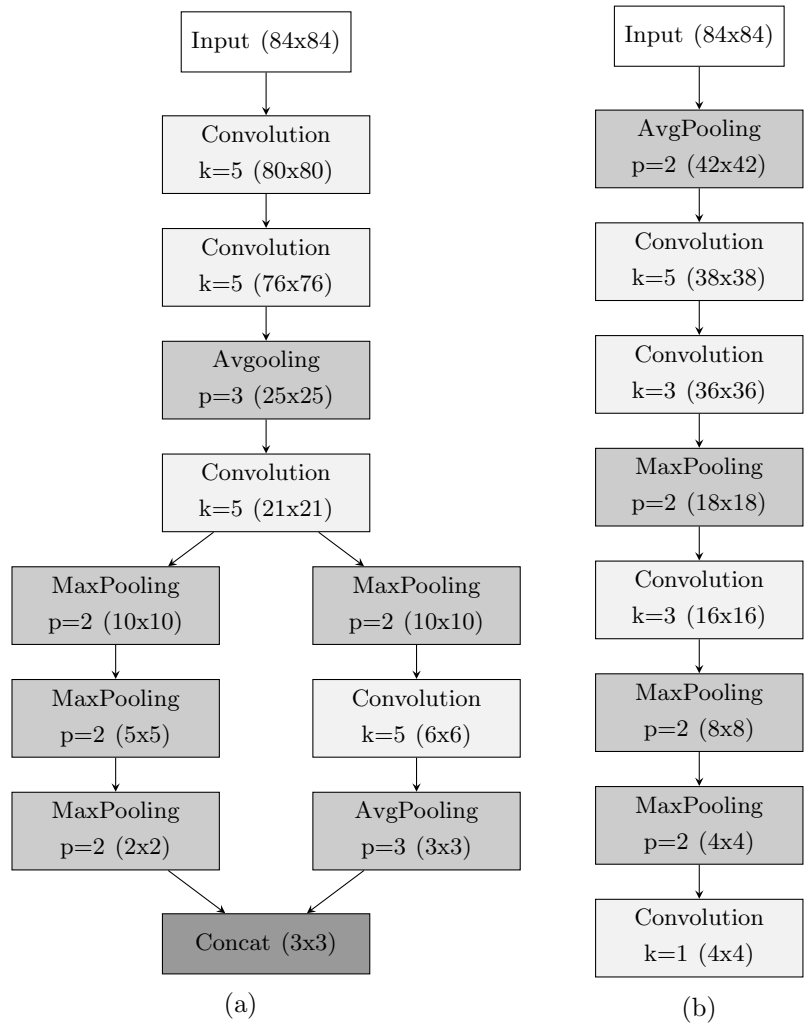

Here we show the best architectures designed by the agent in the three experiments. Figure~\ref{fig:app:networks:training} shows the best architecture per datasets during training (\textit{omniglot}, \textit{vgg\_flower}, and \textit{dtd}). Figure~\ref{fig:app:networks:aircraft} and~\ref{fig:app:networks:cubirds} show the best two architectures during evaluation for \textit{aircraft} and \textit{cu\_birds} respectively. Figure~\ref{fig:app:networks:multibranch} shows the best architectures for the multi-branch experiment. For each architecture we report the early-stop accuracy obtained.

\begin{figure}[ht]
\centering
\begin{subfigure}{.33\textwidth}
\begin{center}
\begin{tikzpicture}

\tikzstyle{input} = [rectangle, minimum width=2cm, minimum height=0.8cm, text centered, draw=black, fill=gray!0, text width=2cm]

\tikzstyle{convolution} = [rectangle, minimum width=2cm, minimum height=0.8cm, text centered, draw=black, fill=gray!10, text width=2.5cm]

\tikzstyle{maxpool} = [rectangle, minimum width=2cm, minimum height=0.8cm, text centered, draw=black, fill=gray!40, text width=2.5cm]

\tikzstyle{avgpool} = [rectangle, minimum width=2cm, minimum height=0.8cm, text centered, draw=black, fill=gray!40, text width=2.5cm]

\tikzstyle{concat} = [rectangle, minimum width=2cm, minimum height=0.8cm, text centered, draw=black, fill=gray!80, text width=2.5cm]

\tikzstyle{arrow} = [->,>=stealth]

\node (l0) [input] at (0,0) {\footnotesize Input (84x84)};

\node (l1) [convolution, below of=l0, yshift=-0.5cm] {\footnotesize Convolution k=5 (80x80)};

\node (l2) [maxpool, below of=l1, yshift=-0.5cm] {\footnotesize MaxPooling p=3 (26x26)};

\node (l3) [convolution, below of=l2, yshift=-0.5cm] {\footnotesize Convolution k=1 (26x26)};

\node (l4) [convolution, below of=l3, yshift=-0.5cm] {\footnotesize Convolution k=5 (22x22)};

\node (l5) [convolution, below of=l4, yshift=-0.5cm] {\footnotesize Convolution k=1 (22x22)};

\node (l6) [convolution, below of=l5, yshift=-0.5cm] {\footnotesize Convolution k=5 (18x18)};

\node (l7) [convolution, below of=l6, yshift=-0.5cm] {\footnotesize Convolution k=1 (18x18)};

\node (l8) [maxpool, below of=l7, yshift=-0.5cm] {\footnotesize MaxPooling p=2 (9x9)};

\node (l9) [avgpool, below of=l8, yshift=-0.5cm] {\footnotesize AvgPooling p=3 (3x3)};

\draw [arrow] (l0) -- (l1);
\draw [arrow] (l1) -- (l2);
\draw [arrow] (l2) -- (l3);
\draw [arrow] (l3) -- (l4);
\draw [arrow] (l4) -- (l5);
\draw [arrow] (l5) -- (l6);
\draw [arrow] (l6) -- (l7);
\draw [arrow] (l7) -- (l8);
\draw [arrow] (l8) -- (l9);

\end{tikzpicture}
\caption{}
\label{fig:app:networks:training:omniglot}
\end{center}
\end{subfigure}%
\begin{subfigure}{.33\textwidth}
\smallskip
\smallskip
\smallskip
\smallskip
\smallskip
\smallskip
\smallskip
\smallskip
\smallskip
\smallskip
\smallskip
\smallskip
\smallskip
\smallskip
\smallskip
\smallskip
\smallskip
\smallskip
\smallskip
\smallskip
\smallskip
\begin{center}
\begin{tikzpicture}

\tikzstyle{input} = [rectangle, minimum width=2cm, minimum height=0.8cm, text centered, draw=black, fill=gray!0, text width=2cm]

\tikzstyle{convolution} = [rectangle, minimum width=2cm, minimum height=0.8cm, text centered, draw=black, fill=gray!10, text width=2.5cm]

\tikzstyle{maxpool} = [rectangle, minimum width=2cm, minimum height=0.8cm, text centered, draw=black, fill=gray!40, text width=2.5cm]

\tikzstyle{avgpool} = [rectangle, minimum width=2cm, minimum height=0.8cm, text centered, draw=black, fill=gray!40, text width=2.5cm]

\tikzstyle{concat} = [rectangle, minimum width=2cm, minimum height=0.8cm, text centered, draw=black, fill=gray!80, text width=2.5cm]

\tikzstyle{arrow} = [->,>=stealth]

\node (l0) [input] at (0,0) {\footnotesize Input (84x84)};

\node (l1) [convolution, below of=l0, yshift=-0.5cm] {\footnotesize Convolution k=3 (82x82)};

\node (l2) [convolution, below of=l1, yshift=-0.5cm] {\footnotesize Convolution k=3 (80x80)};

\node (l3) [maxpool, below of=l2, yshift=-0.5cm] {\footnotesize MaxPooling p=2 (40x40)};

\node (l4) [maxpool, below of=l3, yshift=-0.5cm] {\footnotesize MaxPooling p=2 (20x20)};

\node (l5) [maxpool, below of=l4, yshift=-0.5cm] {\footnotesize MaxPooling p=2 (10x10)};

\node (l6) [maxpool, below of=l5, yshift=-0.5cm] {\footnotesize MaxPooling p=2 (5x5)};





\draw [arrow] (l0) -- (l1);
\draw [arrow] (l1) -- (l2);
\draw [arrow] (l2) -- (l3);
\draw [arrow] (l3) -- (l4);
\draw [arrow] (l4) -- (l5);
\draw [arrow] (l5) -- (l6);

\end{tikzpicture}
\small
\smallskip
\smallskip
\smallskip
\smallskip
\smallskip
\smallskip
\smallskip
\smallskip
\smallskip
\smallskip
\smallskip
\smallskip
\smallskip
\smallskip
\smallskip
\smallskip
\smallskip
\smallskip
\smallskip
\smallskip
\smallskip
\smallskip
\caption{}
\label{fig:app:networks:training:vggflower}
\end{center}
\end{subfigure}%
\begin{subfigure}{.33\textwidth}
\smallskip
\smallskip
\smallskip
\smallskip
\smallskip
\smallskip
\smallskip
\smallskip
\smallskip
\smallskip
\smallskip
\smallskip
\smallskip
\smallskip
\begin{center}
\begin{tikzpicture}

\tikzstyle{input} = [rectangle, minimum width=2cm, minimum height=0.8cm, text centered, draw=black, fill=gray!0, text width=2cm]

\tikzstyle{convolution} = [rectangle, minimum width=2cm, minimum height=0.8cm, text centered, draw=black, fill=gray!10, text width=2.5cm]

\tikzstyle{maxpool} = [rectangle, minimum width=2cm, minimum height=0.8cm, text centered, draw=black, fill=gray!40, text width=2.5cm]

\tikzstyle{avgpool} = [rectangle, minimum width=2cm, minimum height=0.8cm, text centered, draw=black, fill=gray!40, text width=2.5cm]

\tikzstyle{concat} = [rectangle, minimum width=2cm, minimum height=0.8cm, text centered, draw=black, fill=gray!80, text width=2.5cm]

\tikzstyle{arrow} = [->,>=stealth]

\node (l0) [input] at (0,0) {\footnotesize Input (84x84)};

\node (l1) [convolution, below of=l0, yshift=-0.5cm] {\footnotesize Convolution k=3 (82x82)};

\node (l2) [convolution, below of=l1, yshift=-0.5cm] {\footnotesize Convolution k=1 (82x82)};

\node (l3) [convolution, below of=l2, yshift=-0.5cm] {\footnotesize Convolution k=3 (80x80)};

\node (l4) [maxpool, below of=l3, yshift=-0.5cm] {\footnotesize MaxPooling p=2 (40x40)};

\node (l5) [maxpool, below of=l4, yshift=-0.5cm] {\footnotesize MaxPooling p=2 (20x20)};

\node (l6) [maxpool, below of=l5, yshift=-0.5cm] {\footnotesize MaxPooling p=2 (10x10)};

\node (l7) [maxpool, below of=l6, yshift=-0.5cm] {\footnotesize MaxPooling p=2 (5x5)};

\draw [arrow] (l0) -- (l1);
\draw [arrow] (l1) -- (l2);
\draw [arrow] (l2) -- (l3);
\draw [arrow] (l3) -- (l4);
\draw [arrow] (l4) -- (l5);
\draw [arrow] (l5) -- (l6);
\draw [arrow] (l6) -- (l7);

\end{tikzpicture}
\smallskip
\smallskip
\smallskip
\smallskip
\smallskip
\smallskip
\smallskip
\smallskip
\smallskip
\smallskip
\smallskip
\smallskip
\smallskip
\smallskip
\caption{}
\label{fig:app:networks:training:dtd}
\end{center}
\end{subfigure}
\caption{Best architectures designed for the training datasets. (a) The best architecture for \textit{omniglot}, with early-stop accuracy of 67.11. (b) The best architecture for \textit{vgg\_flower}, with early-stop accuracy of 57.15. (c) The best architecture for \textit{dtd}, with early-stop accuracy of 29.43}
\label{fig:app:networks:training}
\vspace{-0.5cm}
\end{figure}
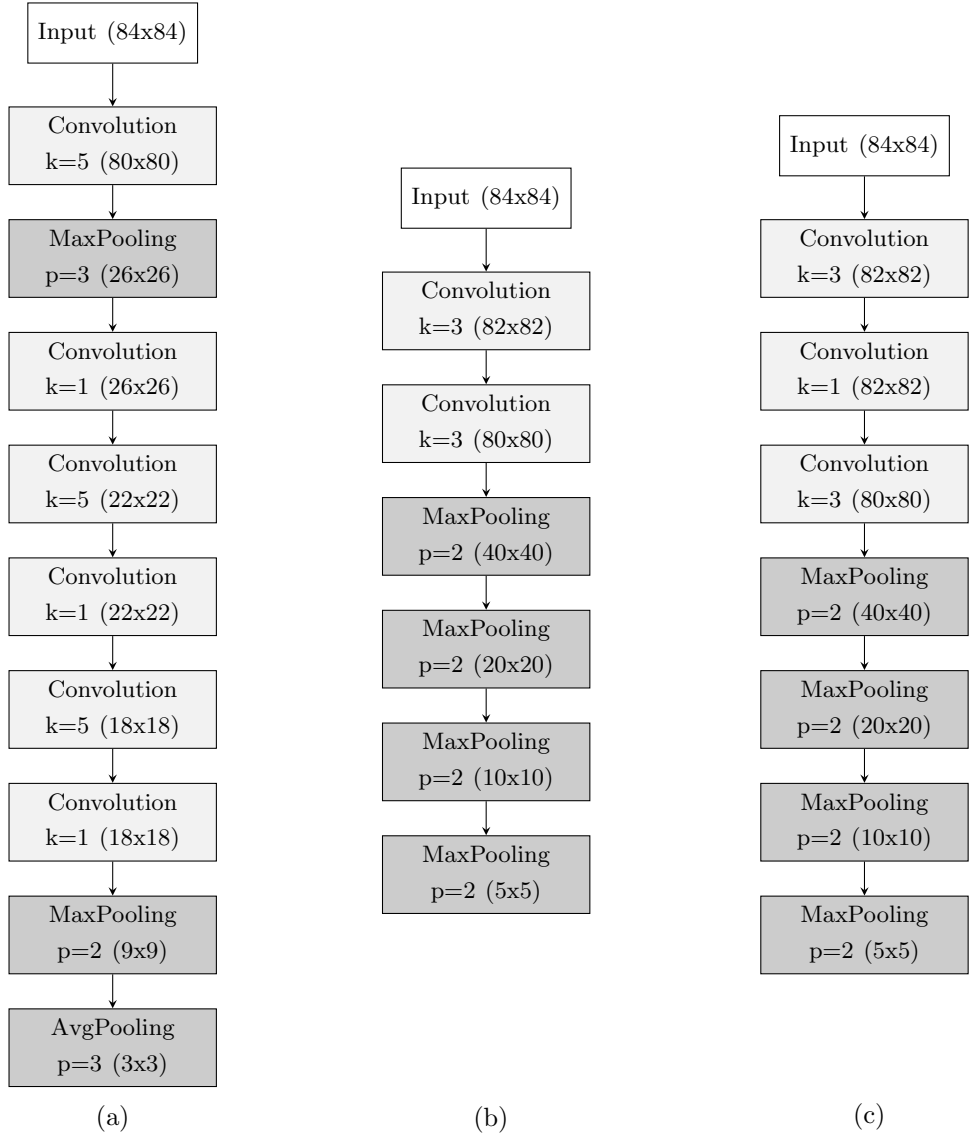

